%% file: main.tex
\documentclass[10pt,journal,compsoc]{IEEEtran}
\ifCLASSOPTIONcompsoc
  \usepackage[nocompress]{cite}
\else
  \usepackage{cite}
\fi
\ifCLASSINFOpdf
\else
\fi
\input{math_commands.tex}

\usepackage{tikz}
\usepackage{pgfplots}
\pgfplotsset{compat=default}
\usetikzlibrary{pgfplots.groupplots}

\usepackage{graphicx}
\usepackage{amsmath}
\usepackage{amssymb}
\usepackage{booktabs}
\usepackage{algpseudocode}
\usepackage{algorithm}
\usepackage{subcaption}
\usepackage{bbm}
\usepackage{pifont}
\usepackage{multirow}
\usepackage{tabu}
\usepackage{multicol}
\usepackage{booktabs}
\usepackage{rotating}

\usepackage{amsmath}
\usepackage{amssymb}
\usepackage{mathtools}
\usepackage{multirow}
\usepackage{array, makecell}
\usepackage{tabularx}
\usepackage{tabu}
\usepackage{multicol}
\usepackage{booktabs}
\usepackage{adjustbox}
\usepackage[normalem]{ulem}

\usepackage[pagebackref,breaklinks,colorlinks]{hyperref}

\usepackage[capitalize]{cleveref}
\crefname{section}{Sec.}{Secs.}
\Crefname{section}{Section}{Sections}
\Crefname{table}{Table}{Tables}
\crefname{table}{Tab.}{Tabs.}

\usepackage{xspace}

\makeatletter
\DeclareRobustCommand\onedot{\futurelet\@let@token\@onedot}
\def\@onedot{\ifx\@let@token.\else.\null\fi\xspace}

\def\eg{\emph{e.g}\onedot} 
\def\ie{\emph{i.e}\onedot} 
\def\cf{\emph{cf}\onedot} 
 \def\vs{\emph{vs}\onedot}

\makeatother

\makeatletter
\@namedef{ver@everyshi.sty}{}
\makeatother

\begin{document}
%
\title{ MED-VT++: Unifying Multimodal Learning with a Multiscale Encoder-Decoder Video Transformer}
%
%
%
%


\author{
{\small $^{1}$York University, $^{2}$Ontario Tech University}
\\
{\small 
\texttt{\{karimr31, zhufl, msiam\}@eecs.yorku.ca}, \texttt{wildes@cse.yorku.ca}}
}

\author{Rezaul Karim,
He Zhao, 
Richard P. Wildes, 
Mennatullah Siam
\\
\IEEEcompsocitemizethanks{
\IEEEcompsocthanksitem R.\ Karim, H.\ Zhao and R.\ Wildes are with the Department
of Electrical Engineering and Computer Science, York University, Canada.\\
M.\ Siam is with the Ontario Tech University, Canada.\\
E-mail:%
\texttt{karimr31@eecs.yorku.ca,zhufl@eecs.yorku.ca, wildes@cse.yorku.ca,mennatullah.siam@ontariotechu.ca}}
\thanks{Manuscript received January 17, 2023; }}

\IEEEtitleabstractindextext{%
\begin{abstract}
In this paper, we present an end-to-end trainable unified multiscale encoder-decoder transformer that is focused on dense prediction tasks in video. The presented Multiscale Encoder-Decoder Video Transformer (MED-VT) uses multiscale representation throughout and employs an optional input beyond video (e.g., audio), when available, for multimodal processing (MED-VT++). Multiscale representation at both encoder and decoder yields three key benefits: (i) implicit extraction of spatiotemporal features at different levels of abstraction for capturing dynamics without reliance on input optical flow, (ii) temporal consistency at encoding and (iii) coarse-to-fine detection for high-level (e.g., object) semantics to guide precise localization at decoding. Moreover, we present a transductive learning scheme through many-to-many label propagation to provide temporally consistent video predictions. We showcase MED-VT/MED-VT++ on three unimodal video segmentation tasks (Automatic Video Object Segmentation (AVOS), actor-action segmentation and Video Semantic Segmentation (VSS)) as well as a multimodal segmentation task (Audio-Visual Segmentation (AVS)). Results show that the proposed architecture outperforms alternative state-of-the-art approaches on multiple benchmarks using only video (and optional audio) as input, without reliance on optical flow. Finally, to document details of the model's internal learned representations, we present a detailed interpretability study, encompassing both quantitative and qualitative analyses.
\end{abstract}
}

\maketitle

\IEEEdisplaynontitleabstractindextext

%
\IEEEpeerreviewmaketitle


%
%
%
%

\input{medvt/1_intro}

\input{medvt/2_related_works}

\input{medvt/3_method}

\input{medvt/4_tasks}

\input{medvt/5_experiments}

\input{medvt/6_interpret}

\input{medvt/7_conclusion}
\bibliographystyle{IEEEtran}
\bibliography{main}

%

\begin{IEEEbiography}[{\includegraphics[width=1.1in,height=1.25in,clip,keepaspectratio]{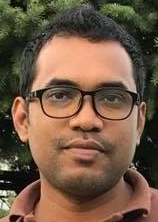}}]{Rezaul Karim}
is currently a Ph.D. candidate at York University, Canada, supervised by Professor Richard P. Wildes, with a research focus on deep learning for video understanding. He received his B. Sc. in Computer Science and Engineering from BUET, Bangladesh and his M.Sc. in Computer Science from the University of Manitoba, Canada. Honours include the VISTA Graduate Scholarship, Lassonde School of Engineering Carswell Scholarship, Lassonde Graduate Entrance Scholarship, Manitoba Graduate Scholarship and University of Manitoba Graduate Fellowship. His main areas of research interest are computational vision and
artificial intelligence.
\end{IEEEbiography}

\begin{IEEEbiography}[{\includegraphics[width=1.0in,clip,keepaspectratio]{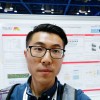}}]{He Zhao}
is currently a machine learning researcher at Borealis AI, Canada. He received his Ph.D. from York University, Canada, supervised by Professor Richard P. Wildes, with a research focus on deep learning for video understanding. He received his B. Sc. in Electrical and Electronics Engineering from Zhengzhou University, China and his M.Sc. in Computer Science from the University of Florida, USA. His main areas of research interest are computational vision and
artificial intelligence.
\end{IEEEbiography}


\begin{IEEEbiography}[{\includegraphics[width=1in,height=1.25in,clip,keepaspectratio]{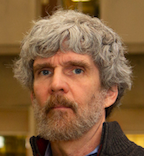}}]{Richard P. Wildes} (Member, IEEE) received the Ph.D. degree from the Massachusetts Institute of Technology in 1989. Subsequently, he joined Sarnoff Corporation in Princeton, New Jersey, as a Member of the Technical Staff in the Vision Technologies Lab. In 2001, he joined the Department
of Electrical Engineering and Computer
Science at York University, Toronto, where he is
a Professor, a member of the Centre for Vision
Research and a Tier I York Research Chair. He also is a visiting research scientist at Samsung Artificial Intelligence Center (SAIC), Toronto.
Honours include receiving a Sarnoff Corporation
Technical Achievement Award, the IEEE D.G. Fink Prize Paper Award
and twice giving invited presentations to the US National Academy of
Sciences. His main areas of research interest are computational vision,
especially video understanding, and artificial intelligence.
\end{IEEEbiography}

\begin{IEEEbiography}[{\includegraphics[width=1.1in,height=1.5in,clip,keepaspectratio]{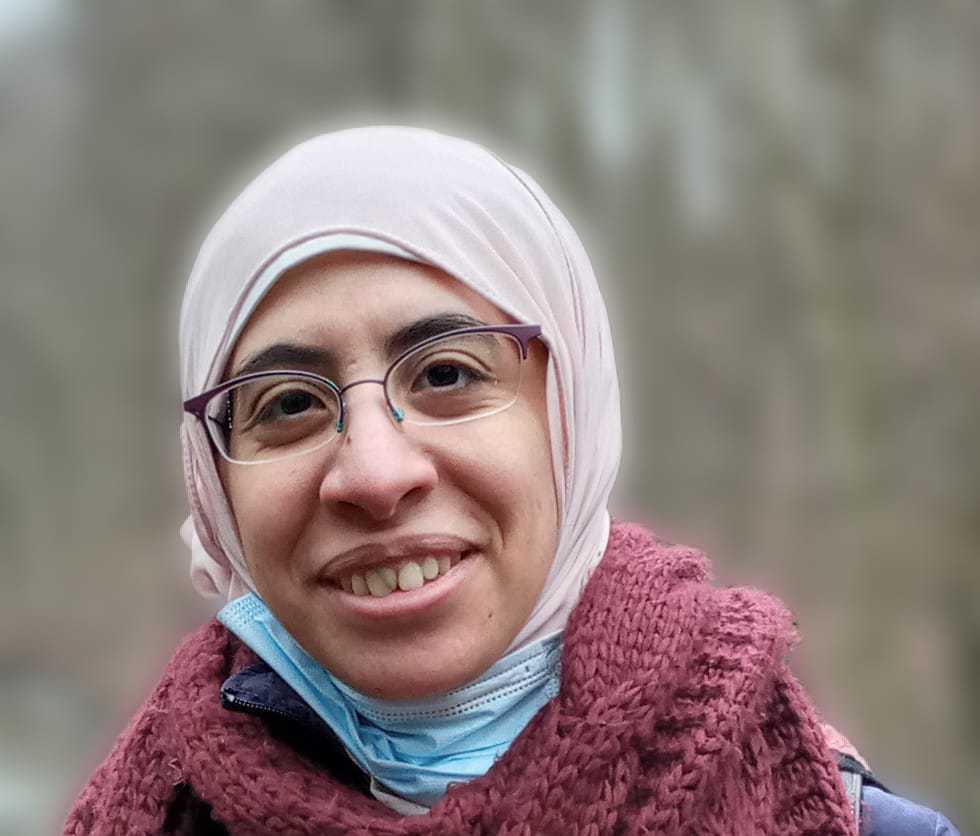}}]{Mennatullah Siam}
is an Assistant Professor in Electrical, Computer and Software Engineering department at Ontario Tech University, where she joined in July 2023. Her research expertise is in the area of computer vision, deep learning and robot vision, where she has published in top-tier conferences, \eg CVPR, ICCV, ICRA and IROS. Her work is focused on pixel-level video understanding and its interplay with data efficient learning techniques. She was a postdoctoral fellow at York University with Professors Richard P. Wildes and Konstantinos G. Derpanis as well as a Vector research affiliate. She previously obtained her PhD from University of Alberta with Professor Martin Jagersand with work on the intersection of video object segmentation and few- shot object segmentation. She is also an active promoter of equity in Computer Vision research through her community work in Deep Learning Indaba.
\end{IEEEbiography}



\end{document}

%% file: math_commands.tex
\usepackage{amsmath,amsfonts,bm}

\def\eqref#1{equation~\ref{#1}}

\def\1{\bm{1}}

\def\vs{{\bm{s}}}

\DeclareMathAlphabet{\mathsfit}{\encodingdefault}{\sfdefault}{m}{sl}
\SetMathAlphabet{\mathsfit}{bold}{\encodingdefault}{\sfdefault}{bx}{n}



%% file: medvt/1_intro.tex
\section{Introduction}\label{intro}

\IEEEPARstart{V}{ideo} transformers have shown tremendous ability to establish data relationships across space and time without the local biases inherent in convolutional and other similarly constrained approaches. This ability has resulted in application of video transformers to a wide range of video understanding tasks \cite{khan2022transformers}. Multiscale processing has the potential to enhance further the learning abilities of transformers through cross-scale learning \cite{chen2021crossvit,cheng2021masked,fan2021multiscale,li2021improved,wang2022crossformer}. Resulting multiscale features can combine the semantic richness available through deep-layer abstractions with fine localization available at shallower layers. Additionally, cross-attention between multiple modalities can yield complementary feature fusion in the encoder. This ability is achieved by attending to the relevant information from different modalities, which can facilitate effective query generation. For scientific \cite{sundararajan2017axiomatic}, application \cite{pan2021ia} and societal \cite{tan2018learning, wu2018beyond} reasons, such advances will be most impactful and widely deployed if their operations are interpretable. 

\begin{figure}[t!]
    \centering
    \resizebox{0.5\textwidth}{!}{ 
    \includegraphics[width=\textwidth]{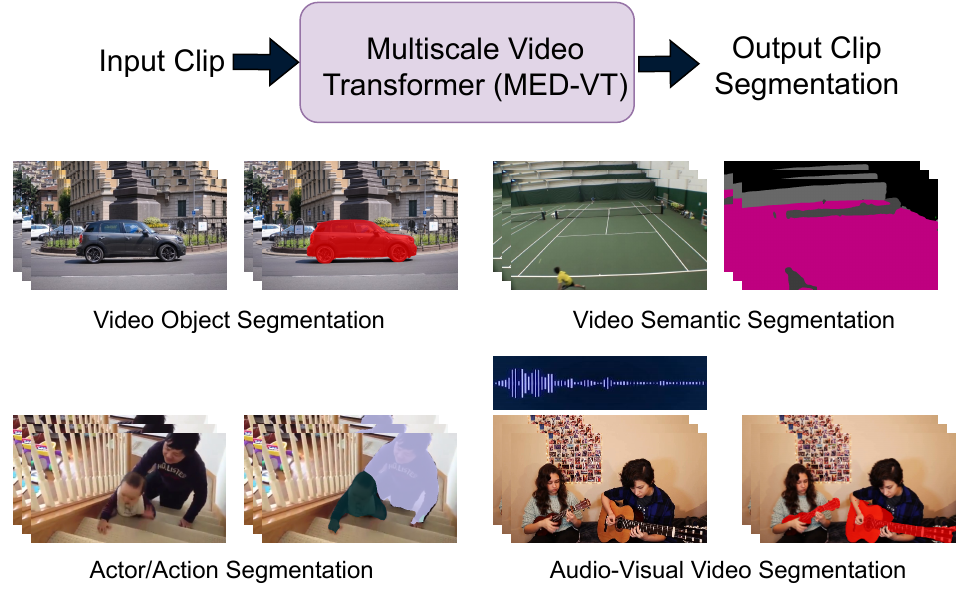}
    }
    \caption{Multiscale video transformer (MED-VT, MED-VT++) on Different Unimodal and Multimodal Video Segmentation Tasks. For unimodal tasks, such as primary object segmentation, video semantic segmentation and actor-action segmentation, MED-VT takes an input clip and estimates target masks. For multimodal video segmentation, \eg audio-visual segmentation, MED-VT++ takes audio as additional input and uses it as a context cue in segmentation. 
    }
    \label{fig:teaser}
\end{figure}

A gap exists, however, as prior research efforts focused on multiscale processing in either encoding or decoding in video transformers, leaving the potential integration throughout the model largely unaddressed. 
Recent work has focused on multiscale transformer encoding~\cite{fan2021multiscale,li2021improved}, yet does not incorporate multiscale processing in the transformer decoder. Other work has proposed multiscale transformer decoding~\cite{cheng2021masked}, yet was designed mainly for single images and did not consider the structured prediction nature inherent to video, \eg the importance of temporal consistency. 
Complementary work has investigated 
specialized architectures for multimodal feature fusion and query generation~\cite{  botach2022end,wu2022language,miao2023spectrum,pan2022wnet,zhou2022audio, mao2023contrastive, gao2023avsegformer}. However, such efforts have not led to models that can process both unimodal and multimodal data in a unified video transformer framework.
Notably, the majority of all the above approaches have given very limited attention to supporting their architectures with interpretability studies.

In response, we present the first Multiscale Encoder Decoder Video Transformer (MED-VT) equipped with a seamless multimodal extension (MED-VT++), which allows optional use of an additional modality when available. Figure~\ref{fig:teaser} overviews MED-VT/MED-VT++ as an architecture that is applicable to a wide range of video segmentation tasks. 
At encoding, its within and between-scale attention mechanisms allows it to capture both spatial, temporal and integrated spatiotemporal information. At decoding, it introduces learnable coarse-to-fine queries that allow for precise target delineation, while enforcing temporal consistency among predicted masks through transductive learning~\cite{vapnik200624,zhou2003learning}. Furthermore, we provide an interpretability analysis of our architecture to reveal the nature of its learned representations. 

We primarily illustrate the utility of MED-VT on the task of Automatic Video Object Segmentation (AVOS) and further extend to actor-action segmentation, Video Semantic Segmentation (VSS) and Audio-Visual Segmentation (AVS). All of these tasks are important and challenging, as they are enablers of many subsequent visually-guided operations, \eg, autonomous driving and augmented reality. In all cases, we operate in the unsupervised inference setting, \ie without using a mask in the first frame of a video to indicate the segmentation target, in contrast to the semi-supervised setting.
These tasks share challenges common to any video segmentation task (\eg target deformation and clutter). 
Notably, however, the requirement of complete automaticity imposes extra challenges, as there is no benefit of per video initialization.  Lacking prior information, solutions must exploit appearance (\eg, colour, texture and shape) as well as motion to garner as much information as possible. 

MED-VT and MED-VT++ respond to these challenges. Their within and between scale attention mechanisms capture both appearance and motion information as well as yield temporal consistency. Their learnable coarse-to-fine queries allow semantically laden information at deeper layers to guide finer scale features for precise object delineation. Their transductive learning through many-to-many label propagation ensures temporally consistent predictions.  

This paper builds on our conference paper~\cite{karim2023med} and
extends it in three major ways: First, to demonstrate the generalization of MED-VT across various video segmentation settings, we extend it to video semantic segmentation, which poses additional challenges due to the requirement of labelling all pixels. Second, we present a simple extension of MED-VT to encompass multimodal video segmentation, while staying within our multiscale framework (MED-VT++). 
We evaluate MED-VT++ on audio-visual segmentation. 
Third, we provide an extensive interpretability study on MED-VT to understand the details of its internal multiscale representations.

%% file: medvt/2_related_works.tex
 
\section{Related work}\label{related}
\textbf{Video segmentation.} We focus on two important aspects of video dense prediction: the multiscale nature of objects and temporally consistent spatial localization for per pixel classification, as well as operation without the expense of optical flow. For tasks, we consider three dense prediction tasks, Automatic Video Object Segmentation (AVOS)~\cite{wang2021survey,karim2023understanding}, actor-action segmentation~\cite{dang2018actor} and Video Semantic Segmentation (VSS)~\cite{yan2020video}, while leaving extensions to instance-aware AVOS~\cite{ventura2019rvos,luiten2020unovost} and tracking~\cite{oh2019video,xie2021efficient} for future work. Dominant approaches to AVOS rely on both colour images and optical flow as input ~\cite{zhou2020motion,ren2021reciprocal,siam2019video,jain2017fusionseg}. Other approaches consider attention to capture recurring objects in a video via simple mechanisms, \eg co-attention~\cite{wang2019zero,lu2019see} . Similarly, dominant approaches to actor-action segmentation depend on optical flow~\cite{dang2018actor,ji2018end}. Conversely, conventional VSS approaches rely on optical flow based warping to propagate their labels~\cite{gadde2017semantic,liu2017surveillance,ding2020every,jain2019accel}; however, other research instead uses inter-frame attention to capture the temporal relations between consecutive frames~\cite{paul2020efficient,wang2021temporal,li2021video,miao2021vspw}. More recent VSS approaches include a hierarchically structured transformer encoder~\cite{xie2021segformer},  dynamic kernels~\cite{li2022video}, inter-frame cross-attention~\cite{sun2022coarse} and tube prediction~\cite{kim2022tubeformer,li2023tube}. 

In complement, there has been considerable recent advances in multimodal video segmentation, with various task formulations, \eg text-referring video segmentation~\cite{xu2018youtube,khoreva2019video,gavrilyuk2018actor}, speech-referring video object segmentation~\cite{pan2022wnet} and audio-visual segmentation~\cite{zhou2022audio}. 
Recent work on audio-visual segmentation presented a specialized temporal pixel-wise audio-visual interaction~\cite{zhou2022audio}, a conditional latent diffusion model~\cite{mao2023contrastive} and transformer based query decoders with audio-visual mixer~\cite{gao2023avsegformer}. In contrast to prior efforts, we present a 
general framework to tackle both unimodal and multimodal video segmentation without domain specific specialized architectures.

\textbf{Multiscale processing.} Multiscale processing is an established technique across computer vision. Some representative examples in the era of convolutional networks include edge detection~\cite{hed}, image segmentation~\cite{unet}, object detection~\cite{lin2017} as well as AVOS~\cite{jain2017fusionseg, zhou2020motion, ren2021reciprocal}. Recently, multiscale processing has been applied with transformers to assist the understanding of single images (\eg classification~\cite{wang2022crossformer}, detection~\cite{zhu2020deformable, chen2021crossvit} and panoptic segmentation~\cite{cheng2021masked}) and videos (\eg action recognition~\cite{fan2021multiscale,li2021improved}). 
However, such transformers are limited by lack of unified multiscale processing  (\ie restricted to the encoding phase) or not readily applicable to video understanding (\ie primarily used for static images~\cite{wang2022crossformer}), in general, and dense video predictions, in particular. In contrast, while our work exploits multiscale information, it makes fuller use in its  multiscale encoder-decoder via \textit{Within} and \textit{Between} scale attention. 

\textbf{Temporal consistency.} 
AVOS models typically benefit from leveraging the principle of global consistency across multiple frames. Early efforts sought such consistency on the feature level by fusing the appearance (\eg RGB images) and motion information (\eg optical flow) of given videos~\cite{jain2017fusionseg, zhou2020motion, ren2021reciprocal}. However, these approaches required additional effort on high-quality flow estimation. Other work focused on enforcing consistency between features computed across time using co-attention~\cite{lu2019see}. A limitation of this approach is its excessive computational overhead, because of its need for multiple inference iterations to yield good results. Recent advances in semi-automatic VOS have devised a lightweight yet efficient counterpart: A prediction-level label propagator that explicitly exploits frame-wise semantic consistency, 
was proposed in a transductive inference setting~\cite{mao2021joint,zhang2020transductive}. Nonetheless, their propagator was confined to a single frame at a time. 

Common approaches in Video Semantic Segmentation (VSS) follow propagation-based segmentation from a key-frame (semi-supervised) or previous frames (automatic) to other frames using flow based warping to propagate their labels~\cite{gadde2017semantic,liu2017surveillance,ding2020every,jain2019accel}. More recent VSS approaches use inter-frame attention to capture the temporal relations between consecutive frames in the feature space, without depending on optical flow~\cite{paul2020efficient,wang2021temporal,li2021video,miao2021vspw,sun2022coarse} . 
Actor-action segmentation models focused on 3D convolution to address temporal consistency during feature extraction, \eg~\cite{ssa2d}. 

In contrast to all these approaches, we designed MED-VT to address temporal consistency throughout the model: Spatiotemporal attention in encoding and decoding yield temporally consistent features; a label propagator operates in the label space to produce consistent final masks. In particular, our label propagator extends existing approaches by considering many-to-many propagation to effectively capture temporal dependency within an entire input clip of a video.

\begin{figure*}[t!]
    \centering
    \includegraphics[width=\textwidth]{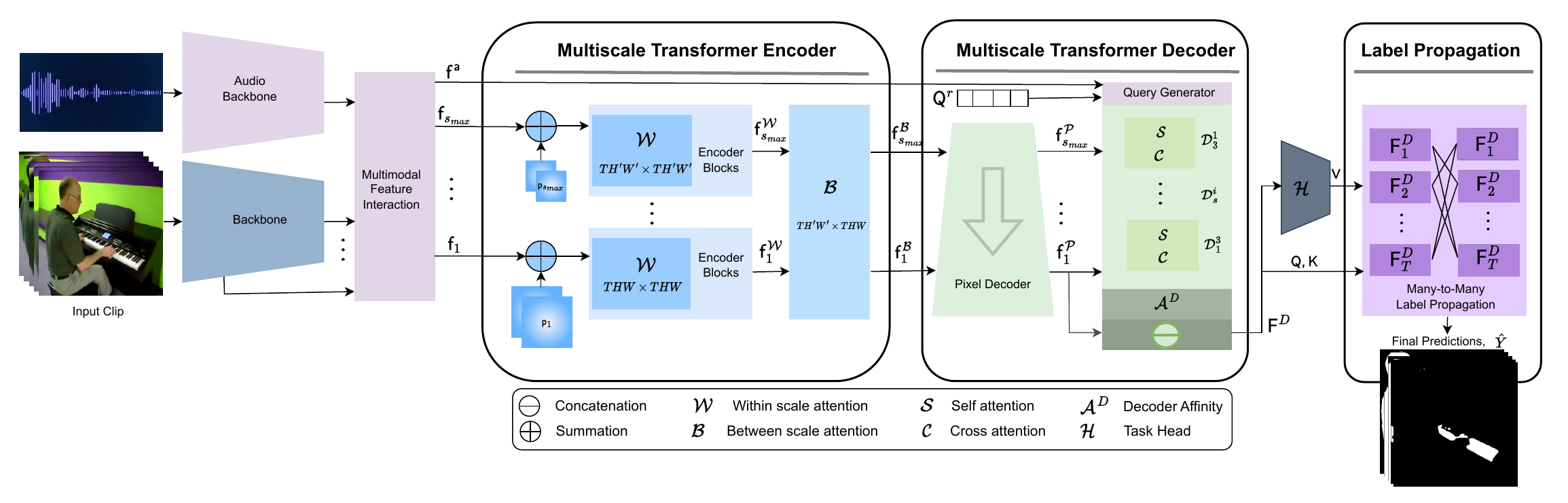} 
    \caption{Detailed MED-VT/MED-VT++ architecture with unified multiscale encoder-decoder transformer, illustrated with application to Audio-Visual Segmentation (AVS). The core MED-VT has five functionally distinct components. (i) Backbone feature extractor to extract per frame features, $\mathsf{f}_s$, at multiple scales, $s\in\{1,\cdots,s_{max}\}$. (ii) Multiscale transformer encoder consisting of spatiotemporal within and between scale attention with resulting features, $\mathsf{f}^{\mathcal{W}}_s$ and $\mathsf{f}^{\mathcal{B}}_s$, resp; the multihead attention transformation, \eqref{eq:multihead}, is used for both. (iii) Multiscale transformer decoder consisting of pixel decoding, which produces decoded features, $\mathsf{f}^{\mathcal{P}}_s$, and a series of mulitscale query learning decoder blocks, $\mathcal{D}^i_s$, for the corresponding $i^{th}$ iteration and scale $s$, each of which entail self and cross attention, again using the multihead attention transformation, \eqref{eq:multihead}. The input to the blocks are the decoded features $\mathsf{f}^{\mathcal{P}}_s$ and the query resulting from the previous block, with a randomized query, $\mathsf{Q}^r$, initialization; the output is a final object query, $\mathsf{Q}^o$. The decoder applies an affinity, \eqref{eq:decoder}, between $\mathsf{Q}^o$ and the finest scale decoded features, $\mathsf{f}^\mathcal{P}_1$, to yield an object attention map, which is concatenated with the finest scale decoded features for final decoder output, $\mathsf{F}^D$. (iv) A task specific head, $\mathcal{H}$, that inputs $\mathsf{F}^D$ to produce initial predictions. (v) Many-to-many label propagation, \eqref{eq:labelProp}, that inputs the initial predictions as values, $\mathsf{V}$,  as well as $\mathsf{F}^D$ as queries, $\mathsf{Q}$, and keys, $\mathsf{K}$, to yield temporally consistent segmentation final masks, $\hat{\mathsf{Y}}$. MED-VT++ adds three additional components: (vi) An additional backbone to extract auxiliary (\eg audio) features, (vii) Multimodal feature interaction that uses bidirectional cross-attention, \eqref{eq:bidir_xy}, and (viii) Multimodal query generator that initializes the query, $\mathsf{Q}^r$, from additional modality feature output from the feature fusion module (\eg, audio feature, $\mathsf{f}^a$), \eqref{eq:decoder_cross_attn_audio}, instead of random initialization used for unimodal segmentation. Our key innovations, outlined in bold boxes, lie in the unified multimodal multiscale encoder-decoder and label propagator. 
    }
    \label{fig:model:detail}
\end{figure*}

\textbf{Interpretability.} There is an increased interest in interpetability for deep-learning-based approaches to artificial intelligence, 
in general~\cite{zhang2021surveyinterpret}, and video segmentation, in particular\cite{karim2023understanding}. In our work, we make special use of a recently presented methodolgy for quantifying the extent to which architectures make use of static  information available in single frames (\eg color, texture and shape) \textit{vs.} dynamic information available from multiple images (\eg motion and dynamic texture)~\cite{kowal2022deeper}. A major finding of that study was that the majority of recent architectures for action recognition and segmentation were strongly biased toward static information. For qualitative analysis, we largely follow previous work that has visualized heat-maps of the first principle component of the high dimensional representations maintained internally by contemporary architectures, \eg~\cite{bakken2020principal,chefer2021generic,zhao2021interpretable,karim2023med,oquab2023dinov2}.



\textbf{Contributions.} In the light of previous work, our contributions are fivefold. (i) We present the first end-to-end multiscale transformer for video understanding that integrates multiscale encoding and decoding, while relying solely on image sequences without optical flow. 
The encoder enables our model to capture spatiotemporal information across scales, while the multiscale decoder provides precise localization. Lack of reliance on optical flow removes challenges in its estimation (\eg computational expense).
(ii) We present the first many-to-many label propagation scheme within a transductive learning paradigm to ensure temporally consistent predictions across an entire input clip. 
(iii) We introduce an attention-based extension from unimodal to multimodal video segmentation.
The approach adds optional modules at the encoder and decoder to integrate multimodal features when available; no domain specific modifications are made, which lends generality. 
(iv) We document the wide applicability of the architecture via empirical evaluation on four video segmentation tasks: Automatic Video Object Segmentation (AVOS), actor-action segmentation, semantic segmentation and Audio-Visual Segmentation (AVS).
(v) We provide an extensive interpretability analysis, both quantitative and qualitative, highlighting the importance of multiscale integration in temporally consistent and robust video segmentation. 
Our code is available at \href{https://rkyuca.github.io/medvt}{rkyuca.github.io/medvt}. 

%% file: medvt/3_method.tex
\section{MED-Video Transformer (MED-VT)}
\label{methods}



\subsection{Overview}\label{sec:overview} 

MED-VT is an end-to-end video transformer that inputs a clip and provides dense segmentation predictions without the need of explicit optical flow. Processing unfolds in five main stages; see Fig.~\ref{fig:model:detail}. (i) A feature pyramid is extracted using a backbone network. 
(ii) The extracted stacks of feature pyramids are processed by a transformer encoder and (iii) decoder. (iv) A task specific head produces initial predictions. (v) A many-to-many temporal label propagator refines the initial predictions by enforcing temporal consistency. Our architecture is unique in its unified approach to encoding and decoding at multiple scales, as well as its use of many-to-many label propagation. 



Feature extraction is standard. Given a video clip, we first extract a set of multiscale features, $F=\{\mathsf{f}_s:s\in S\}$, where $\mathsf{f}_s \in \mathbb{R}^{T \times H_s \times W_s \times C_s}$ represent features extracted at scale $s$, $S = \{1,..,s_{max}\}$ indexes the scale stages from fine to coarse and $\{T, H_s, W_s, C_s\}$ are the clip length, height, width and channel dimension at scale $s$, resp. 
Prior to subsequent processing, backbone features are down projected to $d$ dimensions, $\bar{\mathsf{f}}_s = \phi(\mathsf{f}_s) \in \mathbb{R}^{TH_sW_s\times d}$, where $\phi$ is a simple $1\times 1$ convolutional layer, followed by flattening. 
Our model is not specific to a particular backbone feature extractor; indeed, we illustrate with multiple in Sec.~\ref{exp}. Moreover, our model is not specific to a particular task head and we also illustrate with multiple in Sec.~\ref{exp}. The rest of this section, details our novel encoder-decoder, label propagator and an extension from single modality to multimodality input, \ie from purely video to video plus audio.

\subsection{Multiscale transformer encoder}\label{sec:encoder}
Transformer encoders built on spatiotemporal self-attention mechanisms can capture long range object representation relationships across both space and time for video recognition tasks~\cite{bertasius2021space,bulat2021space,fan2021multiscale}. They thereby naturally support learning of both spatial and temporal features as well as integrated spatiotemporal features.
Notably, however, standard encoders that operate over only coarse scale feature maps
limit the ability to capture fine grained pattern structure as well as fail to support precise localization \cite{wang2021end}. 
To overcome these limitations, our encoder encompasses two main operations of within and between-scale spatiotemporal attention on multiple feature abstraction levels with different resolutions that are extracted from a backbone convolutional network, as discussed in Sec.~\ref{sec:overview}.

We formulate the operations of within and between scale attention via standard multihead attention, $\mathcal{M}$, defined as
\begin{subequations}
    \begin{equation}
   \hspace*{-0.2cm} \mathcal{A}_h(\mathsf{Q}, \mathsf{K}, \mathsf{V}) = \mathcal{S}\left(\frac{1}{\sqrt{d}}\mathsf{Q}\mathsf{W}^q_h(\mathsf{K}\mathsf{W}^k_h)^\top\right)\mathsf{V}\mathsf{W}^v_h,
    \label{eq:singlehead}
\end{equation} 
\text{and}
\begin{equation}
    \mathcal{M}(\mathsf{Q}, \mathsf{K}, \mathsf{V}) =
    \text{Concat}_{h=1}^{N_h} (\mathcal{A}_h(\mathsf{Q}, \mathsf{K}, \mathsf{V}))\mathsf{W}^o,
    \label{eq:multihead}
\end{equation}
\end{subequations}
where $\mathsf{Q}$, $\mathsf{K}$ and $\mathsf{V}$ are query, key and value, resp., while $\mathsf{W}^q_h, \mathsf{W}^k_h$ and $\mathsf{W}^v_h$ are their corresponding learned weight matrices for head $h$, $d$ is feature dimension, $\mathcal{S}$ is the Softmax function and  $\mathsf{W}^o$ is the weight matrix for the final multiheaded output.

\textbf{Within scale attention.} We formulate multihead $\mathcal{W}$ithin scale attention by instantiating multihead attention, \eqref{eq:multihead}, as 
\begin{equation}
\mathcal{W}( \bar{\mathsf{f}}_s, \mathsf{p}_s) =  \mathcal{M}(\bar{\mathsf{f}}_s + \mathsf{p}_s, \bar{\mathsf{f}}_s + \mathsf{p}_s, \bar{\mathsf{f}}_s ), 
\end{equation}
with $\mathsf{p}_s \in \mathbb{R}^{TH_sW_s \times d}$ per scale positional encodings to preserve location information, \cf~\cite{wang2021end}.  
$\mathcal{W}$ is applied successively across multiple encoder layers. We compute the final encoded feature maps, $F^\mathcal{W} = \{\bar{\mathsf{f}}^{\mathcal{W}}_s:s\in S\}$, for each corresponding scale, to capture globally consistent representation of segmentation targets of interest.
Successive application of spatiotemporal within scale attention yields globally coherent representation, otherwise limited by local convolutions.


\textbf{Between scale attention.} For $\mathcal{B}$etween-scale attention, $\mathcal{B}$, we apply attention on the encoded features, $F^{\mathcal{W}}$. Coarse scale feature maps capture rich semantics by virtue of having gone through multiple abstraction layers. Correspondingly, the feature map from a coarser scale, $s$, \ie $\bar{\mathsf{f}}^{\mathcal{W}}_{s}$, is used to affect the immediately finer scale feature map, $\bar{\mathsf{f}}^{\mathcal{W}}_{s-1}$, based on between-scale attention. To achieve this goal, we again use multihead attention, \eqref{eq:multihead}, now as
\begin{equation}
\mathcal{B}( \bar{\mathsf{f}}^{\mathcal{W}}_{s-1},  \mathsf{p}_{s-1}, \bar{\mathsf{f}}^{\mathcal{W}}_{s}, \mathsf{p}_s )=  \mathcal{M}(\bar{\mathsf{f}}^{\mathcal{W}}_{s-1} + \mathsf{p}_{s-1}, \bar{\mathsf{f}}^{\mathcal{W}}_{s} + \mathsf{p}_s, \bar{\mathsf{f}}^{\mathcal{W}}_{s} ),
\end{equation}
where $\mathsf{p}_s$  and $\mathsf{p}_{s-1}$ are positional embeddings. This operation enhances between-scale communication to promote globally consistent, semantically rich feature maps and is conducted between each pair of adjacent scales.
We denote the output features from between-scale attention, $\mathcal{B}$, as $F^{\mathcal{B}} = \{ \bar{\mathsf{f}}^{\mathcal{B}}_s: s\in S\}$. 



\subsection{Multiscale transformer decoder}\label{sec:decoder}

Our multiscale encoder's between-scale attention promotes spatiotemporal consistency across scales, while the decoder promotes multiscale query learning to localize segmentation target level properties. Our decoder works in two steps: (i) pixel decoding, which propagates coarse scale semantics to fine scale localization and (ii) transformer decoding, which generates adaptive queries. 
\begin{figure}[t] 
    \centering
    \includegraphics[width=0.48\textwidth]{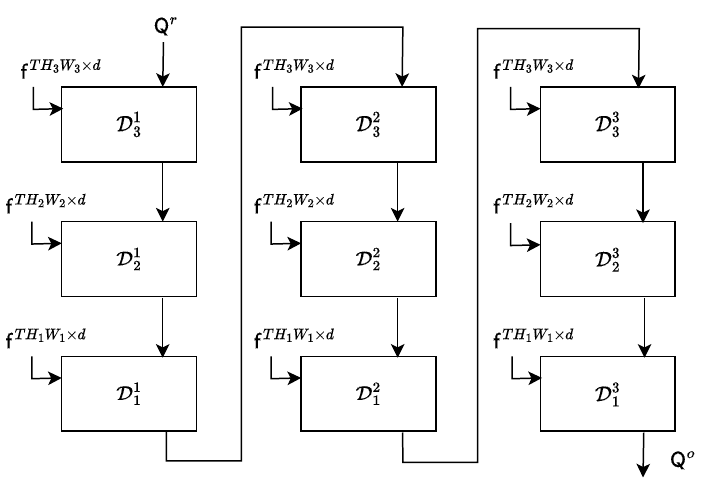} 
    \caption{The decoder stacked coarse-to-fine processing. Our multiscale decoder inputs a multiscale feature pyramid, $F^{\mathcal{P}}$, and randomly initialized queries, $\mathsf{Q}^r$, and outputs final object queries, $\mathsf{Q}^o$. The input is processed coarse-to-fine and iteratively through multiple decoder blocks, $\mathcal{D}^i_s$, with $s$ indicating input feature scale and $i$ indicating iteration. For simplicity, we show $s=3$ scales and $i=3$ iterations, with $\mathsf{f}^{\_}$ denoting features from each level of the pyramid, $F^{\mathcal{P}}$, where corresponding dimensions of the three levels are, $TH_3W_3\times d, TH_2W_2\times d, TH_1W_1\times d$, resp.}
    \label{fig:adaptive_queries}
    
\end{figure}

\textbf{Pixel decoding.} In pixel decoding we seek to propagate semantics of coarse scale features to finer scales. For this purpose, we use a Feature Pyramid Network (FPN) \cite{lin2017}. The FPN works top down from coarse features with highest abstraction to fine features by injecting coarser scale information into each finer scale. It thereby allows for better communication from high level to low level semantics with finer details preserved before queries are generated in the actual transformer decoder. The FPN inputs the between-scale attention features, $F^{\mathcal{B}}$, and outputs a feature pyramid $F^{\mathcal{P}}=\{\bar{\mathsf{f}}^{\mathcal{P}}_s:s \in S\}$. 

\textbf{Decoding and adaptive queries.} Improved object queries, $\mathsf{Q}^{o}$, are learned via multiscale coarse-to-fine processing, where the queries adapt to the input features for object localization. 
This multiscale processing enriches the learnable queries so they work better on the finest resolution. These adaptive queries, $\mathsf{Q}^{o} \in \mathbb{R}^{N_q\times d}$, with $N_q$ the number of queries, are learned jointly from the multiscale feature maps, $F^{\mathcal{P}}$, using a series of transformer decoder blocks, $\mathcal{D}^i_s$, operating coarse-to-fine across scales, $s$, and with multiple iterations, $i$; see Fig.~\ref{fig:adaptive_queries}.
Each transformer decoder block, $\mathcal{D}^i_s$, inputs pixel decoded features, $\bar{\mathsf{f}}^{\mathcal{P}}_s$, at a particular scale, $s$. At each iteration, $i$, the blocks operate, coarse-to-fine, with queries output from the previous serving as input (along with decoded features from  $F^{\mathcal{P}}$) to the next. The process iterates $N_d$ times, as notated by superscripts $i$ on the blocks, $\mathcal{D}^i_s$, \ie $i \in \{1...N_d\}$. The entire process starts with a randomly initialized query, $\mathsf{Q}^r$, and culminates in the final adaptive object queries, $\mathsf{Q}^o$. The adaptive queries serve to compactly represent the foreground object with its different appearance changes and deformations within the input clip.


In each decoder block both self and cross attention operate, as in encoding attention, Sec.~\ref{sec:encoder}. 
To define $\mathcal{S}$elf attention, we instantiate multihead attention, \eqref{eq:multihead}, as
\begin{equation}
    \mathcal{S}( \mathsf{Q}^i_s, \mathsf{p}^\mathsf{Q}_s)= \mathcal{M}(\mathsf{Q}^i_s+\mathsf{p}^\mathsf{Q}_s, \mathsf{Q}^i_s+\mathsf{p}^\mathsf{Q}_s, \mathsf{Q}^i_s),
    \label{eq:decoder_self_attn}
\end{equation} 
with $\mathsf{Q}^i_s$ the input query to block $\mathcal{D}^i_s$ and $p^\mathsf{Q}_s \in \mathbb{R}^{N_q \times d}$ learnable \textit{query positional embeddings}.
To define $\mathcal{C}$ross attention, we instead instantiate \eqref{eq:multihead} as
\begin{equation}
    \mathcal{C}(\mathsf{Q}^i_s, \mathsf{p}^\mathsf{Q}_s,  \bar{\mathsf{f}}^{\mathcal{P}}_s, \mathsf{p}_s, \hat{\mathsf{p}}^{\sigma}_s  )=\mathcal{M}(\mathsf{Q}^i_s + \mathsf{p}^\mathsf{Q}_s, \bar{\mathsf{f}}^{\mathcal{P}}_s + \mathsf{p}_s + \hat{\mathsf{p}}^{\sigma}_s, \bar{\mathsf{f}}^{\mathcal{P}}_s ),
    \label{eq:decoder_cross_attn}
\end{equation}
 with  $\mathsf{p}_s \in \mathbb{R}^{TH_sW_s \times d}$ \textit{feature positional embeddings} and  $\hat{\mathsf{p}}^{\sigma}_s \in \mathbb{R}^{TH_sW_s\times d}$ derived from learnable scale embeddings, $\mathsf{p}^{\sigma}_s \in \mathbb{R}^{1\times d}$, after being repeated across $T, H_s, W_s$; this operation allows cross attention to be scale sensitive, \cf~\cite{cheng2021masked}.

After all decoder blocks have produced the query, $\mathsf{Q}^o$, a final cross attention block is used to establish affinity between the query and finest scale features, $\bar{\mathsf{f}}^{\mathcal{P}}_{1}$, and thereby generate an object attention map, $\mathsf{F}^{\mathcal{A}}$. 
Since only the query and features are considered, a two argument affinity is used, 
\begin{equation} \label{eq:decoder}
\begin{split}
\mathsf{F}^{\mathcal{A}} & = \mathcal{A}^D(\mathsf{Q}, \mathsf{K}) \\
 & = \text{Concat}_{h=1}^{N_h}\left[\mathcal{S}\left(\frac{1}{\sqrt{d}}\mathsf{Q}\mathsf{W}^q_h(\mathsf{K}\mathsf{W}^k_h)^\top\right)\right]
\end{split}
\end{equation}
with $\mathsf{Q}=\mathsf{Q}^o, \mathsf{K}=\bar{\mathsf{f}}^{\mathcal{P}}_{1}$, $N_h$ the number of heads, $\mathsf{W}^q_h$ and $\mathsf{W}^k_h$ learnable parameters for head $h$ and $\mathcal{S}$ the Softmax function. 
The final decoder output, $\mathsf{F}^D$, is formed as the concatenation of the finest scale features, $\mathsf{f}^{\mathcal{P}}_{1}$, with the attention maps, $\mathsf{F}^{\mathcal{A}}$, \ie $\mathsf{F}^D = \mathsf{F}^{\mathcal{A}}\ominus \mathsf{f}^{\mathcal{P}}_{1}$, with $\ominus$ channel-wise concatenation. This augmentation further enhances localization precision.  

\subsection{Many-to-many temporal label propagation}\label{sec:lprop}
Label propagation is a standard technique that can be used in transductive reasoning ~\cite{zhou2003learning}. In semi-supervised VOS, it was proposed to train an end-to-end model to propagate the labels from many/all previous frames, $\{\cdots, t-1\}$, to a single current frame, $t$, hence causal many-to-one propagation within a transductive setting~\cite{mao2021joint}. This operation provides structured prediction across frames instead of independent predictions per frame. 
We extend this idea by allowing label propagation from all other frames, $\{\cdots, t-1, t+1, \cdots\}$, in a clip to each frame, $t$, hence many-to-many label propagation. This extended operation enforces structured prediction across all frames in a clip.

In our many-to-many label propagator, three operators are applied sequentially: (i) a label encoder, $\mathcal{E}_L$, (ii) a spatiotemporal affinity based label propagator using masked attention, $\mathcal{M}^m$, and (iii) a label decoder, $\mathcal{D}_L$. The input to the encoder is an initial prediction, $\mathsf{Y}^\prime =\mathcal{H}(\mathsf{F}^D)$, generated by a task head, $\mathcal{H}$, from the output of the previous decoding, $\mathsf{F}^D$. The label encoder then generates an encoding of dimension $D$ from the initial predictions, $\mathsf{Y}^\prime$, and flattens it, $ \bar{\mathsf{Y}} = \mathcal{E}_L( \mathsf{Y}^\prime) \in \mathbb{R}^{TH_1W_1 \times D}$, \cf~\cite{mao2021joint}.
The label propagator extends the encoded labels temporally in a many-to-many fashion. The label decoder, $\mathcal{D}_L$, takes these propagated encoded labels and generates the final class-wise predictions.
The label encoder, $\mathcal{E}_L$, is a similar CNN  to that used elsewhere~\cite{mao2021joint}. The label decoder, $\mathcal{D}_L$, is a three-layer CNN.

We devise the label propagator as a masked attention module~\cite{vaswani2017attention, sun2019videobert} to capture the long-distance dependencies between labels while preserving efficiency. The mask, $\mathsf{M} \in \mathbb{R}^{TH_1W_1 \times TH_1W_1}$, restricts attention to regions centered around the predicted data point, akin to the notion of \textit{clique} in conditional random fields~\cite{krahenbuhl2011efficient} or, more generally, graph theory. The mask can be defined to promote communication between data points in a wide variety of fashions (\eg, within frame, between frames, many-to-one, many-to-many)~\cite{krahenbuhl2011efficient}.
We use this mechanism for 
temporal many-to-many propagation to encourage information sharing among different frames. The mask of two arbitrary positions is set to, $\mathsf{M}_{ij} = -\infty$, if they are in the same frame, otherwise as zero. Formally, the mask is defined as

\begin{equation}
\mathsf{M}_{ij} = \begin{cases}
    -\infty,& \text{if } \tau(i) = \tau(j)\\
    0,              & \text{otherwise}.
\end{cases}
\end{equation}
where $\tau(\cdot)$ returns the frame index of given data points.
Masked attention is then defined by augmenting the standard multihead attention, \eqref{eq:multihead} and \eqref{eq:singlehead}, to become
\begin{subequations}
   \begin{equation}
   \begin{multlined}
    \mathcal{A}_h^m(\mathsf{Q}, \mathsf{K}, \mathsf{V}, \mathsf{M} ) = \\ 
    \mathcal{S}\left(\frac{1}{\sqrt{d}}\mathsf{Q}\mathsf{W}^q_h(\mathsf{K}\mathsf{W}^k_h)^\top + \mathsf{M} \right)\mathsf{V}\mathsf{W}^v_h,
   \end{multlined}
\label{eq:singlehead_masked}
\end{equation} 
\text{and}
\begin{equation}
    \mathcal{M}^m(\mathsf{Q}, \mathsf{K}, \mathsf{V}, \mathsf{M}) =
    \text{Concat}_{h=1}^{N_h} (\mathcal{A}_h^m(\mathsf{Q}, \mathsf{K}, \mathsf{V}, \mathsf{M}))\mathsf{W}^o,
    \label{eq:masked}
\end{equation}
\end{subequations}
where $\mathsf{Q}, \mathsf{K}, \mathsf{V}, \mathsf{M}$ are the queries, keys, values and mask resp., while $\mathsf{W}^q_h$, $\mathsf{W}^k_h$, $\mathsf{W}^v_h$ are learnable parameters for head $h$, $\mathcal{S}$ is the Softmax function and $W^o$ is the final weighting for combining heads. Our label propagator instantiates masked attention, \eqref{eq:masked}, as 
\begin{equation}\label{eq:labelProp}
    \tilde{\mathsf{Y}} = \mathcal{M}^m(\bar{\mathsf{F}}^{D}, \bar{\mathsf{F}}^{D}, \mathsf{\bar{Y}}, \mathsf{{M}} ), 
\end{equation}
where $\bar{\mathsf{F}}^{D} \in \mathbb{R}^{TH_1W_1\times (d+N_h)}$ is the flattened decoded features. As defined, this operation propagates labels across all data points in the entire clip, both spatial position and frames, unlike previous efforts that were limited to propagating to the current frame only~\cite{mao2021joint}. 

Finally, the overall class-wise predictions, $\hat{\mathsf{Y}}$, are produced by 
combining the propagated label decodings, $\mathcal{D}_L(\tilde{\mathsf{Y}})$, and the initial predictions from the segmentation head, $\mathsf{Y}^\prime$ 
according to
\begin{equation}
\hat{\mathsf{Y}} = \frac{1}{2}(\mathcal{D}_L(\tilde{\mathsf{Y}}) + {\mathsf{Y^\prime}}).\label{eq:kludge}
\end{equation}
We combine the initial predictions per frame and the propagated predictions from all other frames because while label propagation enforces temporal consistency, it also can sacrifice boundary precision due to the smoothing it incurs. The final combination, \eqref{eq:kludge}, provides both temporal consistency and precise boundaries. 

\subsection{Multimodal extension (MED-VT++)}\label{sec:medvtplus}

We have extended MED-VT to multimodal video segmentation tasks, \ie to encompass an additional input signal beyond video, as illustrated in Fig.~\ref{fig:model:detail}. The specific additional modality that we have considered is audio; although, our approach to incorporating that signal is not specific to that particular modality. 
To  extend MED-VT for this setting, we added three additional components: (i) a new modality feature extractor  (ii) an attention based multimodal fusion mechanism and (iii) a context-based query generator. For the sake of generality, we make no assumptions about the nature of the auxiliary modality, except that it can be represented as backbone features, $\mathsf{f}^a$.


We developed an attention based context module immediately after the feature extractors for information exchange between the features of two modalities. In particular, we use a bidirectional cross-attention mechanism~\cite{wangbidirattn} for information exchange between the audio features and each of the four stages of the visual features. The bidirectional cross attention mechanism is a simplification of cross-attention from one modality, $\mathsf{X}$, to the other modality, $\mathsf{Y}$, and vice versa, yet performed in one shared attention matrix to reduce compute and memory footprint. To provide a concrete explanation of this bidirectional multiheaded cross-attention mechanism, we show the computation for a single attention head since the multihead mechanism remains the same. In conventional cross attention, there are two inputs, say $\mathsf{X}$ and $\mathsf{Y}$, one of which is used as query, $\mathsf{Q}$, and the other as key and value, $\mathsf{K}$ and $\mathsf{V}$. Rewriting~\eqref{eq:singlehead} using two inputs for cross-attention with $\mathsf{Q=X}, \mathsf{K=Y}, \mathsf{V=Y}$ and renaming $\mathsf{W}^q_h,\mathsf{W}^k_h$ as $\mathsf{W}^x_h,\mathsf{W}^y_h$, we can formulate computation of a single head for cross-attention as
\begin{equation}
   \hspace*{-0.2cm} \mathsf{X^\prime} =\mathcal{A}_h(\mathsf{X}, \mathsf{Y}, \mathsf{Y}) =  \mathcal{S}\left(\frac{1}{\sqrt{d}}\mathsf{X}\mathsf{W}^x_h(\mathsf{Y}\mathsf{W}^y_h)^\top\right)\mathsf{Y}\mathsf{W}^v_h.
    \label{eq:singlehead_cross_xy}
\end{equation} 

Similarly, computation of a single head of another cross-attention with  $\mathsf{Q=Y}, \mathsf{K=X}, \mathsf{V=X}$ can be expressed as 
\begin{equation}
   \hspace*{-0.2cm} \mathsf{Y^\prime} =\mathcal{A}_h(\mathsf{Y}, \mathsf{X}, \mathsf{X}) = \mathcal{S}\left(\frac{1}{\sqrt{d}}\mathsf{Y}\mathsf{W}^y_h(\mathsf{X}\mathsf{W}^x_h)^\top\right)\mathsf{X}\mathsf{W}^v_h.
    \label{eq:singlehead_cross_yx}
\end{equation} 

Inspection of~\eqref{eq:singlehead_cross_xy} and~\eqref{eq:singlehead_cross_yx} shows that in realizing the two cross-attention operations, we can compute the attention matrix for one and reuse it for the other by taking the transpose. So, we can formulate a single head of the bidirectional cross-attention as 
\begin{subequations}
    \begin{equation}
   \hspace*{-0.2cm} \mathsf{A} =\frac{1}{\sqrt{d}}\mathsf{X}\mathsf{W}^x_h(\mathsf{Y}\mathsf{W}^y_h)^\top,
    \label{eq:attn_xy}
\end{equation} 
\text{and}
\begin{equation}
    \left(\mathsf{X^\prime}, \mathsf{Y^\prime}\right) = \left(\mathcal{S}\left(\mathsf{A}\right)\mathsf{Y}\mathsf{W}^v_h,  \mathcal{S}\left(\mathsf{A}^\top\right)\mathsf{X}\mathsf{W}^v_h\right). 
    \label{eq:bidir_xy}
\end{equation}
\end{subequations}
Finally, we include a residual connection from the same modality feature to compute, $\mathsf{X^{\prime\prime}} = \mathsf{X^\prime} + \mathsf{X}$, and similarly for $\mathsf{Y^{\prime\prime}}$. As standard~\cite{he2016deep}, the residual connection is used to avoid diminished gradients. 

MED-VT++ uses a bidirectional encoder for each scale, $s$, of the visual features. The same auxiliary features, $\mathsf{f}^a \in \mathbb{R}^{T \times c}$, are paired with all the scales of the visual features, where $T$ is the number of frames and $c$ is the dimension. As a preliminary, the backbone  visual, $\mathsf{f}_s$, and auxiliary features, $\mathsf{f}^a$, are projected to a common dimensionality, $d$, via separate linear layers and flattening to yield $\bar{\mathsf{f}}_s$ and $\bar{\mathsf{f}}^a$, resp. Then, as one input to the bidirectional encoder we use the projected visual features from each scale, \ie
$\mathsf{X_s}=\bar{\mathsf{f}}_s$ for the $s^{th}$ scale. 
For the other input, we always use the same projected auxiliary features, $\bar{\mathsf{f}}^a$, \ie  
$\mathsf{Y}_s=\bar{\mathsf{f}}^a$ for the $s^{th}$ scale. Using these definitions of $\mathsf{X}_s$ and $\mathsf{Y}_s$ in \eqref{eq:attn_xy} and \eqref{eq:bidir_xy} followed by residual connections yields a pair of visual, $\bar{\mathsf{f}}^{\prime\prime}_s$, and auxiliary, $\bar{\mathsf{f}}^{a\prime\prime}_s$, features for each scale, $s$. The visual features, $\bar{\mathsf{f}}^{\prime\prime}_s$, are  used as input to next stage of the transformer encoder, exactly as in the original MED-VT. The auxiliary features, $\bar{\mathsf{f}}^{a\prime\prime}_s$, are used to provide context to the MED-VT decoder, as described in the remainder of this section.

In MED-VT, for unimodal video segmentation, the initial queries of the transformer decoder, $\mathsf{Q}^r$, are randomly initialized and subsequently
updated through self- and cross-attention mechanisms to get object queries, $\mathsf{Q}^o$. When an additional modality is available, \eg audio, it can be used for context-based initialization. 
In particular, we input randomly initialized queries, $\mathsf{Q}^r$, and averaged bidirectionally encoded auxiliary features,
\begin{equation}\label{eq:avgAudio}
\tilde{\mathsf{f}}^a = \frac{1}{s_{max}} \sum\limits_{s=1}^{s_{max}} \bar{\mathsf{f}}^{a\prime\prime}_s, 
\end{equation}
to an additional transformer decoder, which operates prior to MED-VT's original decoder, to yield context-based queries, $\mathsf{Q}^a$. Here, the feature averaging is motivated by the fact that the additional decoder only inputs a single feature, not a set of (multiscale) features. 
The queries from this additional transformer decoder are used as input to the multiscale transformer decoder instead of $\mathsf{Q}^r$ (Section~\ref{sec:decoder}). Next, we detail how to obtain, $\mathsf{Q}^a$. 

Self-attention in the query generator is formulated analogously to \eqref{eq:decoder_self_attn} in the MED-VT decoder. Omitting the iterations, $i$, for simplicity, we have
\begin{equation}
    \mathcal{S}(\mathsf{Q}^r, \mathsf{p}^a)= \mathcal{M}(\mathsf{Q}^r+ \mathsf{p}^a, \mathsf{Q}^r+\mathsf{p}^a, \mathsf{Q}^r),
    \label{eq:decoder_self_attn_audio}
\end{equation} 
with $\mathsf{Q}^r$ the input to the query generator, $\mathsf{p}^a \in \mathbb{R}^{N_q \times d}$ learnable \textit{query positional embeddings} and the output taken as $\mathsf{Q}^a = \mathcal{S}(\mathsf{Q}^r, \mathsf{p}^a)$.
Analogously, cross-attention for the query generator is similar to \eqref{eq:decoder_cross_attn} and yields
\begin{equation}
    \mathcal{C}(\mathsf{Q}^a, \mathsf{p}^a,  \tilde{\mathsf{f}}^{a}, \tilde{\mathsf{p}}^a)=\mathcal{M}(\mathsf{Q}^a + \mathsf{p}^a, \tilde{\mathsf{f}}^{a} + \tilde{\mathsf{p}}^a, \tilde{\mathsf{f}}^{a} ),
\label{eq:decoder_cross_attn_audio}
\end{equation}
with  $\tilde{\mathsf{p}}^a \in \mathbb{R}^{T \times d}$ \textit{learnable audio feature positional embeddings} and $\mathsf{Q}^a = \mathcal{C}(\mathsf{Q}^a, \mathsf{p}^a,  \tilde{\mathsf{f}}^{a}, \tilde{\mathsf{p}}^a)$ taken as the final output queries. The self- and cross-attention operations, \eqref{eq:decoder_self_attn_audio} and \eqref{eq:decoder_cross_attn_audio}, are applied consecutively for $m$ decoding layers to yield the final query, $\mathsf{Q}^a$. These output queries serve as input to the MED-VT transformer decoder, \ie $\mathsf{Q}^0_s$ in \eqref{eq:decoder_self_attn} and \eqref{eq:decoder_cross_attn}. 


The rest of the MED-VT architecture remains the same, except that we do not employ label propagation. Label propagation is not used because of the nature of the task: The visual segmentation targets can change abruptly depending on the audio signal (\eg a person is segmented only in frames where they are talking, but not in other frames). 

Notably, MED-VT++ does not entail any modality specific changes to MED-VT. Features from a source other than audio could serve as complimentary to visual features in the bidrectional encoder. This generality opens the possibility of future work that explores other modalities with MED-VT++, \eg text guided referring expression segmentation~\cite{seo2020urvos}.


\subsection{End-to-end training}
Earlier~\cite{karim2023med}, a two-stage training scheme was employed for MED-VT, where the encoder-decoder is first trained and then the label propagator is trained while freezing the rest of the architecture. In contrast, our updated version employs an end-to-end training mechanism for the full model. As before, task specific losses are employed (detailed in Sec.~\ref{sec:tasks}), but to enable end-to-end training an additional auxiliary loss is applied in all cases. 
The main loss, $\mathcal{L}_m$, is with respect to the final model prediction. The auxiliary loss, $\mathcal{L}_a$, is with respect to the initial prediction directly following the task head, \ie before label propagation.
The same task specific losses are used in both. 
The final loss function, $\mathcal{L}$, is a weighted sum of the main and auxiliary losses,  \ie 
\begin{equation}\label{eq:loss}
    \mathcal{L} = w_m\mathcal{L}_m + w_a \mathcal{L},
\end{equation}
with $w_m$ and $w_a$ weights. 

This addition allows us to train the model end-to-end in a simpler and faster way compared to our previous training scheme. Moreover, this end-to-end training also resulted in better performance compared to the two-stage approach.

%% file: medvt/4_tasks.tex
\section{Tasks}\label{sec:tasks}

\textbf{Automatic video object segmentation.}
Automatic Video Object Segmentation (AVOS) is defined
as grouping pixels into foreground object masks representing the primary object in a video
sequence, where primary is distinguished as the object showing dominant motion or distinguishing
appearances across the frames~\cite{wang2021survey}. For this task, we train the model using a combination of distribution-based and region-based losses. In particular, we combine the distribution-based focal~\cite{lin2017focal} and the region-based Dice~\cite{milletari2016v} losses. Notably, this combination ameliorates the challenge of class imbalance ~\cite{lin2017focal,milletari2016v}, as in both our tasks, background pixels are more frequent than other classes. To support many-to-many label propagation in AVOS we use the entire clip ground  truth to compute the loss.

\textbf{Actor-action segmentation.}
Actor-action segmentation is defined as delineating regions from a given video with the corresponding
label of an actor-action tuple, with actors being objects that are participants in some actions~\cite{xu2015can}. For actor-action segmentation, we train the model using only distribution-based loss. In particular, we use the distribution-based multiclass focal cross entropy loss~\cite{lin2017focal}. For the A2D dataset that is used for evaluation, ground truth labels are available only for the centre frame. Therefore, we segment all the frames in a clip together, but minimize the loss for the center frame only. Since the whole clip is processed together, the model learns to segment all the frames in a clip with supervision from only the centre clip ground truth. This approach allows us to apply label propagation in a similar way as for the AVOS task. 


\textbf{Video semantic segmentation.}
Video Semantic Segmentation (VSS), also referred as Video Scene Parsing (VSP), aims to label all the pixels in a scene according to semantic categories of thing and stuff classes in an instance agnostic fashion~\cite{wang2021survey,yan2020video}. For VSS/VSP, the only change is in the number of outputs of the last layer of the task head, which is set to number of semantic classes. For this task, we train the model using distribution-based cross-entropy and the region-based Dice~\cite{milletari2016v} losses. We did not use any additional synthetic data augmentation techniques, such as clip-level copy-paste, which is used elsewhere~\cite{kim2022tubeformer}.

\textbf{Audio-visual segmentation.}
Audio-Visual Segmentation (AVS) is defined as segmenting objects that are the source of sounds in the corresponding audio stream captured together with the video~\cite{zhou2022audio}. 
There are two formulation of the problem: One aims to segment single sound sources; the second aims to segment multiple sound sources. This problem poses additional challenges, as the same object (\eg a person) is segmented as foreground in frames where it is producing sound (\eg singing or talking) and is considered as background in frames when not producing any sound (\ie silent). 
For training, we follow the same objectives used in earlier AVS work~\cite{zhou2022audio}, \ie cross-entropy loss and Kullback–Leibler (KL) divergence for the multiple sound source setting and only cross-entropy for the single sound source setting.

%% file: medvt/5_experiments.tex
\section{Empirical evaluation}
\label{exp}
\subsection{Experiment design}
\label{sec:exp_details}

\begin{table*}[h]
  \centering
  \aboverulesep=0ex
   \belowrulesep=0ex
   \begin{adjustbox}{max width=0.8\textwidth}
    \begin{tabular}{@{}cc|ccc|ccc@{}}
    \toprule
    \multicolumn{2}{c|}{\multirow{2}{*}{Measures}} & \multicolumn{3}{c|}{Uses RGB+Flow} & \multicolumn{2}{c}{Uses RGB only} \\
    \multicolumn{2}{c|}{}                   & COD (two-stream)~\cite{lamdouar2020betrayed}  &   MATNet~\cite{zhou2020motion}  &   RTNet~\cite{ren2021reciprocal}  &  COSNet~\cite{lu2019see}& \textbf{Ours} & \textbf{Ours$\dagger$}  
    \\ 
    \midrule
    \multirow{1}{*}{$\mathcal{J}$ Mean $\uparrow$}          &          & 55.3 &  64.2   &  60.7   &  50.7 & 69.6 & \textbf{78.5}\\
    \cmidrule{1-8}
    \multirow{6}{*}{Success Rate $\uparrow $}       &   $\tau = 0.5$     &  0.602 &  0.712   & 0.679    &  0.588  & 0.764 & \textbf{0.877}\\
                                                    &   $\tau = 0.6$     &  0.523 &  0.670   & 0.624    &  0.534  & 0.723 & \textbf{0.851}\\
                                                    &   $\tau = 0.7$     &  0.413 &  0.599   & 0.536    &  0.457  & 0.681 & \textbf{0.798}\\
                                                    &   $\tau = 0.8$     &  0.267 &  0.492   & 0.434    &  0.337  & 0.568 & \textbf{0.706}\\
                                                    &   $\tau = 0.9$     &  0.088 &  0.246   & 0.239    &  0.167  & 0.386 & \textbf{0.468}\\
                                                    &   $SR_{mean}$      &  0.379 &  0.544   & 0.502    &  0.417  & 0.624 & \textbf{0.740}\\
    \bottomrule
    \end{tabular}
    \end{adjustbox}
  \caption{Results of moving camouflaged object segmentation on MoCA dataset with best overall results in \textbf{bold}. Results shown as mean Intersection over Union (mIoU) and localization success rate for various thresholds, $\tau$. Our results are reported with ResNet101 backbone, as used in the state of the art, as well as Video-Swin backbone, labelled with $\dagger$.
  }
  \label{tab:moca:sota}
\end{table*}

\begin{table*}[h]
  \centering
  \aboverulesep=0ex
   \belowrulesep=0ex
   \begin{adjustbox}{max width=\textwidth}
    \begin{tabular}{@{}cc|ccccccc|ccccc@{}}
    \toprule
    \multicolumn{2}{c|}{\multirow{2}{*}{Measures}} & \multicolumn{7}{c|}{Uses RGB + Optical Flow} & \multicolumn{4}{c}{Uses RGB only} \\
    \multicolumn{2}{c|}{}                                  &   MATNet~\cite{zhou2020motion}  &   RTNet~\cite{ren2021reciprocal}  &  FSNet ~\cite{ji2021full}& HFAN~\cite{pei2022hierarchical} & TMO~\cite{cho2023treating}& PMN~\cite{lee2023unsupervised}  & Isomer~\cite{yuan2023isomer} & COSNet~\cite{lu2019see} &  DFNet~\cite{zhen2020learning}   & IMP~\cite{lee2022iteratively} & \textbf{Ours} & \textbf{Ours$\dagger$}  \\
    \midrule
    \multirow{3}{*}{$\mathcal{J}$}          &    Mean   $\uparrow $    &  -/82.4   &  84.3/85.6   &  82.1/83.4 & 86.8/88.0 & 85.6/-  & 85.4/85.6 & \textbf{88.8}/- & -/80.5&  -/83.4  & 84.5/-   & 83.3/83.7 & 85.0/86.7  \\
                                            &    Recall $\uparrow $    &  -/94.5   &   -/96.1     &  -   & 96.1/96.2 & - & - & - & -/94.0   &   -        &   92.7/-   & 93.6/93.3 & 95.6/ \textbf{96.6} \\
                                            &    Decay  $\downarrow$   &  -/5.5    &   -          &  -    & 4.3/4.5    & -& -&-& -/\textbf{0.0}   &  -  & 2.8/-    & 0.05/0.05 & 0.06 /0.07           \\
    \cmidrule{1-14}
    \multirow{3}{*}{$\mathcal{F}$}          &   Mean    $\uparrow $    &  -/80.7   & 83.0/84.7    &  83.0/83.1 & 88.2/89.3& 86.6/- & 86.4/86.2 & \textbf{91.1}/- & -/79.4      & -/81.8   & 86.7/- & 84.3/83.7 & 87.6/86.9  \\
                                            &   Recall  $\uparrow $    &  -/90.2   & -/93.8       &  - & 95.3/\textbf{95.4} & - &  -  & -      & -/90.4   &  -       & 93.3/- & 93.6/93.7 & 95.1/94.5  \\
                                            &   Decay   $\downarrow$   &  -/4.5    &  -           &  - &  1.1/2/0 & - & - &- & -/\textbf{0.0}   &  -  & 0.8/-  & 0.03/0.03 & 0.04/ 0.04          \\
    \bottomrule
    \end{tabular}
    \end{adjustbox}
  \caption{Results on DAVIS'16 validation set. For those using post processing (\eg, conditional random fields~\cite{akhter2020epo, zhou2020motion, ren2021reciprocal, ji2021full, wang2019learning, lu2019see, wang2019zero,zhen2020learning}, instance pruning~\cite{yang2019anchor, zhen2020learning}, multiscale inference~\cite{zhen2020learning}), results shown as $x/y$, with $x$ and $y$ results without and with post processing, resp. Boundary F-measure, $\mathcal{F}$, and mean Intersection over Union (mIoU), $\mathcal{J}$ are shown. 
  We show our results with ResNet-101 backbone and Video-Swin backbone, indicated by $\dagger$. Best results in \textbf{bold}. 
  }
  \label{tab:davis16:sota}
\end{table*}

\begin{table*}[h]
\centering
  \aboverulesep=0ex
   \belowrulesep=0ex
    \begin{adjustbox}{max width=0.98\textwidth}
	    \begin{tabular}{@{}c|ccccc|cccccc@{}}
		\toprule
		\multirow{2}{*}{Category} & \multicolumn{5}{c|}{Uses RGB+Flow} & \multicolumn{6}{c}{Uses RGB only} \\
		 & MATNet~\cite{zhou2020motion}  &   RTNet~\cite{ren2021reciprocal}  & 
   HFAN~\cite{pei2022hierarchical} & TMO~\cite{cho2023treating}& PMN~\cite{lee2023unsupervised} &
   PDB~\cite{song2018pyramid}   & AGS~\cite{wang2019learning}   &  COSNet~\cite{lu2019see} & AGNN~\cite{wang2019zero}  &   \textbf{Ours}  &   \textbf{Ours$\dagger$}  \\ 
		\midrule
		Airplane(6)  & 72.9  & 84.1   & 84.7          & 85.7 &   -    & 78.0  	    & 87.7 & 81.1  & 81.1          &  \textbf{88.9}   & \textbf{88.9} \\
		Bird(6)      & 77.5  & 80.2   & 80.0          & 80.0 &   -    & 80.0  	    & 76.7 & 75.7  & 75.9          & 73.6             & \textbf{85.4} \\
		Boat(15)     & 66.9  & 70.1   & 72.0          & 70.1 &   -    & 58.9		& 72.2 & 71.3  & 70.7          & 77.4             & \textbf{77.9} \\
		Car(7)       & 79.0  & 79.5   & 76.1          & 78.0 &   -    & 76.5		& 78.6 & 77.6  & 78.1          & 77.7             & \textbf{86.6} \\
		Cat(16)      & 73.7  & 71.8   & 76.0          & 73.6 &   -    & 63.0		& 69.2 & 66.5  & 67.9          & 79.5             & \textbf{83.4} \\
		Cow(20)      & 67.4  & 70.1   & 71.2          & 70.3 &   -    & 64.1		& 64.6 & 69.8  & 69.7          & \textbf{76.1}    &   75.3        \\
		Dog(27)      & 75.9  & 71.3   & 76.9          & 76.8 &   -    & 70.1		& 73.3 & 76.8  & 77.4          & 75.9             & \textbf{81.1} \\
		Horse(14)    & 63.2  & 65.1   & \textbf{71.0} & 66.2 &   -    & 67.6		& 64.4 & 67.4  & 67.3          & 68.6             &   67.7 \\
		Motorbike(10)& 62.6  & 64.6   & 64.3          & 58.6 &   -    & 58.3		& 62.1 & 67.7  & 68.3          & 65.9             & \textbf{69.2} \\
		Train(5)     & 51.0  & 53.3   & 61.4          & 47.0 &   -    & 35.2		& 48.2 & 46.8  & 47.8          & 70.1             & \textbf{77.2}\\
		\midrule
		Mean     	 & 69.0  & 71.0   & 73.4          & 71.5 &   71.8 & 65.4		& 69.7 & 70.5  & 70.8		   & 75.4  &    \textbf{79.3}\\
		\bottomrule
	\end{tabular}
\end{adjustbox}
\caption{Results on YouTube-Objects. Results shown as mean Intersection over Union (mIoU) per category as well as overall average across all categories. $\dagger$ indicates our model with Video-Swin backbone. Best results are highlighted in \textbf{bold}.}
\label{tab:ytbobj:sota}
\end{table*}

\begin{figure*}[h]
	\begin{center}
		\setlength\tabcolsep{0.7pt}
		\def\arraystretch{0.5}
		\resizebox{0.99\textwidth}{!}{
                \begin{tabular}{ccccccc}
                & \multicolumn{2}{c}{\scriptsize{DAVIS'16}} & \multicolumn{2}{c}{\scriptsize{MoCA}} & \multicolumn{2}{c}{\scriptsize{YouTube Objects}}
                \\
               \rotatebox{90}{\hspace{6pt}\scriptsize{Image}}
                &
			\includegraphics[height=0.068\textwidth]{figures/breakdance/img/00060.jpg}
			&
			\includegraphics[height=0.068\textwidth]{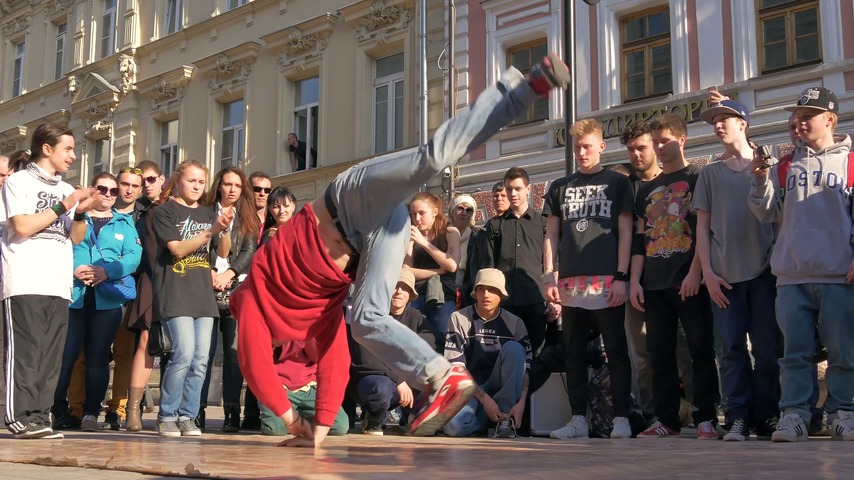}
                &
			\includegraphics[height=0.068\textwidth]{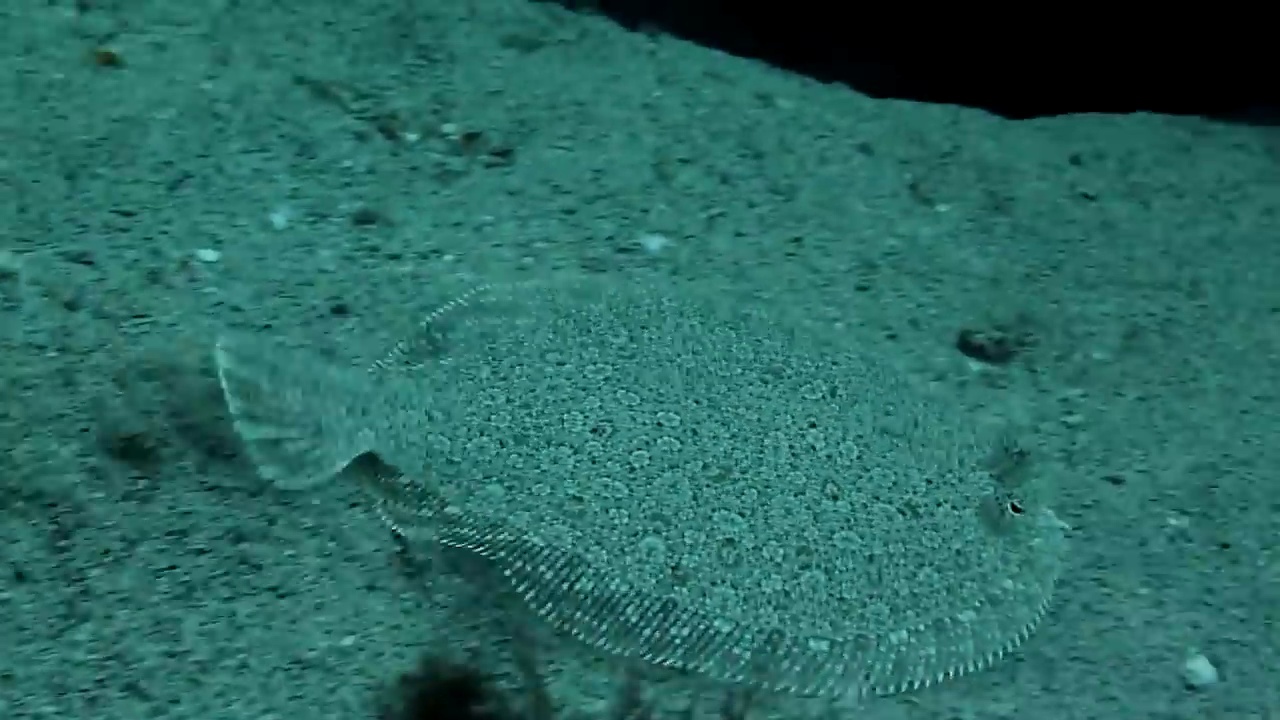}
			&
			\includegraphics[height=0.068\textwidth]{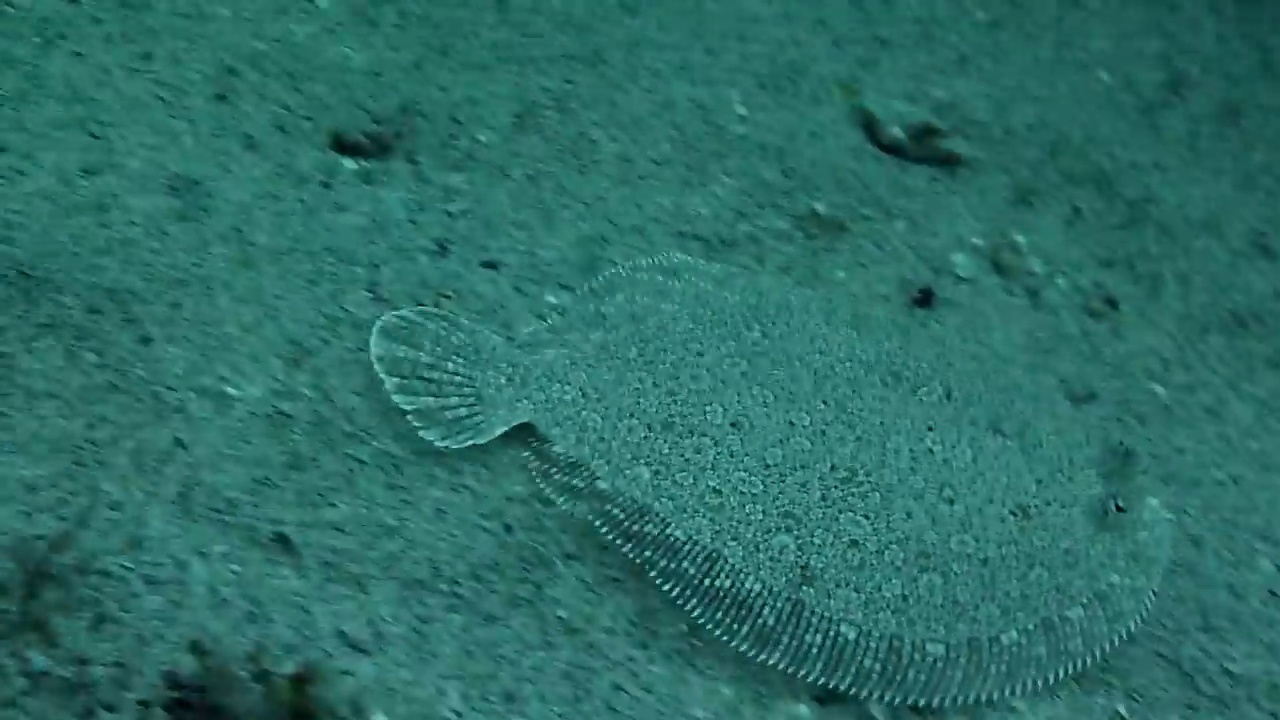}
                &
			\includegraphics[height=0.068\textwidth]{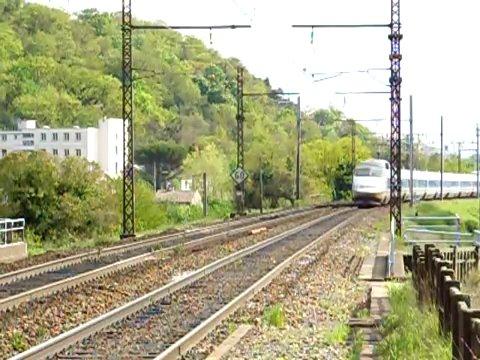}
			&
			\includegraphics[height=0.068\textwidth]{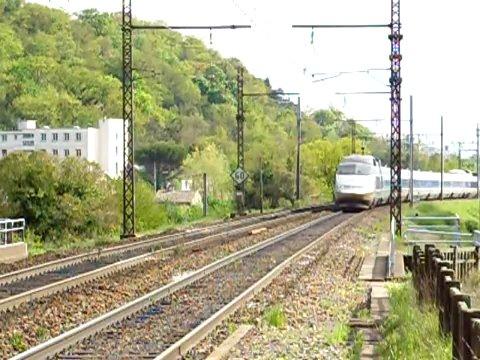}
                \\
                \rotatebox{90}{\hspace{10pt}\scriptsize{GT}}
                &
			\includegraphics[height=0.068\textwidth]{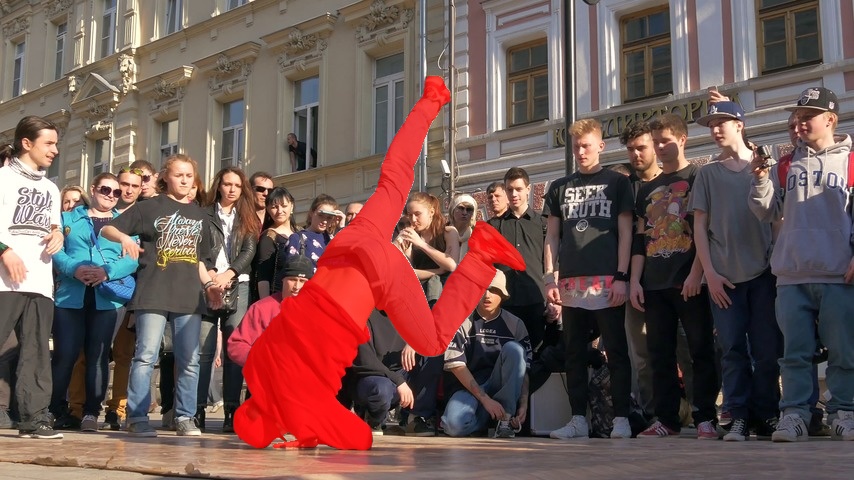}
                &
			\includegraphics[height=0.068\textwidth]{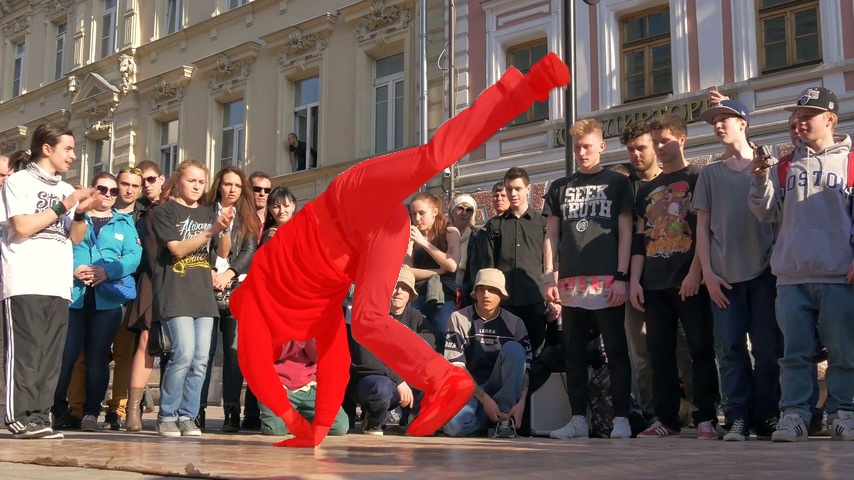}
                &
			\includegraphics[height=0.068\textwidth]{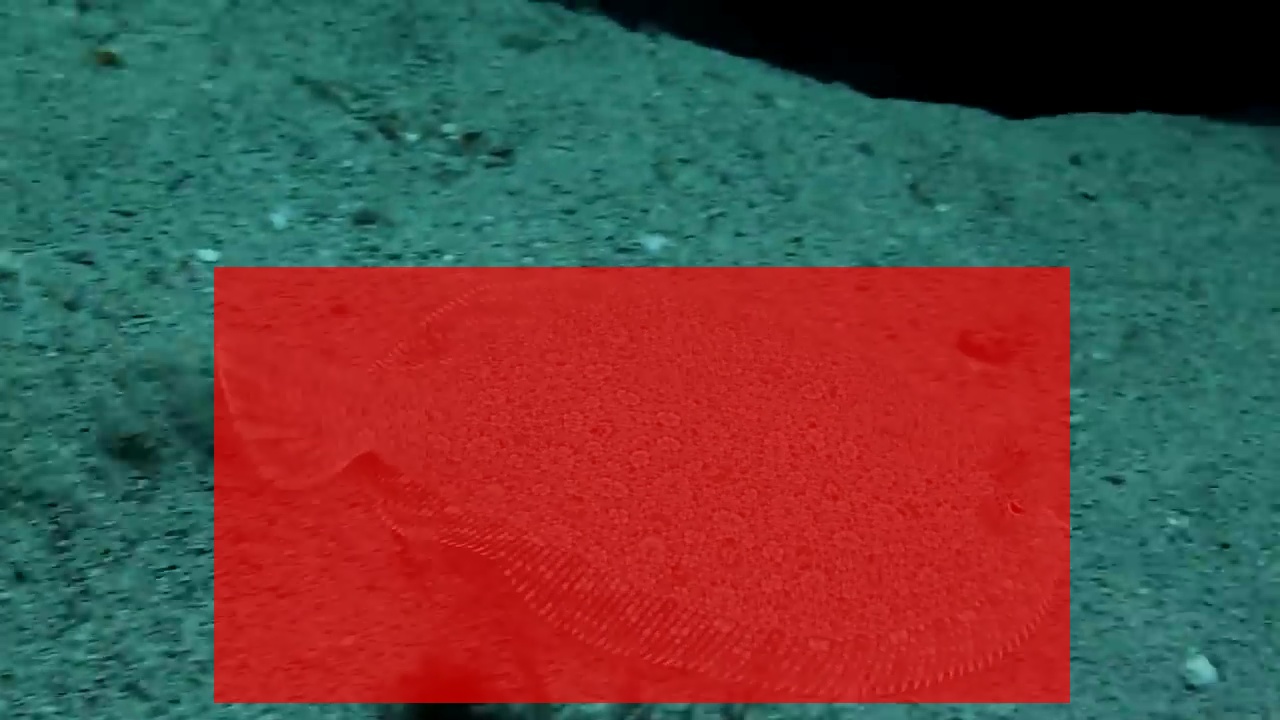}
                &
			\includegraphics[height=0.068\textwidth]{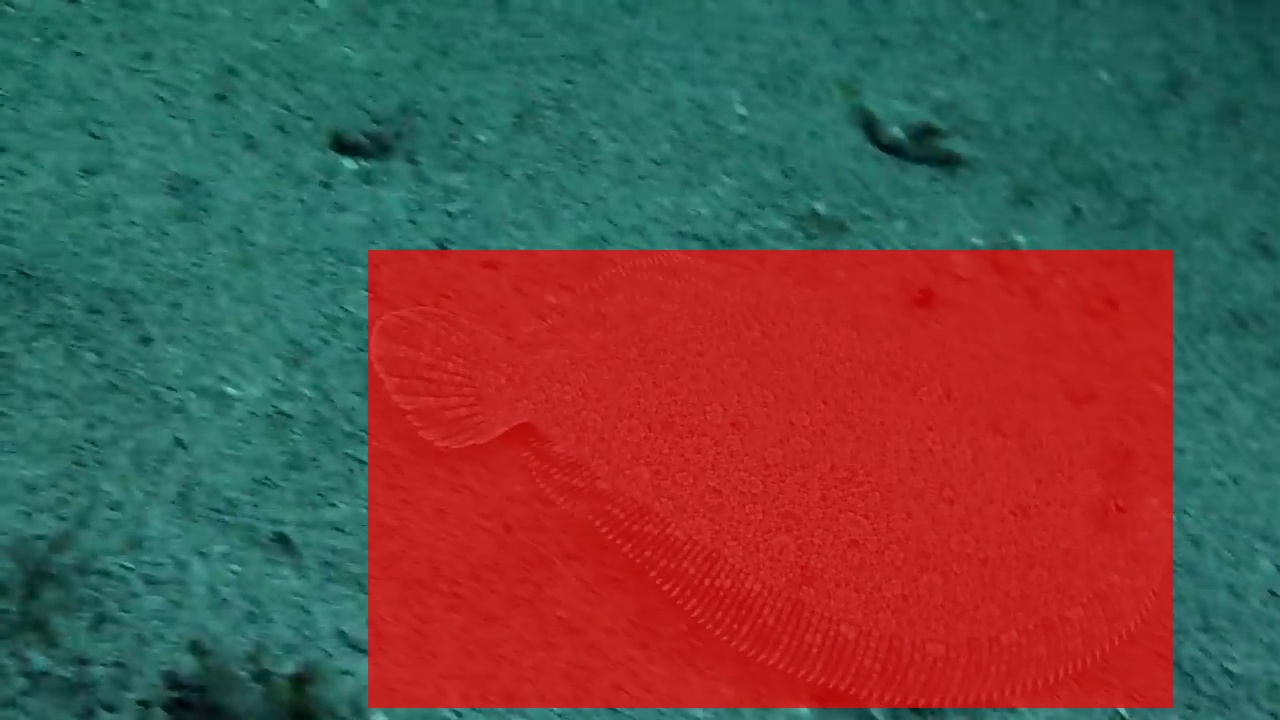}
                &
			\includegraphics[height=0.068\textwidth]{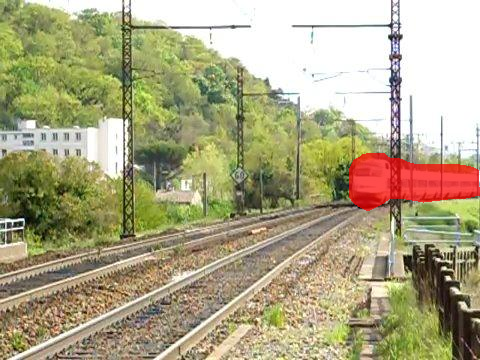}
                &
			\includegraphics[height=0.068\textwidth]{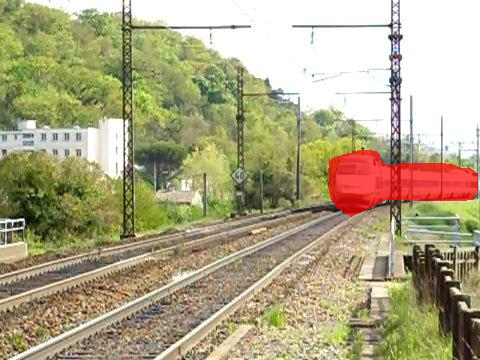}
                \\
                \rotatebox{90}{\hspace{5pt}\scriptsize{Baseline}}
                &
			\includegraphics[height=0.068\textwidth]{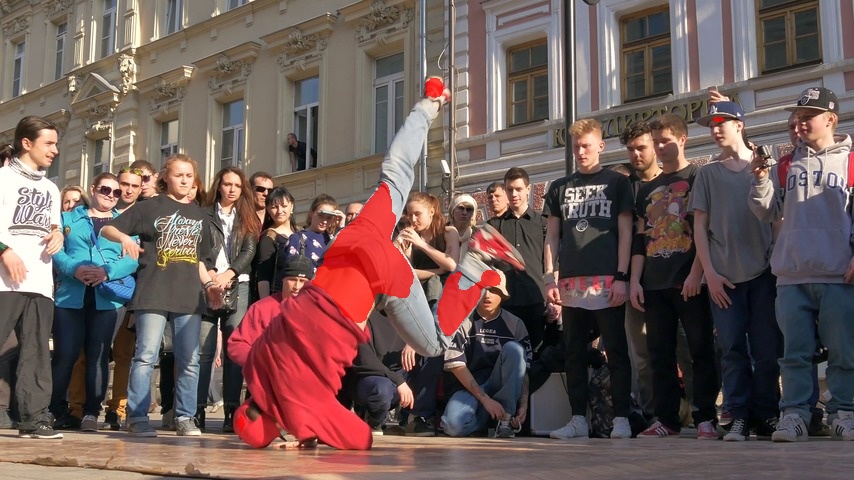}
                &
			\includegraphics[height=0.068\textwidth]{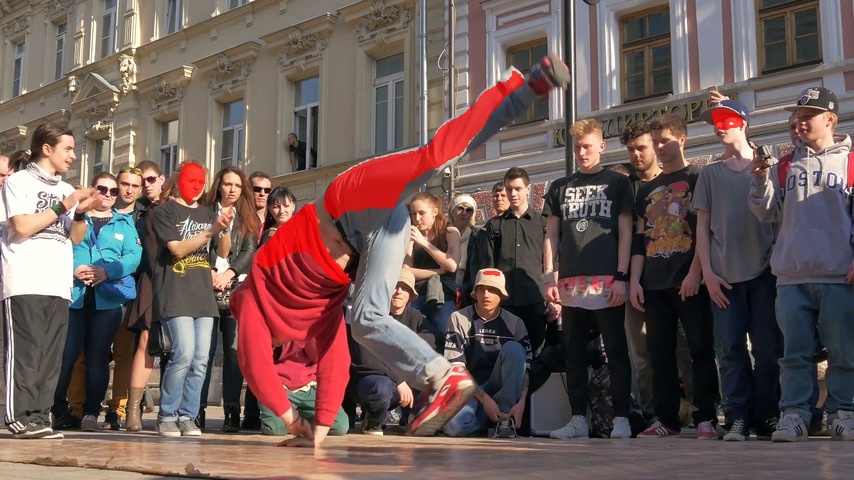}
                &
			\includegraphics[height=0.068\textwidth]{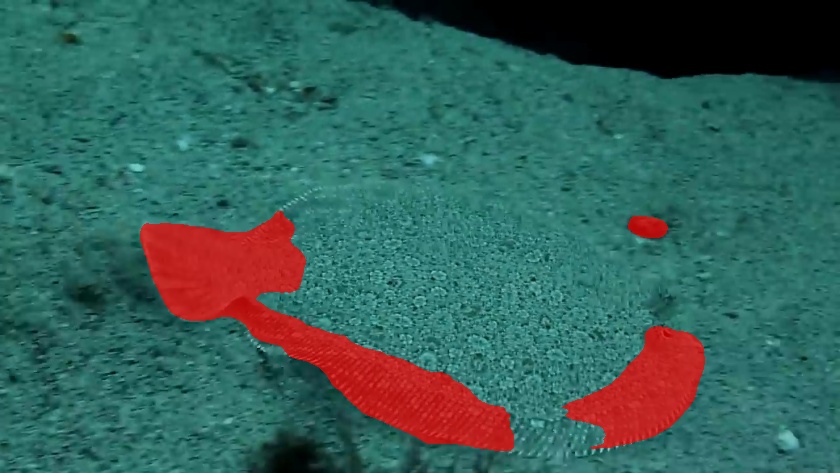}
                &
			\includegraphics[height=0.068\textwidth]{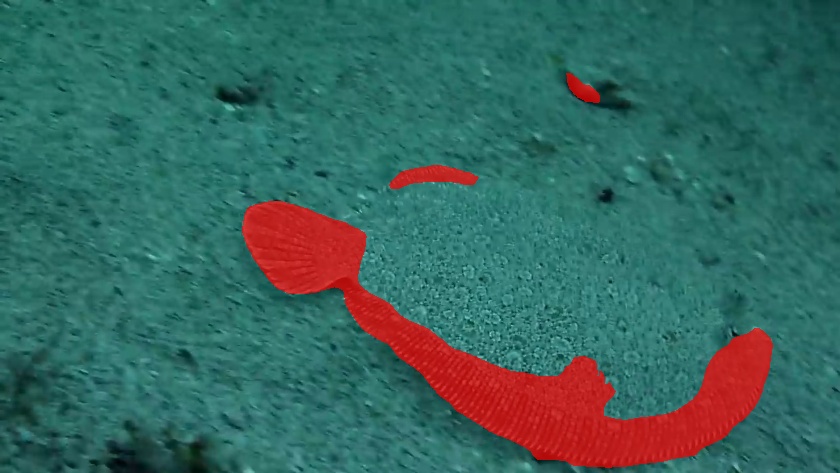}
                &
			\includegraphics[height=0.068\textwidth]{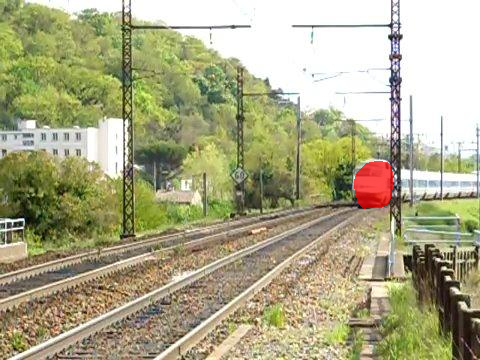}
                &
			\includegraphics[height=0.068\textwidth]{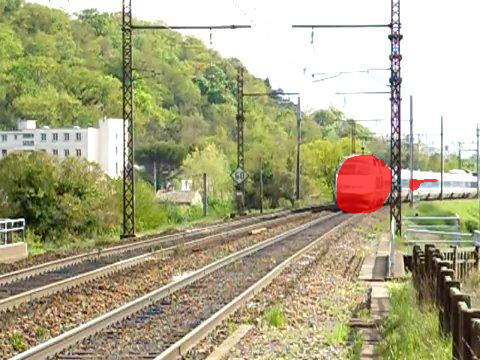}
                \\
                 \rotatebox{90}{\hspace{4pt}\scriptsize{MED-VT}}
                &
			\includegraphics[height=0.068\textwidth]{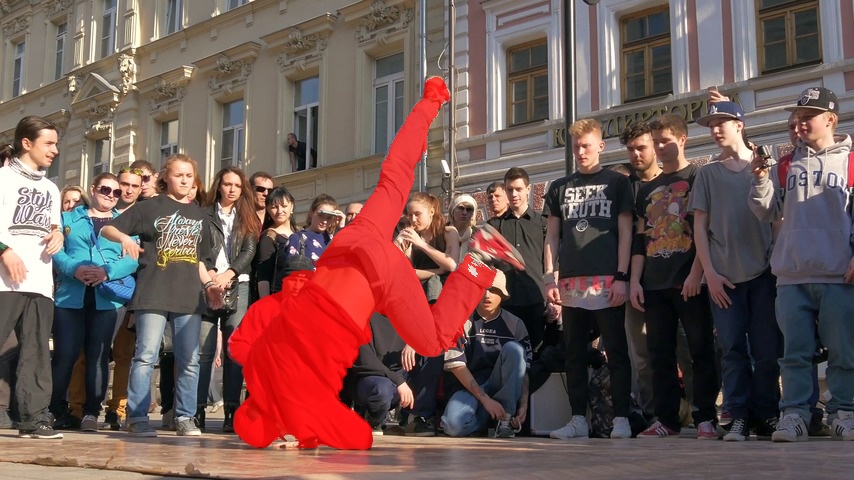}
                &
			\includegraphics[height=0.068\textwidth]{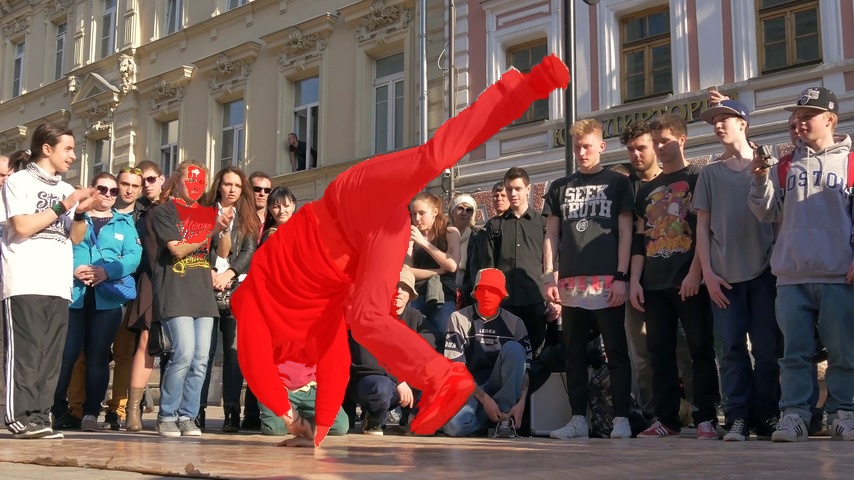}   
                &
			\includegraphics[height=0.068\textwidth]{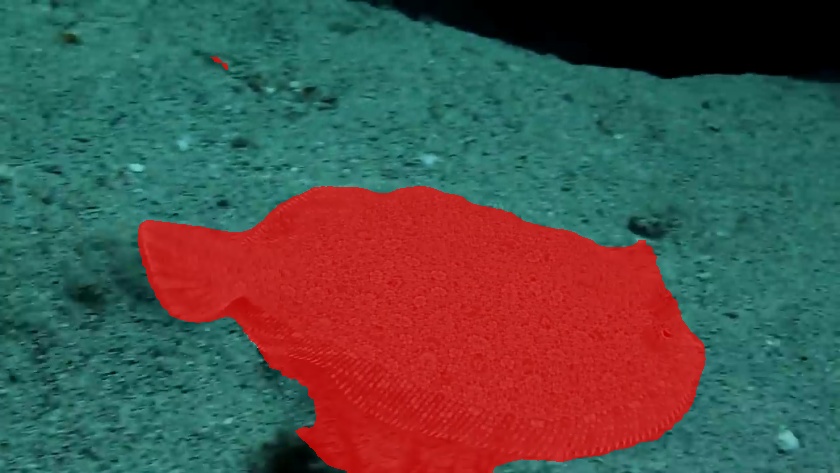}
                &
			\includegraphics[height=0.068\textwidth]{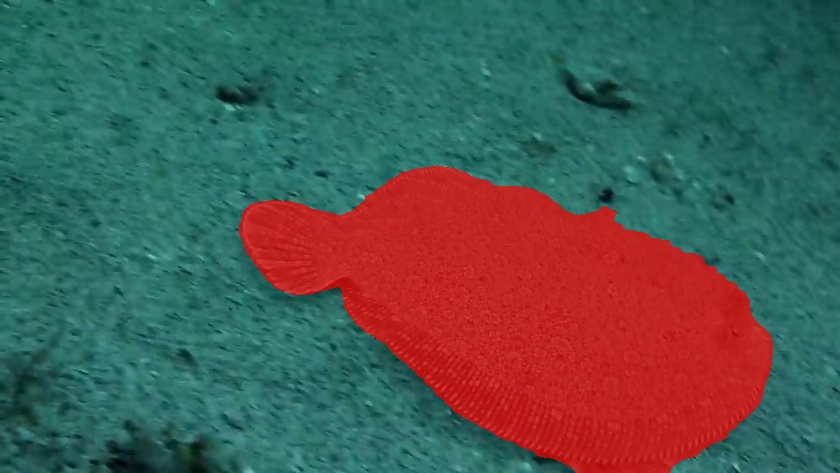}
                &
			\includegraphics[height=0.068\textwidth]{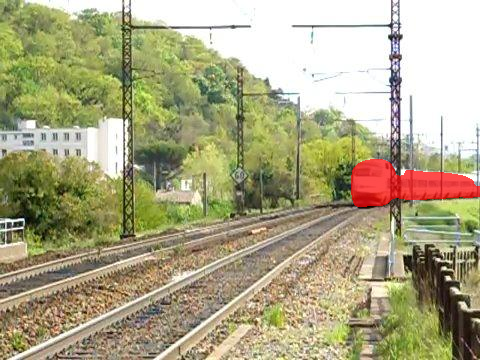}
                &
			\includegraphics[height=0.068\textwidth]{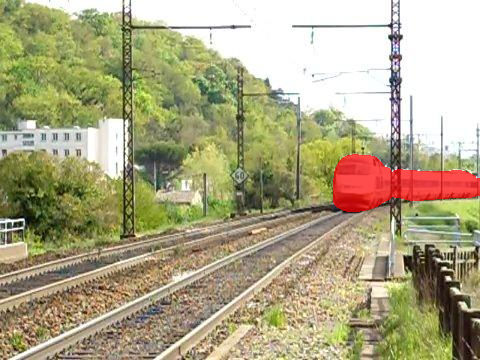}
            \\                
 		\end{tabular}
            }
	   \caption{Qualitative segmentation results (red masks) showing the efficacy of our full model. From top to bottom, rows are arranged as input image, ground truth, our single scale encoder-decoder (\textit{baseline}) and MED-VT. \textbf{Left:} Two frames of DAVIS'16 breakdance. \textbf{Middle:} Two frames of  MoCA Flounder-6. \textbf{Right:} Two frames of  YouTube Objects train shot $0025$. Clearly, the example shows that MED-VT adeptly tackles challenges in complex motion, fine localization, strong camouflage, and partial occlusion in videos.}
	\label{fig:qualitative}
	\end{center}
\end{figure*}

\textbf{Implementation details.} 
For unimodal segmentation, we present results from two backbones that are commonly used on the considered tasks, ResNet-101~\cite{he2016deep} and Video-Swin~\cite{liu2022video}, to extract multiscale feature pyramid, $F$, from a clip of $T=6$ frames by taking features, $\mathsf{f}_s$, as activations at the last block of successive stages. 
For the multimodal segmentation task, we instead use  pretrained PVTv2~\cite{wang2022pvt} and VGGish~\cite{hershey2017cnn} as visual and audio backbone, respectively, because those are used by our comparison approaches. For this task, we use $T=5$ as clip size because the employed AVS dataset only provides five annotated frames per video.

For the multiscale transformer encoder, we use only the two coarsest scale features for memory efficiency reasons; we use six and one encoder blocks for the two smallest scale features. 
In the FPN we combine the two coarsest resolution feature outputs from the encoder with the two finest resolution outputs directly from the backbone. Again, we only use the two coarsest resolutions for memory efficiency. The last three (coarsest) levels from the FPN are input to our multiscale query learning mechanism. For query learning, we use three feature scales and three iterations, $N_d=3$, resulting in nine decoding layers. All multihead attention operations and the attention block, $\mathcal{A}^D$, have, $N_h=8$, attention heads and use a channel of dimension $d=384$. All input backbone features, \ie visual features for all cases and audio features for multimodal case are projected to dimension $d=384$  using a linear projection layer before subsequent processing. The weights for the loss function, $\mathcal{L}$, are $w_m=1.0$ and $w_a=0.5$.

The final segmentation head, $\mathcal{H}$, is a three-layer convolutional module. The final decoding layer of $\mathcal{H}$ is defined according to the segmentation task to match the number of categories, \ie, foreground \textit{vs.} background for AVOS and AVS, 43 for actor-action segmentation and 124 for VSS to encompass the classes in the dataset under consideration.

\textbf{Inference, datasets and evaluation protocols.} 
For inference we use the same clip length as in training, $T=6$, in a sliding window with the predicted logits upsampled to the original image size for AVOS, actor-action and VSS.
Each temporal window of frames serves as input to our model to predict the segmentation of its centre frame. For the DAVIS'16 dataset only, we use multiscale inference postprocessing wherein inference is conducted at multiple scales and subsequently averaged~\cite{chen2017deeplab}, as it is standard with that dataset to present results with and without postprocessing; although, the postprocessing methods vary. 
In case of audio-visual video segmentation, we use $T=5$ and infer the whole sequence at once without using a sliding window.

In addition to comparing our results to extant state-of-the-art approaches, to highlight the particular benefits of multiscale processing, we also present qualitative and ablation results with a single scale baseline architecture. This baseline operates with only single scale
transformer encoder and decoder using just the top
layer feature map of the backbone feature extractor, \ie
the coarsest scale.

For AVOS, we test on three standard datasets: DAVIS'16 ~\cite{perazzi2016benchmark}, YouTube-Objects~\cite{prest2012learning} and MoCA (Moving Camouflaged Animals)~\cite{lamdouar2020betrayed}. DAVIS'16 is a widely adopted AVOS benchmark, while YouTube-Objects is another large-scale VOS dataset. MoCA is the most challenging motion segmentation dataset available, as in the absence of motion the camouflaged animals are almost indistinguishable from the background by appearance alone (\ie colour and texture). For actor-action segmentation, we use the A2D dataset~\cite{xu2015can}, which is standard for that task. For video semantic segmentation, we use the VSPW dataset~\cite{miao2021vspw}, which is a large-scale video semantic segmentation dataset 
with pixel-level annotations for 124 categories. For multimodal audio-visual segmentation, we use the AVSBench dataset\cite{zhou2022audio}, which has two subsets: one with single sound sources 
and the other with mutliple sound sources. 
For all datasets we use its standard evaluation protocol.

\subsection{Comparison to the state of the art}
\label{sec:soa}

\subsubsection{Video object segmentation}
\label{sec:vos_sota}

\textbf{MoCA.} Table~\ref{tab:moca:sota} shows MoCA results, with comparison to the previous state of the art. Since the dataset provides only bounding box annotations, following standard protocol, we compare maximum bounding box of our segmentation mask to compute region similarity. It is evident that MED-VT outperforms all others by a notable margin when using the same backbone (\ie ResNet-101) as the previous state of the art~\cite{zhou2020motion,ren2021reciprocal}. Moreover, MED-VT performance improves even further when using the recent attention-based Video-Swin backbone. Interestingly, even though our model does not use optical flow, it succeeds on this dataset where motion is the primary cue to segmentation due to the camouflaged nature of the animals. This fact supports the claim that our encoder is able to learn rich spatiotemporal features, even without optical flow input.

\textbf{DAVIS'16.}
Table~\ref{tab:davis16:sota} shows DAVIS'16 results, with and without postprocessing. With the Video-Swin backbone, MED-VT outperforms alternatives that do not employ optical flow information like MED-VT in mean/recall F-measure, $\mathcal{F}$, and mean/recall IoU, $\mathcal{J}$. When comparing to  alternatives that use optical flow as additional information, MED-VT still outperforms the majority of approaches, except for HFAN~\cite{pei2022hierarchical} and Isomer~\cite{yuan2023isomer}, where MED-VT is competitive. Again when reverting to the ResNet101 backbone for MED-VT, MED-VT outperforms all similar alternatives that work directly on video frames (\ie, RGB without optical flow) on mean F-measure and mIoU while being competitive with the approaches relying on additional optical flow input. Notably, MED-VT relies only on RGB frames while other state-of-the-art approaches use optical flow as an additional input to exploit object motion. Moreover, most of the DAVIS'16 results include CRF postprocessing~\cite{krahenbuhl2011efficient}. In contrast, we do not employ such complex  postprocessing; rather, we follow a simpler multiscale inference strategy similar to that used by another approach~\cite{zhen2020learning}.

There is evidence that success on DAVIS'16 is largely driven by the ability to capitalize on single frame/static appearance information (\eg, colour, texture), rather than dynamic (\eg, motion) information \cite{kowal2022deeper}. Unlike MED-VT,   RTNet~\cite{ren2021reciprocal}, uses extra pre-training on a saliency segmentation dataset, which aligns with success on DAVIS'16 being tied to single frame information.
Nevertheless, not only can MED-VT be competitive with RTNet (\eg on mean F-measure without postprocessing) when using the same ResNet101 backbone, but by switching MED-VT to the Video-Swin backbone it yields better performance on boundary accuracy and also slightly better performance on mIoU, without an extra dataset or optical flow.



\textbf{YouTube-Objects}. Table~\ref{tab:ytbobj:sota} shows YouTube-Objects results. For overall results averaged across categories, it is seen that our approach once again outperforms all others by a considerable margin when using the standard ResNet-101 backbone without optical flow and further improves using Video-Swin. Also, it is observed that although the performance of most of the alternatives varies widely across different categories (\eg, \textit{Airplane} \vs \textit{Train}), our approach is consistent across categories, which demonstrates its robustness. This observation is especially evident in the most challenging category, ``Train'', where our approach outperforms the previous state-of-the-art approach that uses optical flow input by up to 16\%

Figure~\ref{fig:qualitative} shows qualitative results on three AVOS videos using the segmentation masks overlaid with input images in red. In the figure, the rows are arranged as input image, ground truth, our single scale encoder-decoder (\textit{baseline}), and our model (\textit{MED-VT}). The left two columns show two frames from the breakdance video in DAVIS'16 exhibiting complex, deforming motion and requires fine localization precision to delineate limbs. The middle two columns show the flounder\_6 sequence from MoCA exhibiting strong camouflage. The right two columns show shot 0025 from the train class of YouTube Objects, which exemplifies an object with complex motion and partial occlusion. MED-VT deals with all of these challenges in a temporally consistent fashion, with consistency from the encoder as well as label propagator and localization from the decoder.

\subsubsection{Actor-action segmentation}
\label{sec:aa_sota}

\textbf{A2D.} Table~\ref{tab:a2d} compares our approach on actor-action segmentation to a number of alternatives, which typically use optical flow as an extra input, across different feature backbones. Our results are consistently better than the alternatives, even when we operate under the simplest setting, \ie one input modality (RGB images) and weaker features, \ie ResNet101 \textit{vs.} I3D used by the previous state-of-the-art (SSA2D). When trained and tested with a stronger backbone (yet maintaining only RGB input), the improvements are even more notable, \eg, we outperform the previous best (SSA2D\cite{ssa2d}) by more than 10\% with Video-Swin features~\cite{liu2022video}. 

Figure~\ref{fig:qualitative_a2d} (first two columns) show qualitative segmentation results comparing MED-VT to the baseline algorithm on the A2D dataset. The example shows that MED-VT can delineate segmentation regions precisely with correct actor-action classes compared to the baseline where both segmentation and classification errors are present. Notice, for example, the superior ability of MED-VT to delineate the boundary between the non-moving adult and the climbing baby.

\subsubsection{Video semantic segmentation}
\label{sec:vss_sota}

\textbf{VSPW.} 
Table~\ref{tab:sota_vss} compares our approach on video semantic segmentation to a number of alternatives on the VSPW validation set. Overall, MED-VT without major task specific modification of the model achieved competitive results on VSPW (on-par with the second best result). Notably,  the best reported results, TubeFormer-DeepLab~\cite{kim2022tubeformer}, used additional data augmentation that has not been released, clip-level copy-paste. 

\begin{figure*}[t!]
	\begin{center}
		\setlength\tabcolsep{0.7pt}
    		\def\arraystretch{0.5}
	       	\resizebox{\textwidth}{!}{
                \begin{tabular}{ccccccc}
                    & \multicolumn{2}{c}{\scriptsize{A2D}} & \multicolumn{2}{c}{\scriptsize{VSPW}} & \multicolumn{2}{c}{\scriptsize{AVSBench}}
                    \\
			    \rotatebox{90}{\hspace{6pt}\scriptsize{Image}}
			    &
			    \includegraphics[height=0.068\textwidth,width=0.1207\textwidth]{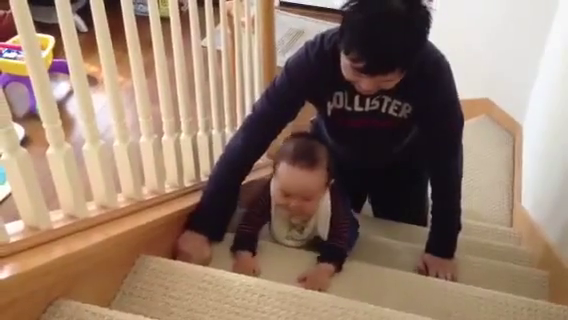}
				&
				\includegraphics[height=0.068\textwidth,width=0.1207\textwidth]{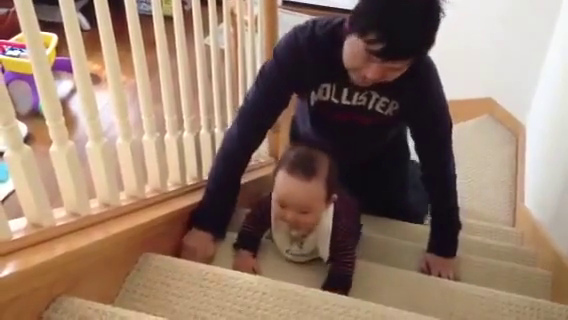}
                    &
				\includegraphics[height=0.068\textwidth,width=0.1209\textwidth]{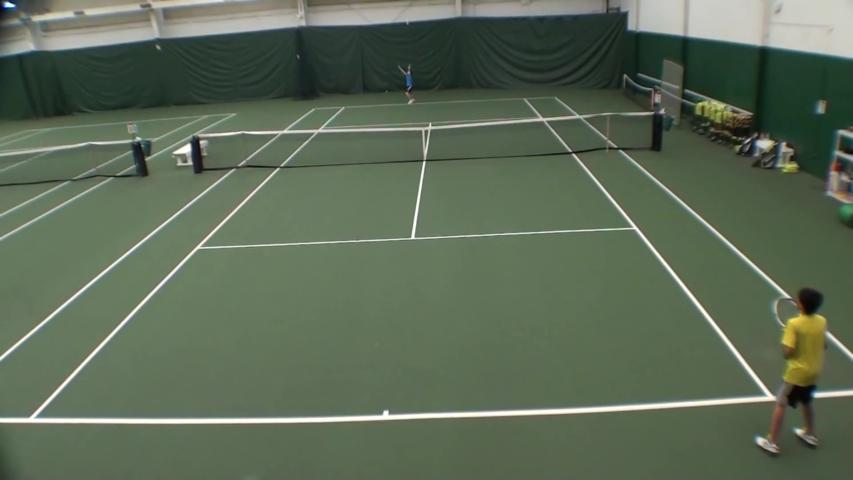}
				&
				\includegraphics[height=0.068\textwidth,width=0.1209\textwidth]{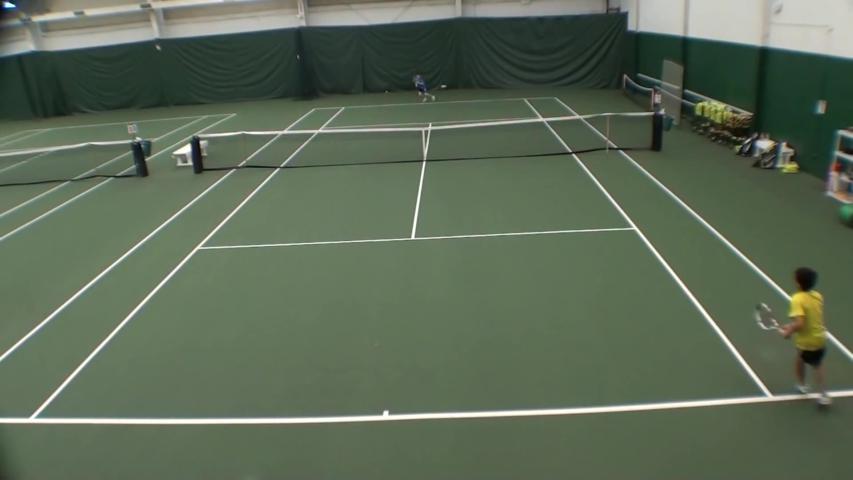}
				&
                    \includegraphics[height=0.068\textwidth,width=0.068\textwidth]{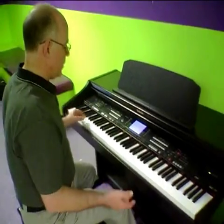}
				&
				\includegraphics[height=0.068\textwidth,width=0.068\textwidth]{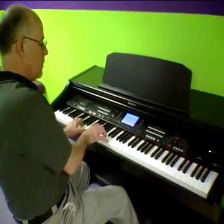}
				\\
				\rotatebox{90}{\hspace{10pt}\scriptsize{GT}}
				&
				\includegraphics[height=0.068\textwidth,width=0.1207\textwidth]{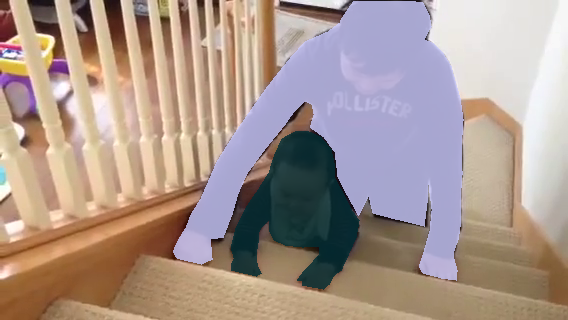}
				&
				\includegraphics[height=0.068\textwidth,width=0.1207\textwidth]{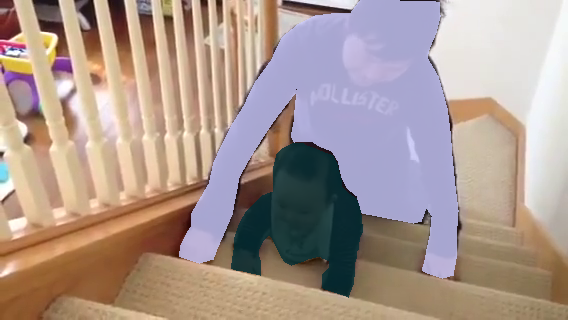}
				&
				\includegraphics[height=0.068\textwidth,width=0.1209\textwidth]{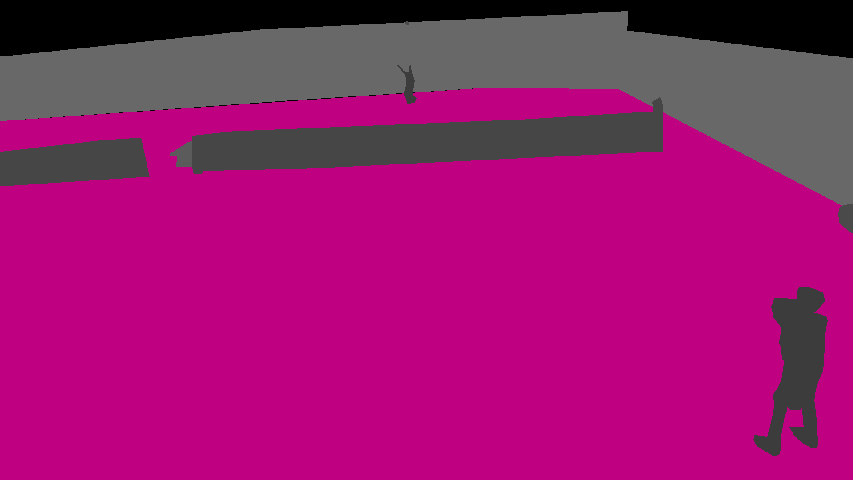}
				&
				\includegraphics[height=0.068\textwidth,width=0.1209\textwidth]{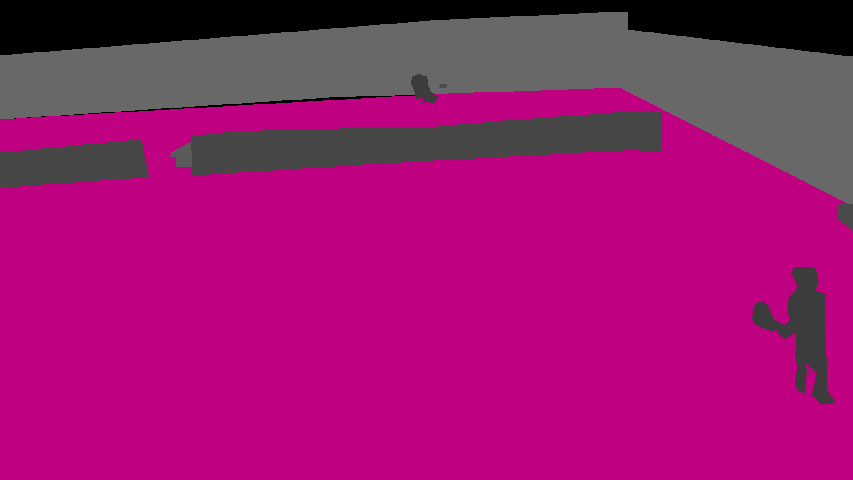}
				&
                    \includegraphics[height=0.068\textwidth,width=0.068\textwidth]{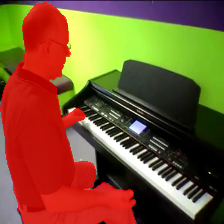}
				&
				\includegraphics[height=0.068\textwidth,width=0.068\textwidth]{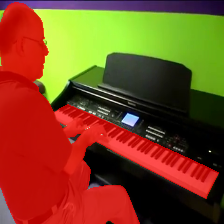}
				\\
				\rotatebox{90}{\hspace{5pt}\scriptsize{Baseline}}
				&
				\includegraphics[height=0.068\textwidth,width=0.1207\textwidth]{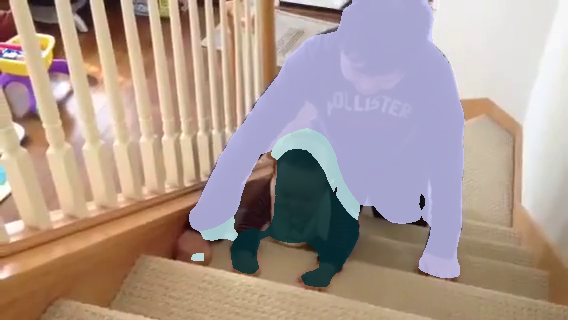}
				&
				\includegraphics[height=0.068\textwidth,width=0.1207\textwidth]{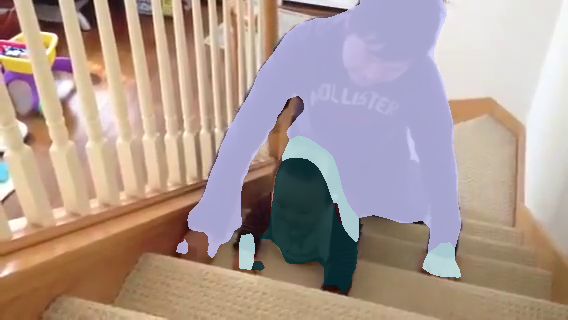}
                    &
				\includegraphics[height=0.068\textwidth,width=0.1209\textwidth]{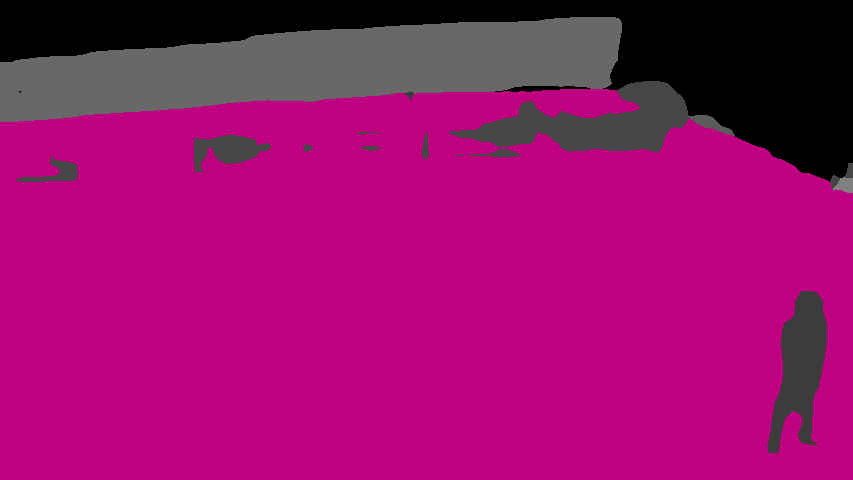}
				&
				\includegraphics[height=0.068\textwidth,width=0.1209\textwidth]{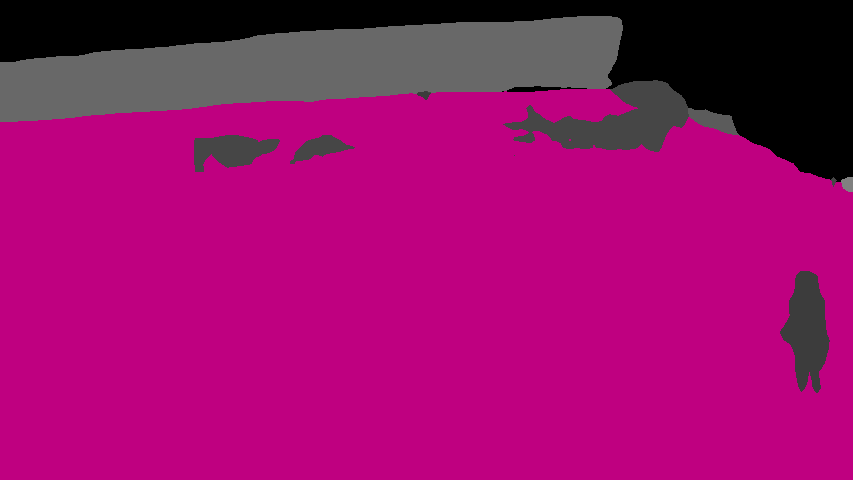}
				&
                    \includegraphics[height=0.068\textwidth,width=0.068\textwidth]{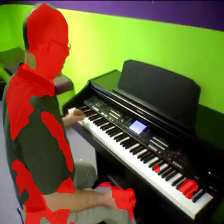}
				&
				\includegraphics[height=0.068\textwidth,width=0.068\textwidth]{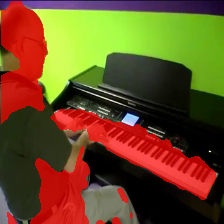}
				\\
				\rotatebox{90}{\hspace{4pt}\scriptsize{MED-VT}}
				&
				\includegraphics[height=0.068\textwidth,width=0.1207\textwidth]{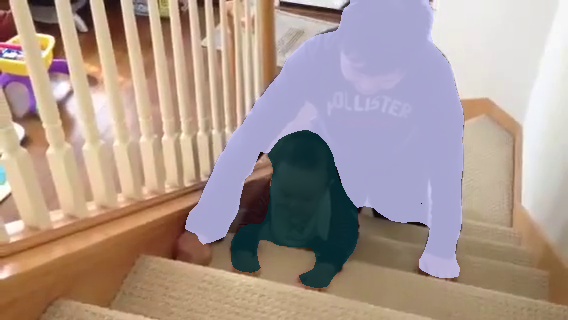}
				&
				\includegraphics[height=0.068\textwidth,width=0.1207\textwidth]{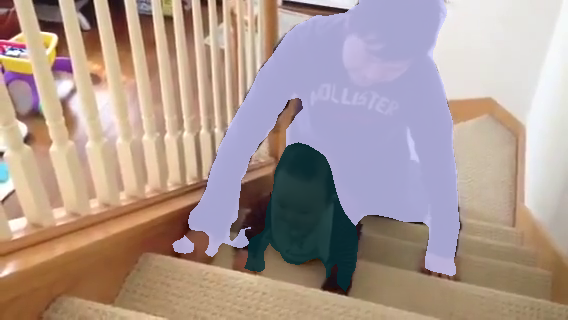}
                    &
				\includegraphics[height=0.068\textwidth,width=0.1209\textwidth]{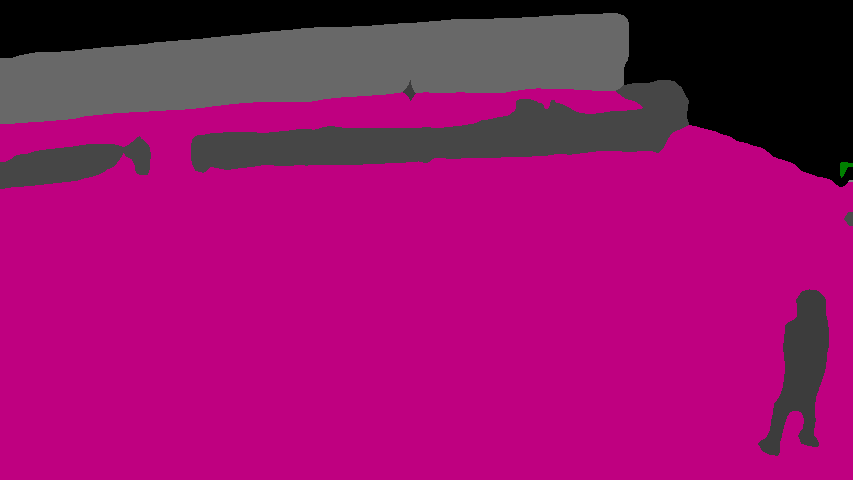}
				&
				\includegraphics[height=0.068\textwidth,width=0.1209\textwidth]{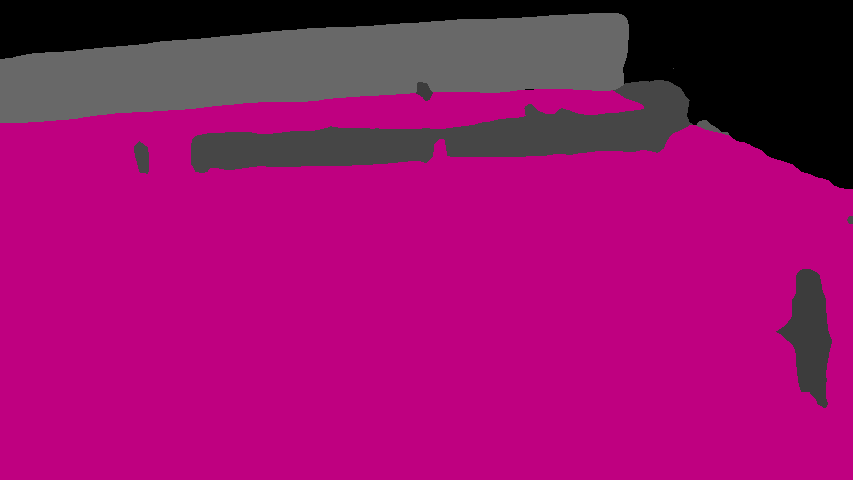}
				&
                    \includegraphics[height=0.068\textwidth,width=0.068\textwidth]{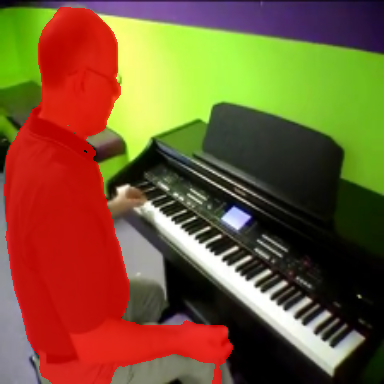}
				&
				\includegraphics[height=0.068\textwidth,width=0.068\textwidth]{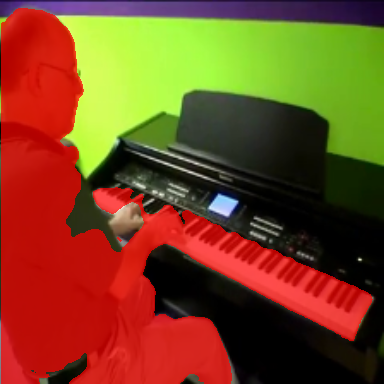}
				\\
                    
		\end{tabular}
        }
	\caption{Qualitative segmentation results comparing MED-VT to baseline algorithm on A2D, VSPW and AVSBench datasets. \textbf{A2D:} Two frames of $2yu9Qkdo4HY$ with $<$adult, none$>$ and $<$baby, climbing$>$ actor-action tuples. \textbf{VSPW:} Two frames of $8aIZCJKQL1s$. \textbf{AVSBench:} Two frames from AVSBench where the audio has sounds for `man talking' as well as `man talking and playing piano'. MED-VT segments with fine precision and classifies correctly in scenarios of multiple objects, multiple actions, multiple sound sources, even in scenarios involving articulated parts and complex object motions.
         }
	\label{fig:qualitative_a2d}
	\end{center}
\end{figure*}

Figure~\ref{fig:qualitative_a2d} (second two columns) show qualitative segmentation results comparing MED-VT to the baseline algorithm on the VSPW dataset. These visual examples show that MED-VT can better delineate semantic categories compared to the baseline (\eg tennis net).  

\begin{table}[t]
   \centering
   \begin{adjustbox}{width=0.4\textwidth}
    \begin{tabular}{l c c c}
              \toprule
               Method & Input & Backbone & mIoU \\
               \midrule
               Ji et al.~\cite{ji2018end} & & ResNet-101 & 36.9 \\
               Dang et al.~\cite{dang2018actor} & RGB + Optical Flow & ResNet-101 & 38.6 \\
            
               SSA2D~\cite{ssa2d} & & I3D & 39.5 \\
                \midrule
                  \multirow{2}{*}{\textbf{MED-VT (Ours)}}& \multirow{2}{*}{RGB} & ResNet-101 & 45.7 \\
                 &  & Video-Swin & \textbf{57.1} \\
               \bottomrule
          \end{tabular}
   \end{adjustbox}
    \caption{State-of-the-art comparisons for actor-action segmentation on A2D dataset; best results in \textbf{bold}. Results given as mean Intersection over Union (mIoU). 
    }
    \label{tab:a2d}
\end{table}
\begin{table}[t]
       \centering
       \begin{adjustbox}{width=0.49\textwidth}
            \begin{tabular}{lccccc}
            \toprule
               Method & Backbone & mIoU & fwIoU & $mVC_8$ & $mVC_{16}$\\ 
              \midrule
             TCB~\cite{miao2021vspw}                     &ResNet101         & 37.8          & 59.5          &86.9           & 82.1 \\
             Video K-Net~\cite{li2022video}              &ResNet101         & 38.0          & -             &87.2           & 82.3 \\   
             CFFM~\cite{sun2022coarse}                   &MiT-B5            & 49.3          & 65.8          &90.8           & 87.1  \\
             MRCFA~\cite{sun2022mining}                  &MiT-B5            & 49.9          & 66.0          &90.9           &87.4 \\    
             TubeFormer-DeepLab~\cite{kim2022tubeformer} &Axial-ResNet50x64 & \textbf{63.2} & -             & \textbf{92.1} & 88.0\\
             Tube-Link~\cite{li2023tube}                 &Swin-L            & 59.7          &  -            & 90.3          & \textbf{88.4}\\
             \cmidrule{1-6}
             \textbf{MED-VT (Ours) }                     &Swin-B            & 58.1          & 72.9 &89.7           & 83.3\\
             \textbf{MED-VT (Ours) }                     &Swin-L            & 59.6          & \textbf{73.1} &90.5           & 86.7\\
            \bottomrule
           \end{tabular}
           \end{adjustbox}
    \caption[State of the art on video semantic segmentation]{State of the art on video semantic segmentation reporting mean intersection over union (mIoU), frequency-weighted intersection over union (fwIoU) and video consistency ($mVC_8$, $mVC_{16}$). Best results in \textbf{bold}.}
    \label{tab:sota_vss}
\end{table}

\subsubsection{Audio-visual segmentation}
\label{sec:avs_sota}

\textbf{AVSBench}. Table~\ref{tab:sota_avsbench} compares our approach on audio-visual segmentation to the most recent alternatives on the AVSBench dataset~\cite{zhou2022audio}. 
MED-VT++ outperforms the prior state of the art on both the single (S4) and multiple (MS3) sound source settings under both mean intersection over union and F1 scores. These results indicate that the multimodal extension of MED-VT, with a simple attention based feature interaction across different modalities, can produce superior results compared to modality specific specialized feature interaction mechanisms~\cite{zhou2022audio}. 

Qualitative visualizations for this task are shown in Figure~\ref{fig:qualitative_a2d} (last two columns) We observe that MED-VT++ accurately segments the sound source object corresponding to `man talking' and `man talking and playing piano' in the video frames. In the baseline model, person is not properly segmented and false positive segmentation of the piano occurs, even when the piano is not playing. In contrast, MED-VT++ successfully follows the sound sources to segment the desired objects.

\begin{table}[t]
     \centering
    \begin{adjustbox}{max width=0.45\textwidth}
    \begin{tabular}{l|c|c|c|c}
          \toprule
          \multirow{2}{*}{Method}& \multicolumn{2}{c|}{S4}& \multicolumn{2}{c}{MS3}\\
          \cmidrule{2-5}
           & $M_J$  & $M_F$ & $M_J$ & $M_F$ \\
           \midrule
          AVS~\cite{zhou2022audio} & 78.74 & 87.9 & 54.00 & 64.5 \\
          DiffusionAVS~\cite{mao2023contrastive} & 81.38 & 90.2 & 58.18 & 70.9 \\ 
          AVSegFormer~\cite{gao2023avsegformer} & 83.06 & 90.5 & 61.33 & 73.0 \\
          \textbf{MED-VT++ (Ours)} & \textbf{83.22} & \textbf{94.4} & \textbf{66.61} & \textbf{75.5} \\
           \bottomrule
      \end{tabular}
    \end{adjustbox}
    \caption[State of the art on audio-visual video segmentation]{State of the art on audio-visual video segmentation reporting mean intersection over union ($M_J$) and F1-score ($M_F$). S4 and MS3 are two different problem formulations representing semi-supervised single sound source segmentation and multiple sound source segmentation. Best results highlighted in \textbf{bold}.}
    \label{tab:sota_avsbench}    
\end{table}
 
\subsection{Ablation study}
The previous section documented the overall strength of MED-VT, with an integral part of that presentation being a confirmation of our first contribution: state-of-the-art performance without extra optical flow input. In this section we conduct ablation experiments on AVOS (DAVIS'16, MoCA, YouTube-Objects), actor-action segmentation (A2D) and VSS (VSPW), to investigate two more of our contributions: 
the multiscale encoder-decoder video transformer and many-to-many label propagation. 


Table~\ref{tab:ab_ms} shows that  the multiscale decoder immediately improves over the baseline for all of the tasks (AVOS, actor-action segmentation and VSS). Addition of the multiscale encoder further improves the results to demonstrate their complementarity and the importance of unified multiscale encoding-decoding. Moreover, addition of many-to-many label propagation to the unified encoder-decoder consistently leads to the best overall performance. This enhancement can be traced to the label propagation yielding more temporally consistent predictions. Finally, we do a direct analysis of the benefits of our many-to-many label propagation \textit{vs.} an alternative many-to-one label propagation technique, where many-to-one only propagates from previous frames to the current. Table~\ref{tab:ab_lprop} shows consistent improvement with many-to-many label propagation over the many-to-one alternative. 
 

\begin{table}[t]
     \centering
     \aboverulesep=0ex
     \belowrulesep=0ex

    \begin{adjustbox}{max width=0.48\textwidth}
    \begin{tabular}{c|c|c|c|c|c|c|c}
        \toprule
            \scriptsize{DMS} & \scriptsize{EMS}  & \scriptsize{LP} & \scriptsize{DAVIS'16}  & \scriptsize{YouTube Objects} &\scriptsize{MoCA} & \scriptsize{A2D} & \scriptsize{VSPW}     \\
        \midrule
            \scriptsize{-} & \scriptsize{-}       &   \scriptsize{-}  &       \scriptsize{79.5} & \scriptsize{73.9}& \scriptsize{67.5}  &  \scriptsize{52.3}      & 56.3  \\
             \scriptsize{\checkmark} & \scriptsize{-}     &  \scriptsize{-} &   \scriptsize{81.5}& \scriptsize{74.2}  &  \scriptsize{67.7}      &  \scriptsize{54.6}      & 56.7 \\
            \scriptsize{\checkmark} & \scriptsize{\checkmark} & \scriptsize{-} & \scriptsize{82.2} &  \scriptsize{74.4}& \scriptsize{69.1}  & \scriptsize{55.7}& 57.6\\
            \scriptsize{\checkmark} & \scriptsize{\checkmark} & \scriptsize{\checkmark} & \textbf{\scriptsize{83.3}} & \textbf{\scriptsize{75.4}}& \textbf{\scriptsize{69.6}}  &\textbf{\scriptsize{57.1}} & \textbf{\scriptsize{58.1}}\\
        \bottomrule
      \end{tabular}
  \end{adjustbox}
     \caption{Multiscale encoder-decoder and label propagation ablations reporting mIoU. DMS, EMS and LP refer to Decoder Multiscale, Encoder Multiscale and Label Propagation respectively. Best results highlighted in \textbf{bold}. }
     \label{tab:ab_ms}
\end{table}

\begin{table}[t]
     \centering
    \begin{adjustbox}{max width=0.33\textwidth}
    \begin{tabular}{l|c|c|c}
          \toprule
           Method & DAVIS'16  & YouTube Objects &MoCA \\
           \midrule
           - & 82.2 & 74.4 & 69.1 \\
           Mto1 & 81.7 & 74.5 & 68.6  \\
           MtoM & \textbf{83.3} & \textbf{75.4} & \textbf{69.6} \\
           \bottomrule
      \end{tabular}
    \end{adjustbox}
    \caption{Ablation on Many-to-One (Mto1) \textit{vs.} our Many-to-Many (MtoM) label propagation. Best results in \textbf{bold}.}
    \label{tab:ab_lprop}
\end{table}

\begin{table}[t]
     \centering
    \begin{adjustbox}{max width=0.45\textwidth}
    \begin{tabular}{l|c|c|c}
          \toprule
          \multirow{2}{*}{Method}& \multicolumn{3}{c}{Sampling rate}\\
          \cmidrule{2-4}
           & 1  & 5 & 10 \\
           \midrule
           -            & \textbf{78.5} & \textbf{78.0} & \textbf{77.6} \\
          Shuffle frame order in clip & 77.1 & 77.4 & 77.3  \\
          Repeat a single frame in the clip & 70.4 & 72.1 & 72.1 \\
           \bottomrule
      \end{tabular}
    \end{adjustbox}
    \caption{Analysis of temporal dynamics by means of dynamic perturbations on clips using MoCA dataset. Best results highlighted in \textbf{bold}.}
    \label{tab:dpiou}
\end{table}

\input{tkz_figures/time_miou.tex}

\begin{table}[t]
     \centering
    \begin{adjustbox}{max width=0.33\textwidth}
    \begin{tabular}{l|c|c|c}
      \toprule
       Model &  Time & mIoU & Memory\\
       \midrule
       MATNet [54] &  0.9 & 64.2 & \textbf{2577} \\
       RTNet [32]  &  1.9 & 60.7 & 3615 \\
       COSNet [23] & 1.3 & 50.7 & 9255\\
       \textbf{MED-VT (ours)}  & \textbf{0.6} & \textbf{69.6} & 9509 \\
       \bottomrule
    \end{tabular}
    \end{adjustbox}
    \caption{Comparison on run time and memory consumption. Best results highlighted in 
    \textbf{bold}.}
    \label{tab:ab_inference_time}
\end{table}


\subsection{Temporal properties}\label{sec:ab:temporal_dynamics}
\textbf{Dynamic perturbation.}
To document the overall ability of MED-VT to exploit the temporal dimension, without computing optical flow, we consider the impact of corrupting the temporal information in a video. For this experiment we use the MoCA dataset due to its camouflage nature highlighting temporal information. 
We use two variants of temporal perturbation: shuffle the frame order and replicating a single frame  $T$ times to constitute a clip with no temporal dynamics. We perform these experiments with variable temporal sampling rates. Results are shown in Table~\ref{tab:dpiou}. It is seen that shuffling reduces performance for all sampling rates, which demonstrates the importance of frame order. Notably, a far greater performance drop is observed when the temporal information is fully diminished by repeating the same frame. This evidence documents that MED-VT can exploit both correlated information from multiple frames and temporal order in segmenting objects. Conventional video segmentation approaches that simply extend single image segmentation models to process individual frames fail to capture this temporal information, while approaches relying on optical flow limits the ability to consider temporal information to only consecutive frames. Conversely, MED-VT exploits temporal information via clip-based inference with multiscale processing throughout.

\textbf{Temporal consistency.} 
 As a demonstration of temporal consistency, we provide a quantitative result showing IoU over time (\ie across frames) for a sequence from DAVIS'16 and a sequence from MoCA comparing our single scale baseline and MED-VT using ResNet101 backbone. The result is plotted in Figure~\ref{fig:time_iou_analysis}. It is seen that our results are considerably smoother than the baseline, \eg showing less zig-zag pattern, and thereby documenting better temporal consistency in the segmentation.

\subsection{Training and inference efficiency}
We perform all benchmarking experiments on a Linux Server equipped with Intel(R) Xeon(R) W-2155  3.30GHz CPU and an NVIDIA Quadro P6000 24 GB GPU. We run all experiments on a single GPU. The training time is approximately two days. We run other models using their publicly available code and trained models. Table~\ref{tab:ab_inference_time} shows a comparison of inference time, memory use and mIoU score on the MoCA dataset. It is seen that our approach is indeed efficient as it achieves the best performance while being faster in run time. This efficiency is a result of our approach not depending on time consuming optical flow estimation and the parallelization enabled by our ability to process a video clip as a whole rather than sequentially frame-by-frame. Due to the use of a multiscale transformer encoder, our approach requires more GPU memory than the other approaches. An interesting future direction is to investigate memory efficient video transformers, \eg~\cite{Kitaev2020Reformer}.

%% file: tkz_figures/time_miou.tex
\begin{figure}[t]
    \centering
    \resizebox{0.25\textwidth}{!}{
\begin{tikzpicture}
\begin{axis} [
     title={\textbf{Dance-Twirl}},
     title style={at={(axis description cs:0.4,0.02)}, font=\footnotesize
},
     xlabel={Frame},
     ylabel={IoU},
     xmin=0.0, xmax=90.0,
     ymin=0.0, ymax=1.0,
     xtick={0, 10, 20, 30, 40, 50, 60, 70, 80, 90},
     ytick={0.0, 0.2, 0.4, 0.6, 0.8, 1.0},
     x tick label style={font=\tiny},
     y tick label style={font=\tiny},
     x label style={at={(axis description cs:0.5,0.07)},anchor=north,font=\tiny},
     y label style={at={(axis description cs:0.17,.5)},anchor=south,font=\tiny},
     width=6.5cm,
     height=5cm,
     ymajorgrids=false,
     xmajorgrids=false,
      line legend,
     legend style={
        nodes={scale=0.86, transform shape},
        cells={anchor=west},
        legend style={at={(0.822,0.1)},anchor=south,row sep=0.01pt}, font =\footnotesize} ,
     legend image post style={scale=0.86},
] 

\addplot[line width=0.5pt, color=blue, style=solid] coordinates {(0, 0.855314469) (1, 0.678863151) (2, 0.746896436) (3, 0.572181106) (4, 0.596992163) (5, 0.597461804) (6, 0.633387954) (7, 0.639642815) (8, 0.678226309) (9, 0.747622699) (10, 0.783708635) (11, 0.655493896) (12, 0.523995782) (13, 0.639091807) (14, 0.449387742) (15, 0.445892929) (16, 0.589588452) (17, 0.756793643) (18, 0.794712773) (19, 0.816009766) (20, 0.840733479) (21, 0.860644396) (22, 0.890798661) (23, 0.881727952) (24, 0.896047738) (25, 0.865849535) (26, 0.864370748) (27, 0.877011494) (28, 0.858130735) (29, 0.826181755) (30, 0.787583588) (31, 0.801341287) (32, 0.710527263) (33, 0.668362469) (34, 0.674691215) (35, 0.653207098) (36, 0.6746416) (37, 0.642994888) (38, 0.716218049) (39, 0.701006147) (40, 0.671089438) (41, 0.598695711) (42, 0.687085199) (43, 0.741123205) (44, 0.742311197) (45, 0.727480335) (46, 0.685514075) (47, 0.714145171) (48, 0.728478995) (49, 0.796189632) (50, 0.818484476) (51, 0.860462236) (52, 0.864456006) (53, 0.876077559) (54, 0.844049698) (55, 0.855597342) (56, 0.834437899) (57, 0.804113124) (58, 0.851445367) (59, 0.835342953) (60, 0.886070057) (61, 0.789168454) (62, 0.789470526) (63, 0.802004077) (64, 0.832493354) (65, 0.814496223) (66, 0.812924923) (67, 0.82284264) (68, 0.748206478) (69, 0.769687097) (70, 0.839532944) (71, 0.902736988) (72, 0.917974817) (73, 0.938411437) (74, 0.909722066) (75, 0.908824296) (76, 0.899210222) (77, 0.907398865) (78, 0.873641835) (79, 0.884595793) (80, 0.91708817) (81, 0.909463934) (82, 0.918373972) (83, 0.918387188) (84, 0.903817484) (85, 0.908694058) (86, 0.889640821) (87, 0.89471008) (88, 0.904300982) (89, 0.886128169)};

\addplot[line width=0.5pt, color=red, style=solid] coordinates {(0, 0.889082361) (1, 0.895771003) (2, 0.917288505) (3, 0.834525624) (4, 0.826224079) (5, 0.77656817) (6, 0.762726685) (7, 0.743445207) (8, 0.775594185) (9, 0.768876611) (10, 0.788124367) (11, 0.835798979) (12, 0.841400549) (13, 0.850699947) (14, 0.761459438) (15, 0.63769936) (16, 0.868364277) (17, 0.911446949) (18, 0.922896171) (19, 0.906903257) (20, 0.926522954) (21, 0.896960413) (22, 0.896684465) (23, 0.889874029) (24, 0.889901059) (25, 0.868953733) (26, 0.830601093) (27, 0.864027539) (28, 0.863644382) (29, 0.899467033) (30, 0.905548756) (31, 0.907869482) (32, 0.903717419) (33, 0.862539557) (34, 0.869586728) (35, 0.85928508) (36, 0.849318641) (37, 0.808561313) (38, 0.837246347) (39, 0.825108716) (40, 0.841456389) (41, 0.845081155) (42, 0.848390132) (43, 0.849961822) (44, 0.804718218) (45, 0.796345409) (46, 0.793194863) (47, 0.827352135) (48, 0.837317023) (49, 0.875838667) (50, 0.874365088) (51, 0.895624029) (52, 0.868575681) (53, 0.884914025) (54, 0.841245455) (55, 0.832335023) (56, 0.81852723) (57, 0.855164485) (58, 0.907511649) (59, 0.90768629) (60, 0.9055728) (61, 0.859067637) (62, 0.871888879) (63, 0.88667657) (64, 0.869028771) (65, 0.876299199) (66, 0.885487173) (67, 0.882801448) (68, 0.899026104) (69, 0.916713004) (70, 0.912643178) (71, 0.911195428) (72, 0.914220743) (73, 0.945063694) (74, 0.948994011) (75, 0.93706354) (76, 0.933730256) (77, 0.947336888) (78, 0.949396161) (79, 0.952394308) (80, 0.946153006) (81, 0.947337161) (82, 0.94831585) (83, 0.937318118) (84, 0.916201309) (85, 0.913426854) (86, 0.91074877) (87, 0.904292691) (88, 0.920649376) (89, 0.903345933)};

\legend{ \textbf{Baseline}, \textbf{MEDVT}}

\end{axis}
\end{tikzpicture}
}%
\resizebox{0.25\textwidth}{!}{
\begin{tikzpicture}
\begin{axis} [
     title={\textbf{Flounder}},
     title style={at={(axis description cs:0.40,0.02)},font=\footnotesize},
     xlabel={Frame},
     ylabel={IoU},
     xmin=0.0, xmax=90.0,
     ymin=0.0, ymax=1.0,
     xtick={0, 10, 20, 30, 40, 50, 60, 70, 80, 90},
     ytick={0.0, 0.2, 0.4, 0.6, 0.8, 1.0},
     x tick label style={font=\tiny},
     y tick label style={font=\tiny},
     x label style={at={(axis description cs:0.5,0.07)},anchor=north,font=\tiny},
     y label style={at={(axis description cs:0.17,.5)},anchor=south,font=\tiny},
     width=6.5cm,
     height=5cm,
     ymajorgrids=false,
     xmajorgrids=false,
      line legend,
     legend style={
        nodes={scale=0.86, transform shape},
        cells={anchor=west},
        legend style={at={(0.822,0.1)},anchor=south,row sep=0.02pt}, font =\footnotesize
    },
     legend image post style={scale=0.86},
] 

\addplot[line width=0.5pt, color=blue, style=solid] coordinates {(0, 0.538128327) (1, 0.5474301) (2, 0.750334415) (3, 0.771076764) (4, 0.519976866) (5, 0.710891282) (6, 0.35299771) (7, 0.483496797) (8, 0.775940491) (9, 0.727888231) (10, 0.691085182) (11, 0.736645164) (
12, 0.372499294) (13, 0.847432705) (14, 0.772912142) (15, 0.576610682) (16, 0.361497445) (17, 0.10839733) (18, 0.580395574) (19, 0.878296267) (20, 0.854369506) (21, 0.813055606) (22, 0.75714168) (23, 0.759269041) (24, 0.711987791) (25, 0.273972089) (26, 0.768147406) (27, 0.708416729) (28, 0.7500634) (29, 0.688892176) (30, 0.690570514) (31, 0.720894665) (32, 0.77564539) (33, 0.712164105) (34, 0.967714853) (35, 0.799478557) (36, 0.955241914) (37, 0.751685127) (38, 0.675733629) (39, 0.686893918) (40, 0.617366071) (41, 0.702840959) (42, 0.787137288) (43, 0.731648631) (44, 0.824182299) (45, 0.829109285) (46, 0.740811851) (47, 0.718994673) (48, 0.756185803) (49, 0.779986428) (50, 0.617917399) (51, 0.19402564) (52, 0.226181935) (53, 0) (54, 0) (55, 0.064799657) (56, 0.443729747) (57, 0.797272448) (58, 0.905791272) (59, 0.932751968) (60, 0.881485249) (61, 0.626422218) (62, 0.635662399) (63, 0.785696631) (64, 0.756806064) (65, 0.729552786) (66, 0.84692603) (67, 0.949678612) (68, 0.959418184) (69, 0.901165988) (70, 0.947257647) (71, 0.886353232) (72, 0.951957918) (73, 0.845586439) (74, 0.904481183) (75, 0.945066426) (76, 0.934469158) (77, 0.845512163) (78, 0.889776341) (79, 0.581331485) (80, 0.904757833) (81, 0.632357889) (82, 0.945438834) (83, 0.674334045) (84, 0.76407265) (85, 0.71386067) (86, 0.802720961) (87, 0.536512401) (88, 0.448414799) (89, 0.394838336) (90, 0.418305079) (91, 0.102383222) (92, 0.42369314) (93, 0.523477172) (94, 0.550727762) (95, 0.7472598) (96, 0.839793288) (97, 0.226568646) (98, 0.259055859) (99, 0.1045868)};

\addplot[line width=0.5pt, color=red, style=solid] coordinates {(0, 0.414026364) (1, 0.609688574) (2, 0.762597286) (3, 0.688026082) (4, 0.774807064) (5, 0.835751532) (6, 0.706684436) (7, 0.682504692) (8, 0.765009546) (9, 0.773784386) (10, 0.733526551) (11, 0.764286235) (12, 0.974194505) (13, 0.96413067) (14, 0.918244827) (15, 0.912263301) (16, 0.825550264) (17, 0.77958476) (18, 0.893296979) (19, 0.951649748) (20, 0.954749148) (21, 0.85011585) (22, 0.785631022) (23, 0.768430753) (24, 0.696014478) (25, 0.886740558) (26, 0.831685755) (27, 0.954632507) (28, 0.973767715) (29, 0.880308281) (30, 0.910852292) (31, 0.917058465) (32, 0.952863179) (33, 0.88141958) (34, 0.961870377) (35, 0.959338751) (36, 0.96434712) (37, 0.955606756) (38, 0.895694597) (39, 0.824312928) (40, 0.772224426) (41, 0.848240869) (42, 0.87778578) (43, 0.942202445) (44, 0.912628656) (45, 0.909180945) (46, 0.900476836) (47, 0.848936302) (48, 0.851584832) (49, 0.812656436) (50, 0.67229257) (51, 0.32837866) (52, 0.103500175) (53, 0) (54, 0) (55, 0.071364209) (56, 0.069416382) (57, 0.200049622) (58, 0.95511282) (59, 0.946741858) (60, 0.939345252) (61, 0.962550384) (62, 0.875558545) (63, 0.8474716) (64, 0.920496055) (65, 0.893348572) (66, 0.902205593) (67, 0.953218638) (68, 0.972019997) (69, 0.909760857) (70, 0.948860922) (71, 0.956849098) (72, 0.948318682) (73, 0.943737598) (74, 0.923316662) (75, 0.966589177) (76, 0.941494272) (77, 0.910145915) (78, 0.944799849) (79, 0.889954313) (80, 0.904781029) (81, 0.879855729) (82, 0.9636532) (83, 0.763949355) (84, 0.812294428) (85, 0.779492878) (86, 0.858601096) (87, 0.903317313) (88, 0.710532245) (89, 0.790085115) (90, 0.847183734) (91, 0.833722456) (92, 0.93300491) (93, 0.884780688) (94, 0.907363839) (95, 0.961135025) (96, 0.908828864) (97, 0.895530123) (98, 0.281758852) (99, 0.243263604)};

\legend{ \textbf{Baseline}, \textbf{MEDVT}}

\end{axis}
\end{tikzpicture}
}
\caption{Temporal consistency analysis using IoU over time on dance-twirl video of DAVIS'16 and flounder video of MoCA. Flounder IoU drops around frame 53 due to sudden large camera motion.}
\label{fig:time_iou_analysis}
\end{figure}
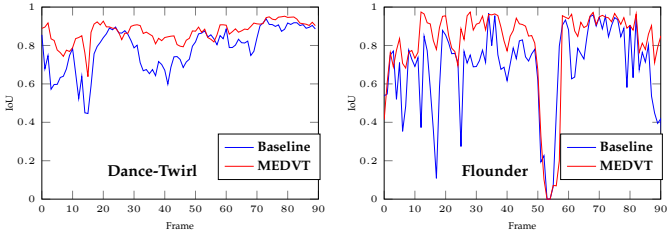

%% file: medvt/6_interpret.tex
\section{Interpretability}\label{sec:interpret}


To gain insight into the learned representations of MED-VT, we provide systematic probes to the architecture at key processing stages: backbone feature extractor, multiscale encoder, multiscale decoder and label propagator. At each stage, we provide both quantitative and qualitative analyses. 
For these investigations, we focus on MED-VT trained on Automatic Video Object Segmentation (AVOS) using the Video-Swin backbone~\cite{liu2022video}. 

\input{tkz_figures/medvt_stdyn}
\begin{figure}[h]
	\begin{center}
		\setlength\tabcolsep{0.7pt}
		\resizebox{0.49\textwidth}{!}{
			\begin{tabular}{ccccccc}
				\includegraphics[width=0.15\textwidth]{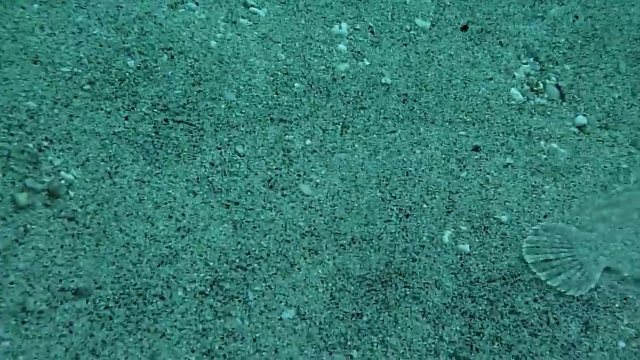}
				&
				\includegraphics[width=0.15\textwidth]{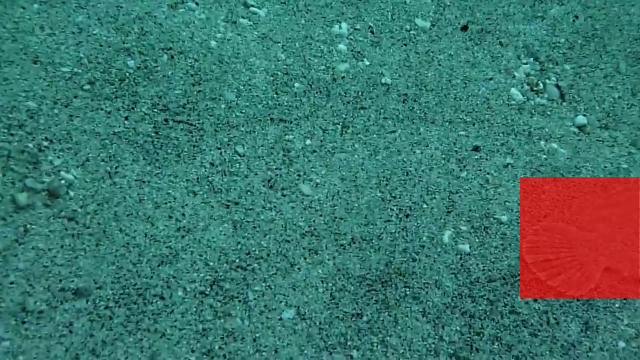}
				& 
				\includegraphics[width=0.15\textwidth]{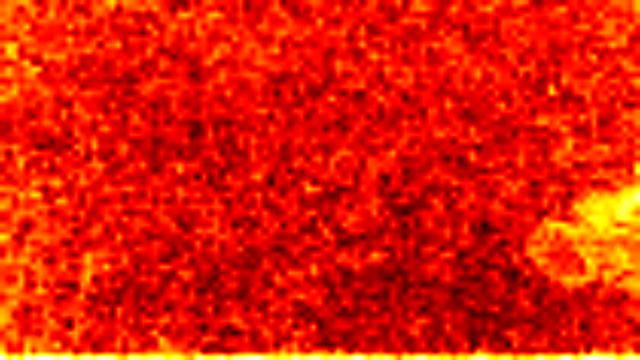}
				&
				\includegraphics[width=0.15\textwidth]{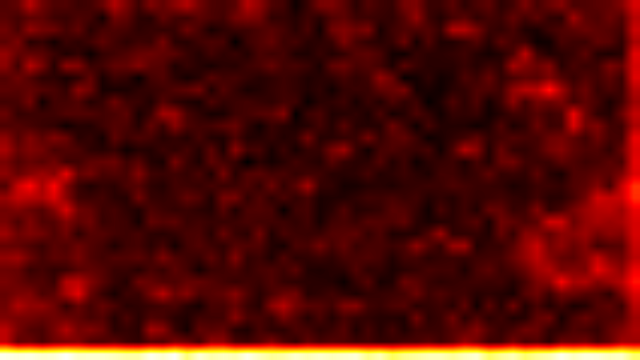}
				&
				\includegraphics[width=0.15\textwidth]{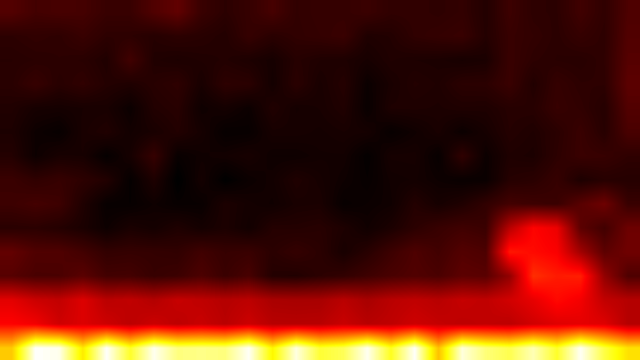}
				&
				\includegraphics[width=0.15\textwidth]{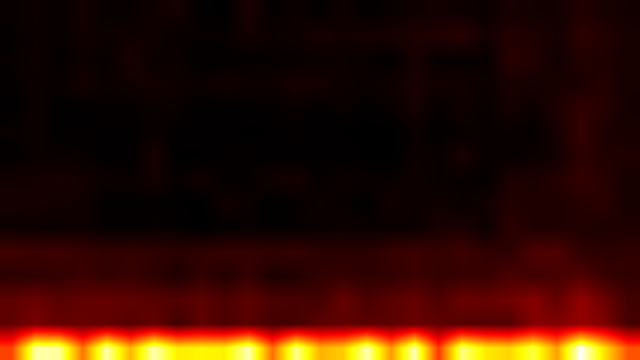}
				\\

				\includegraphics[width=0.15\textwidth]{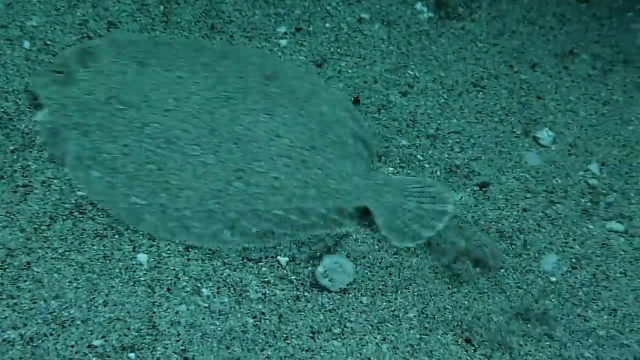}
				&
				\includegraphics[width=0.15\textwidth]{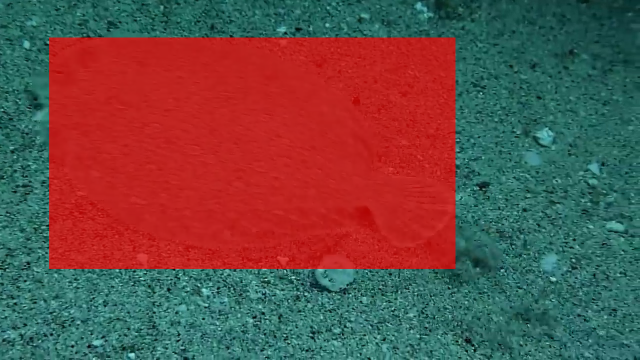}
				& 
				\includegraphics[width=0.15\textwidth]{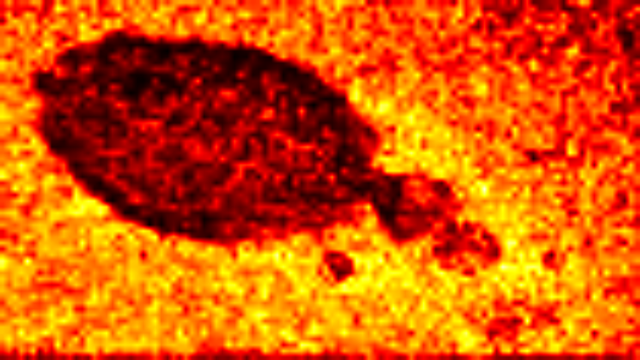}
				&
				\includegraphics[width=0.15\textwidth]{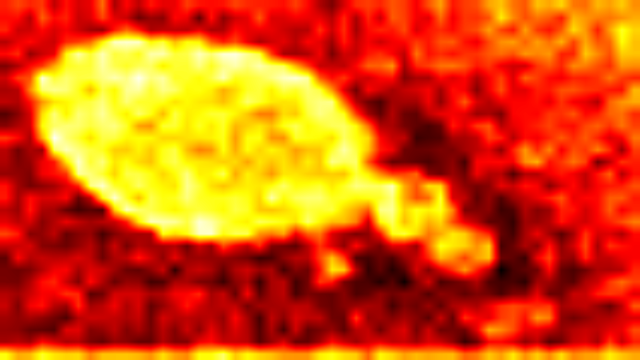}
				&
				\includegraphics[width=0.15\textwidth]{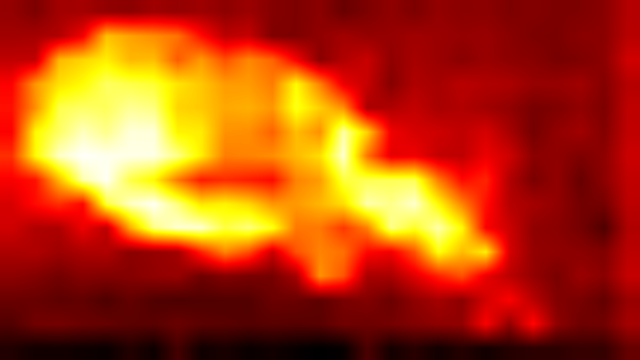}
				&
				\includegraphics[width=0.15\textwidth]{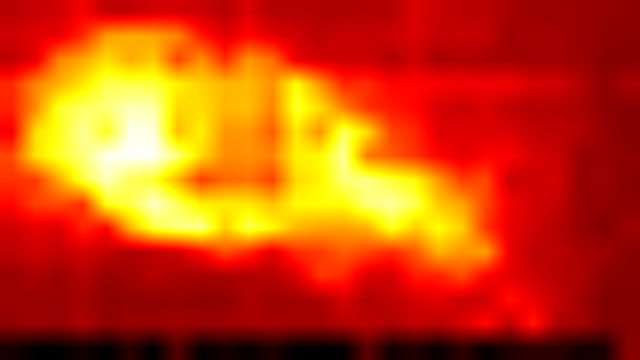}
				\\
				 Image & Ground-Truth & Backbone $F_1$ & Backbone $F_2$ & Backbone $F_3$ & Backbone $F_4$  
                    \\             
		\end{tabular}}
		\caption{Activations (first principle component) at different backbone layers as heat maps. Top and bottom rows show two frames of a camouflaged fish. It is seen that distinct information is captured at different scales (\eg different object parts and textures). Keeping information across all scales yields richest representation. 
        }
		\label{fig:interpret_viz_multiscale}
	\end{center}
\end{figure}

\subsection{Methodology}

\textbf{Quantitative analysis.} To study the latent representations maintained by the backbone, transformer encoder and transformer decoder, we extend a recent method for analysing the extent to which deep models capture static \textit{vs.} dynamic properties in video~\cite{kowal2022deeper}. Here, static is defined as the information that can be captured from a single frame (\eg colour and texture), while the dynamic factor refers to the information conveyed from multiple frames (\eg motion). Together, the static and dynamic factors provide insight into fundamental aspects of video representation.


The original methodology calculated the extent that static \textit{vs.} dynamic properties are represented at layer-wise granularity. This property was calculated in terms of mutual information between pairs of inputs generated from the same video clip, by perturbing either the static or dynamic information. If the mutual information was decreased more during perturbation of one type of information (static or dynamic) compared to other, then that information was deemed to be more characteristic of what is captured at that layer. Static information was perturbed through appearance stylization, \eg~\cite{texler2020interactive}; dynamic information was perturbed through frame shuffling or optical flow jittering. Details are available in the original presentation~\cite{kowal2022deeper}.

In comparison to the original methodology, we employ two necessary modifications for our video object segmentation setup: (i) We modify the dataset used in measuring the static/dynamic bias to be less static biased; (ii) we use a different scheme for perturbing the dynamic information. 
First, in addition to adopting use of the Stylized DAVIS dataset, we extend to generate a stylized version of the MoCA dataset using a similar approach~\cite{texler2020interactive}. The reason for using stylized MoCA is also inspired by the findings of the original work that DAVIS itself is static biased~\cite{kowal2022deeper}. Thus, we chose another dataset that we believe to reduce the static bias by its design that focuses on camouflaged objects, where motion is a critical factor. Second, we repeat the same frame throughout the clip to perturb dynamics in contrast to shuffling frames used in the original methodology. This choice is rooted in our ablation experiments with MED-VT, revealing that in video object segmentation shuffling the frame order is not sufficient to perturb dynamics (Table~\ref{tab:dpiou}); although, it may be the case for action recognition~\cite{kowal2022deeper}. Note, that the previous interpretability study focused on video object segmentation techniques that use appearance (RGB frames) and motion (optical flow). However, our model, MED-VT, relies only on appearance information without optical flow preprocessing; so, optical flow jittering is not applicable. 

For quantitative analysis of label propagation, we employ the Receiver Operating Characteristic (ROC) curve, which supports evaluation of a classification model across different thresholds to show the trade-off between sensitivity and specificity. We use ROC analysis for label propagation because it employs a compact label encoding of the initial predictions in the label space. This encoding intrinsically contains a very different representation compared to the latent features in the intermediate layers. In particular, a quantitative method to interpret classifiers characteristics is suitable for investigating its properties. We show the ROC curves for the original videos \textit{vs.} static or dynamic perturbed videos. For each variant, we extracted the first principal component of the label encoding, computed true positive and false positive rates, and plotted the ROC. 

\textbf{Qualitative analysis.} For qualitative analysis, we visualize MED-VT's internal representations using heat-maps of the intermediate layer features after performing principal component analysis. Since our interest is pixel-wise semantically discriminatory feature generation throughout model stages, we apply principal component analysis to the high-dimensional features and analyze the region-wise distinguishable activation of the main principal component. A rich feature representation for pixel-wise estimation tasks is expected to have features that show activation regionally distinguishable across semantics of the scene. As a representative input for demonstration, we use the $Flounder\_6$ video from the MoCA dataset for the following reasons. First, this sequence represents a very challenging scenario where the object (fish) in the scene is heavily camouflaged and very difficult to distinguish in a single frame. Second, this sequence poses other challenges, where the fish changes motion directions and encompasses various speeds. Third, the object also presents challenges of partially going out of view and re-appearing in the scene. 

\begin{figure*}[h]
	\begin{center}
		\setlength\tabcolsep{0.7pt}
		\def\arraystretch{0.95}
		\resizebox{0.99\textwidth}{!}{
			\begin{tabular}{ccccccccccc}
				\includegraphics[width=0.15\textwidth]{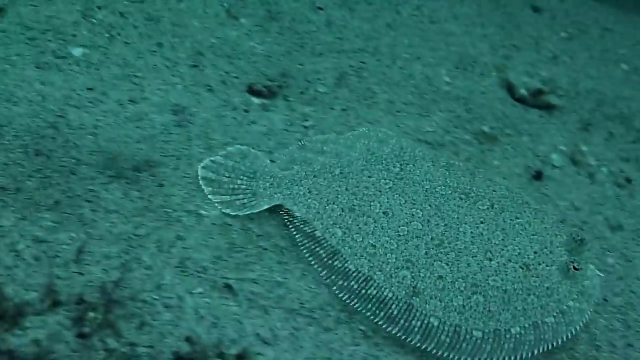}
				&
				\includegraphics[width=0.15\textwidth]{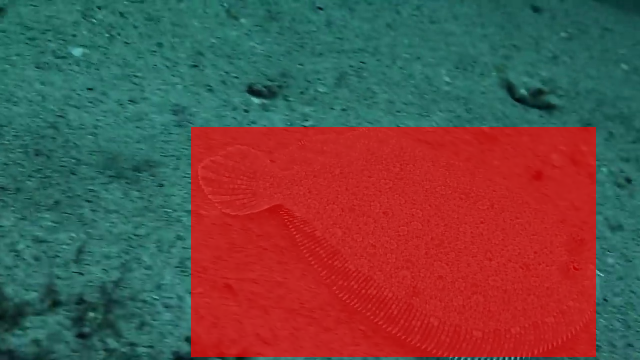}
				& 
				\includegraphics[width=0.15\textwidth]{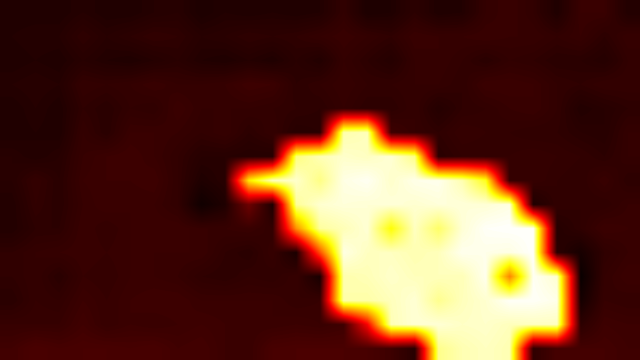}
				&
				\includegraphics[width=0.15\textwidth]{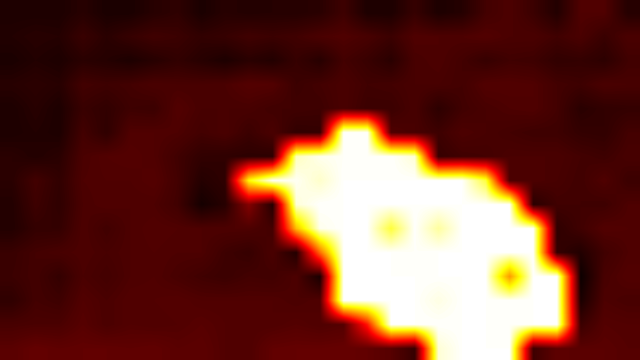}
				&
				\includegraphics[width=0.15\textwidth]{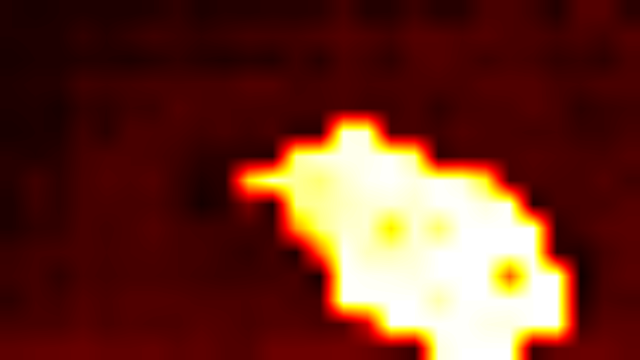}
				&
				\includegraphics[width=0.15\textwidth]{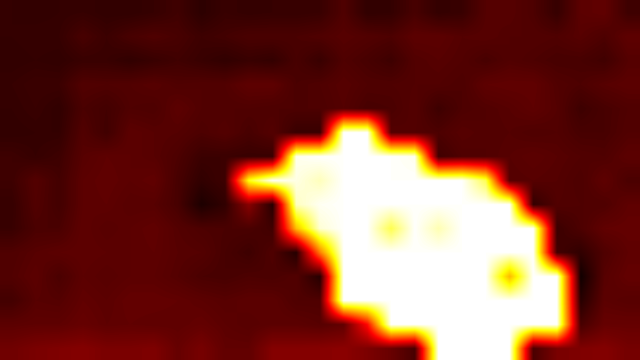}
				&
				\includegraphics[width=0.15\textwidth]{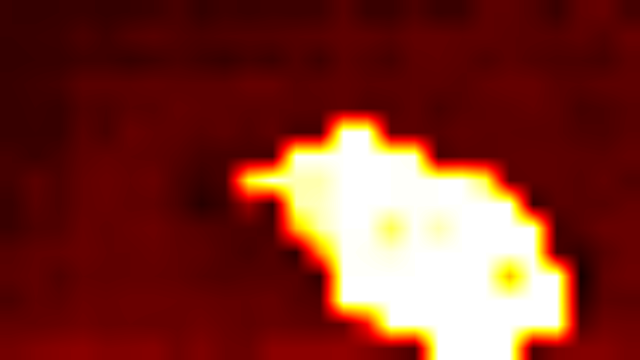}
				&
				\includegraphics[width=0.15\textwidth]{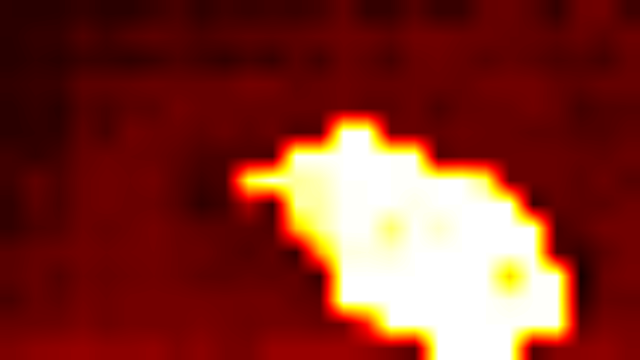}
				&
				\includegraphics[width=0.15\textwidth]{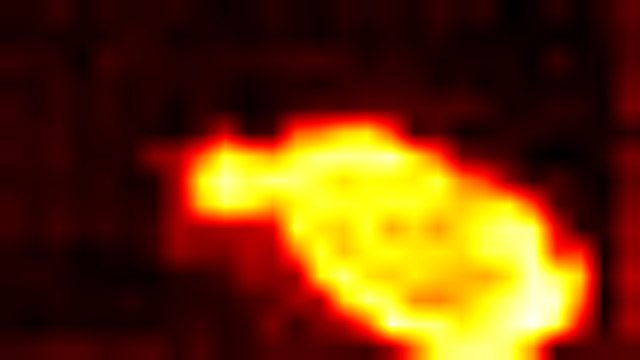}
				&
				\includegraphics[width=0.15\textwidth]{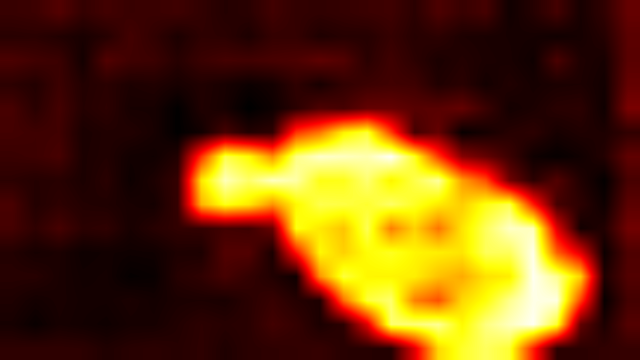}
				&
				\includegraphics[width=0.15\textwidth]{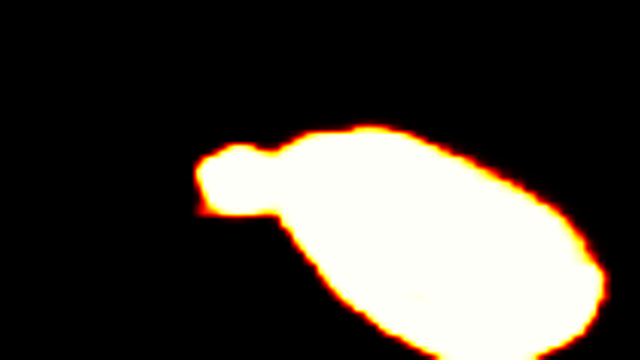}
				\\
				\includegraphics[width=0.15\textwidth]{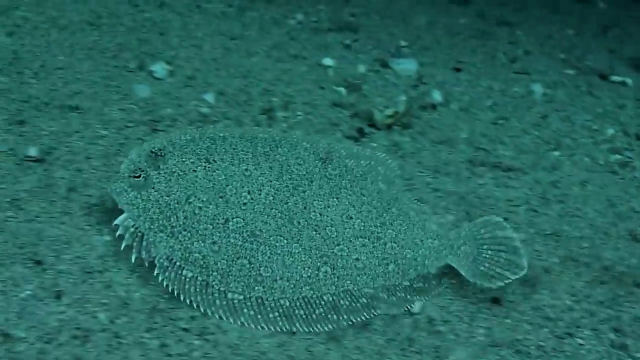}
				&
				\includegraphics[width=0.15\textwidth]{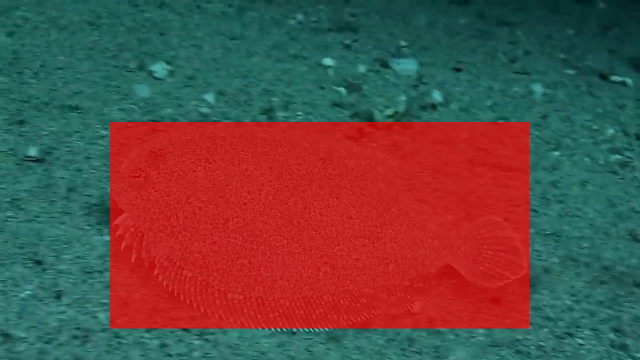}
				& 
				\includegraphics[width=0.15\textwidth]{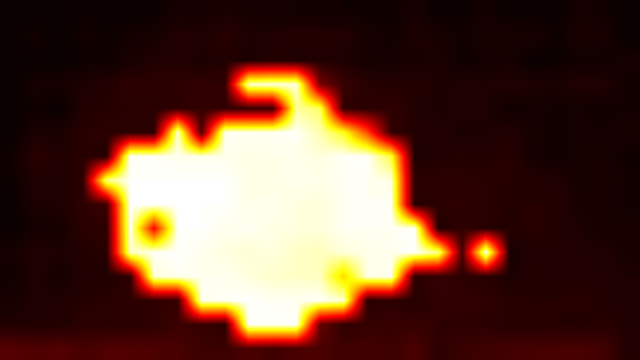}
				&
				\includegraphics[width=0.15\textwidth]{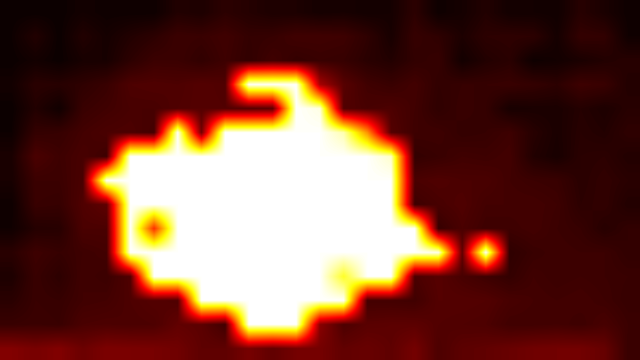}
				&
				\includegraphics[width=0.15\textwidth]{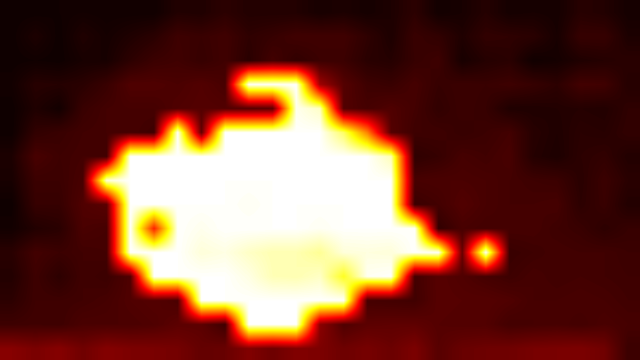}
				&
				\includegraphics[width=0.15\textwidth]{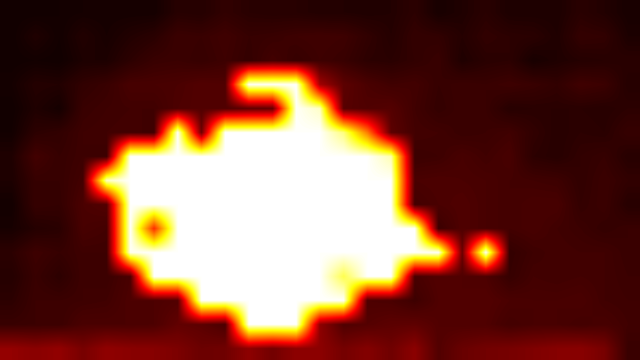}
				&
				\includegraphics[width=0.15\textwidth]{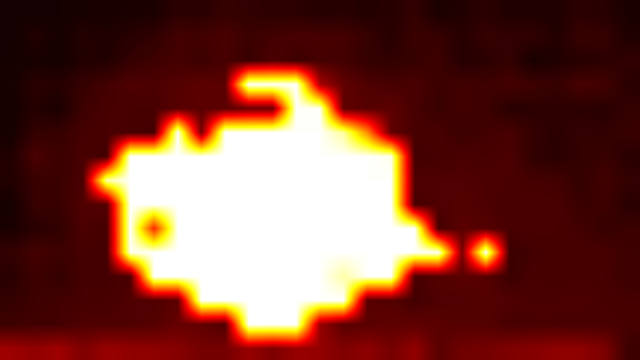}
				&
				\includegraphics[width=0.15\textwidth]{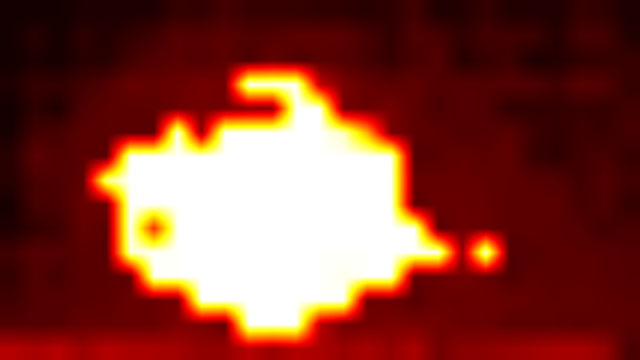}
				&
				\includegraphics[width=0.15\textwidth]{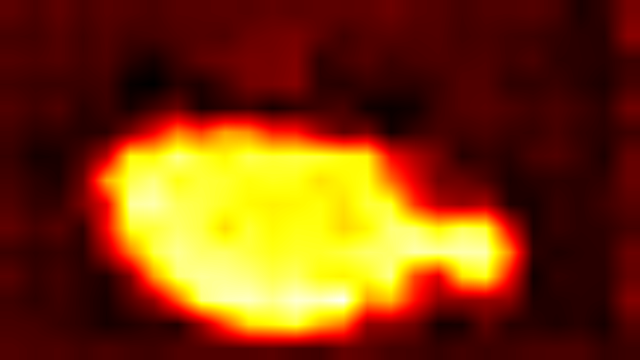}
				&
				\includegraphics[width=0.15\textwidth]{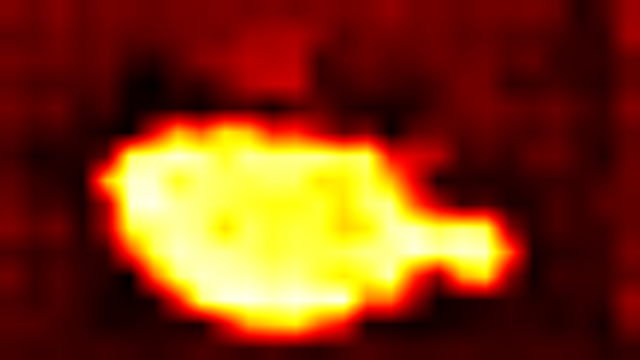}
				&
				\includegraphics[width=0.15\textwidth]{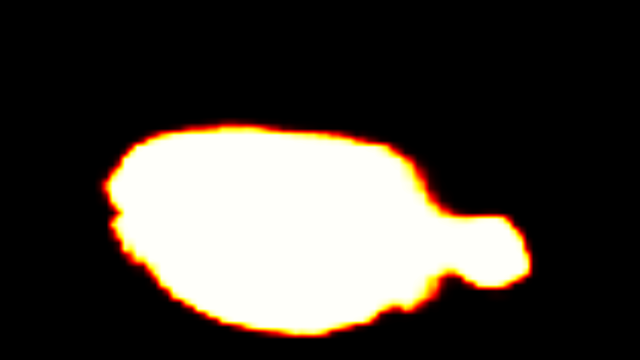}
				\\
				Image & Ground-Truth & WSE $W_{4\times1}$ &WSE $W_{4\times2}$ & WSE $W_{4\times3}$ & WSE $W_{4\times4}$  & WSE $W_{4\times5}$ &WSE $W_{4\times6}$ & WSE $W_{3\times1}$  & BSE $B_{3\times4}$ & Logits $Y$\\             
		\end{tabular}}
		\caption{Activations (first principle component) at different encoder layers as heatmaps. Top and bottom rows show two frames of a camouflaged fish. Left-to-right: Input and bounding box groundtruth, six encoder layers operating on the deepest backbone layer (columns 3-8), between and within scale encoder operating on penultimate background layer (columns 9, 10), logit predictions (column 11). 
  $W_{s\times k}$ denotes $k^{th}$ layer of within scale encoder on stage $s$. $B_{i\times j}$ denotes between scale encoder between stage $i$ and $j$.}
		\label{fig:interpret_viz_msenc}
	\end{center}
\end{figure*}


\subsection{Backbone} 
\label{sec:visual_justification_ms_feat}
\textbf{Quantitative analysis.} 
In Figure~\ref{fig:static_vs_dynamic_bias_tkz} the first four data points in the layer-wise analyses correspond to the four backbone stages. It is seen that  the backbone learns to encode more dynamics and less static information in the MED-VT multiscale model compared to the single scale baseline, with most pronounced differences in stages three and four. 
These results support that using multiscale processing in the downstream transformer compels even the earliest feature extraction, \ie the backbone, to acquire a nuanced understanding of task demands, \eg importance of dynamics in video segmentation, especially where motion is critical, \eg in camouflaged MoCA.


\textbf{Qualitative analysis.} 
The visualizations in Figure~\ref{fig:interpret_viz_multiscale} provides further insights. The backbone features (columns three - six) show that the maximum activation is not always at the last layer. As examples: The first row shows a scenario of severely diminished activation of the object region in the deepest layer feature of the backbone compared to other intermediate layers; the second row reveals a different scenario where the second layer has activation from the object's main body, while other parts (\eg tail) are best captured by deeper layers. These results show that when operating as part of MED-VT the backbone is able to capture important object characterizations across a range of scales and pass all of them forward for subsequent processing, rather than just the deepest layer, which could be lacking in some details.

\begin{figure*}[ht]
	\begin{center}
		\setlength\tabcolsep{0.7pt}
		\def\arraystretch{0.95}
		\resizebox{0.99\textwidth}{!}{
			\begin{tabular}{ccccccccccc}
				\includegraphics[width=0.15\textwidth]{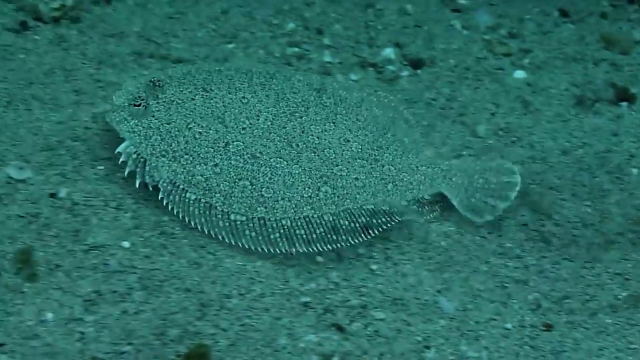}
				&
				\includegraphics[width=0.15\textwidth]{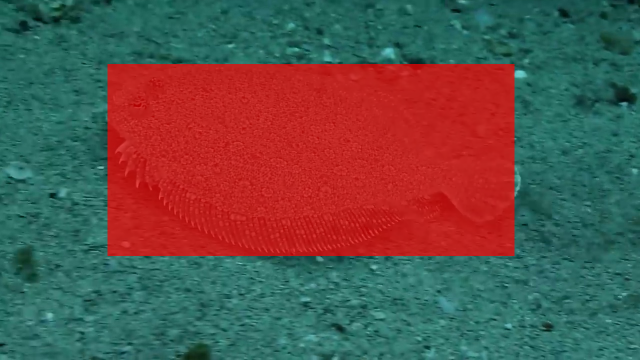}
				& 
				\includegraphics[width=0.15\textwidth]{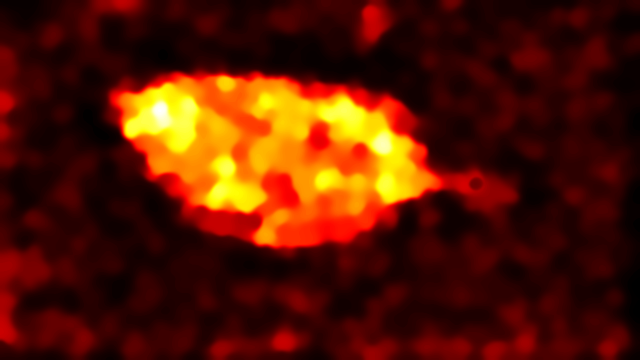}
				&
				\includegraphics[width=0.15\textwidth]{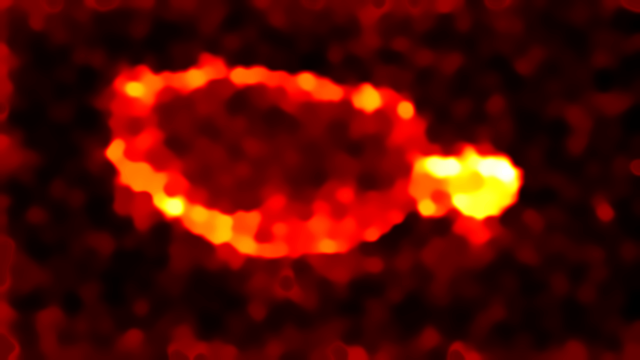}
				&
				\includegraphics[width=0.15\textwidth]{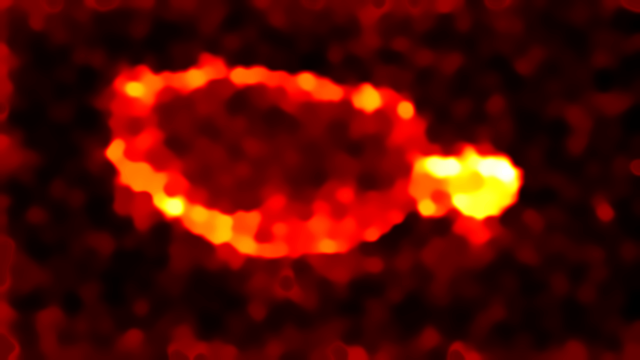}
				&
				\includegraphics[width=0.15\textwidth]{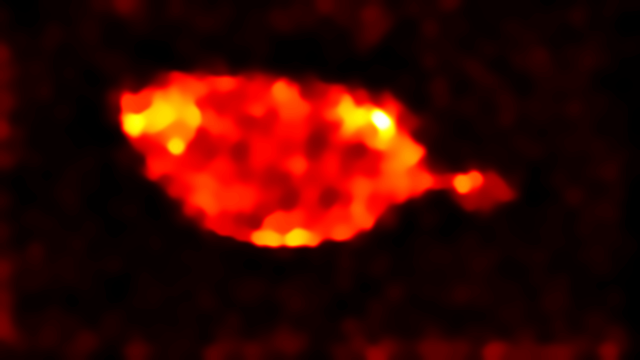}
				&
				\includegraphics[width=0.15\textwidth]{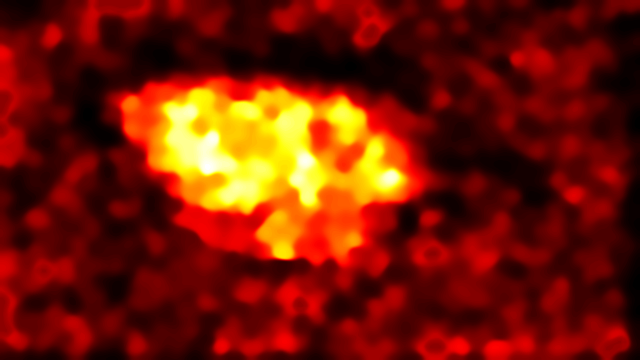}
				&
				\includegraphics[width=0.15\textwidth]{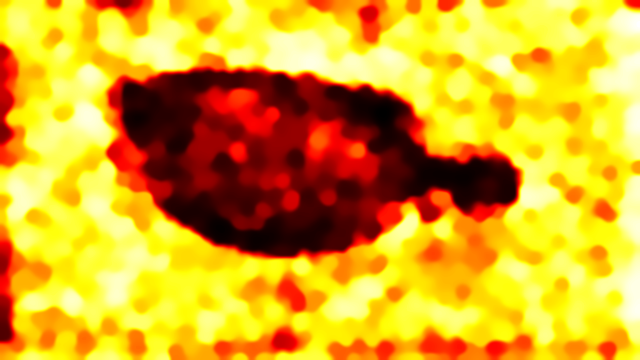}
				&
				\includegraphics[width=0.15\textwidth]{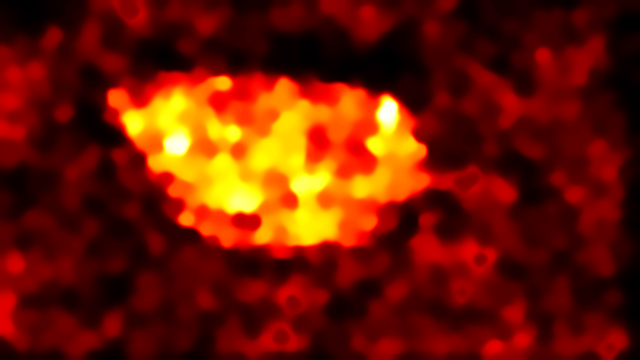}
				&
				\includegraphics[width=0.15\textwidth]{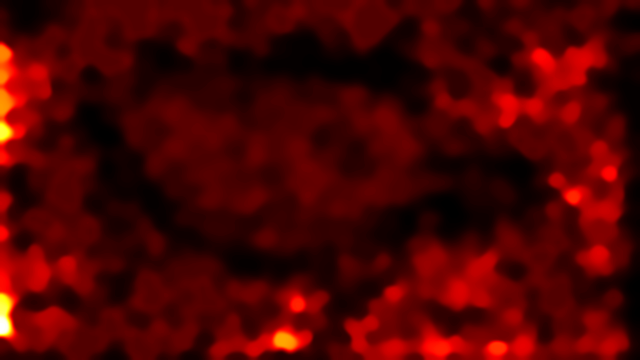}
				&
				\includegraphics[width=0.15\textwidth]{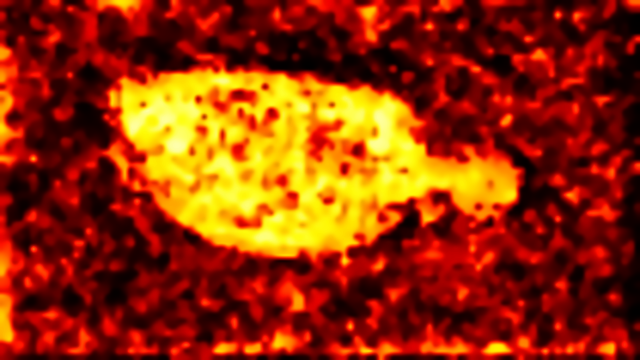}
				\\
				
				\includegraphics[width=0.15\textwidth]{figures/interpretability/flounder_6/img/00076.png}
				&
				\includegraphics[width=0.15\textwidth]{figures/interpretability/flounder_6/gt/00076.png}
				& 
				\includegraphics[width=0.15\textwidth]{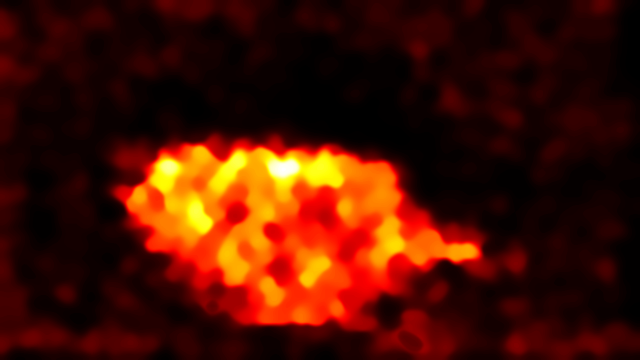}
				&
				\includegraphics[width=0.15\textwidth]{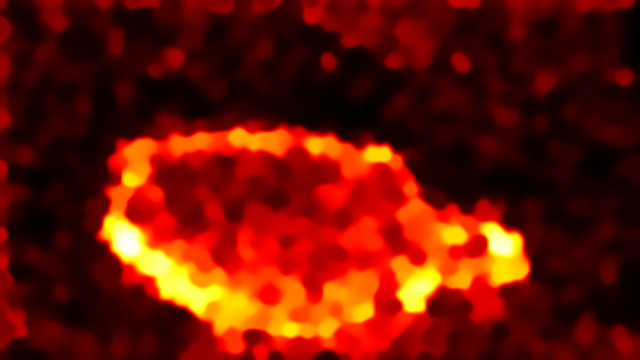}
				&
				\includegraphics[width=0.15\textwidth]{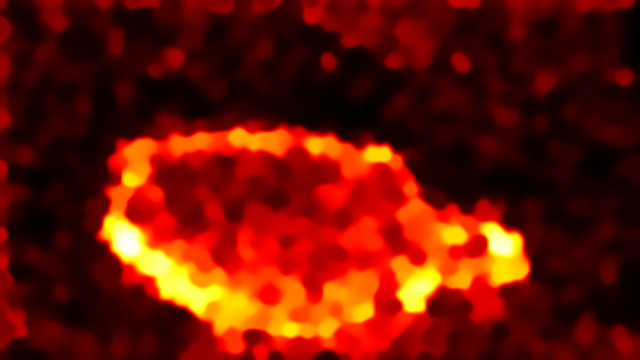}
				&
				\includegraphics[width=0.15\textwidth]{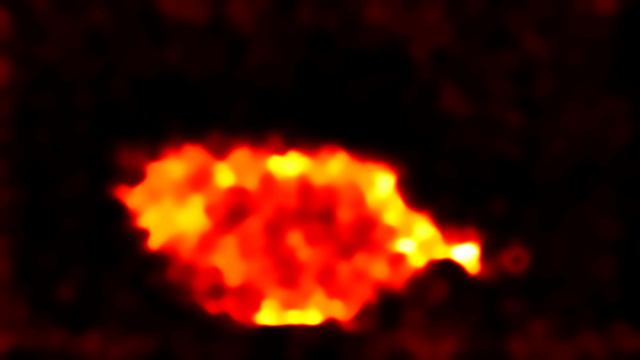}
				&
				\includegraphics[width=0.15\textwidth]{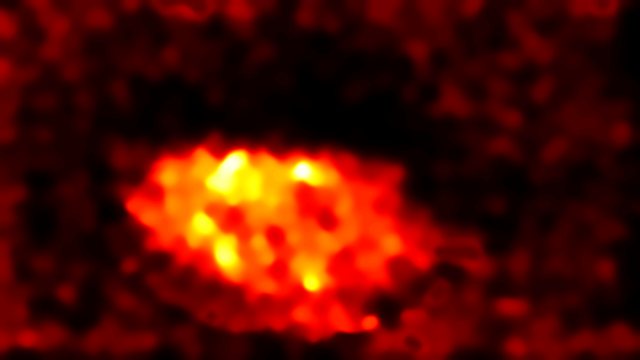}
				&
				\includegraphics[width=0.15\textwidth]{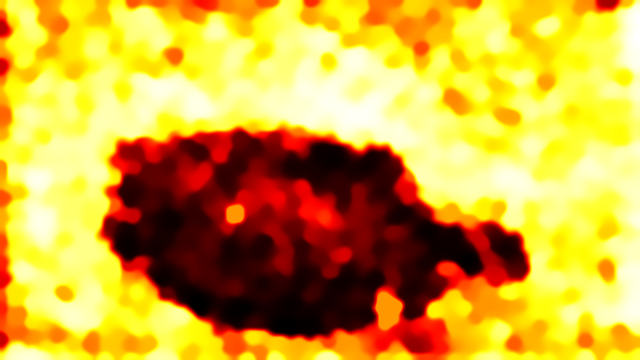}
				&
				\includegraphics[width=0.15\textwidth]{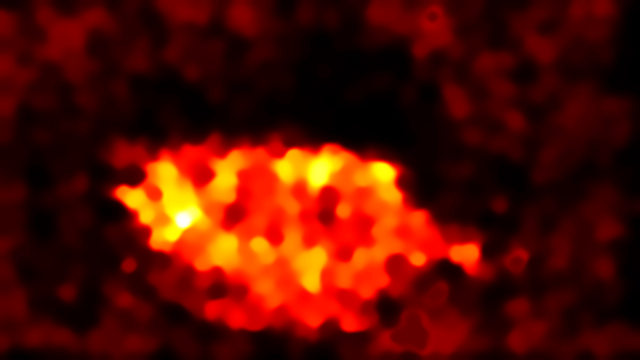}
				&
				\includegraphics[width=0.15\textwidth]{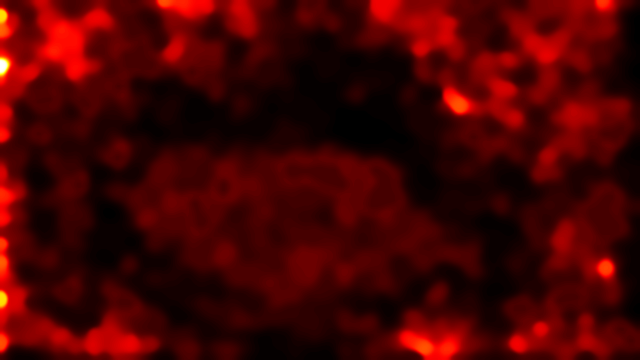}
				&
				\includegraphics[width=0.15\textwidth]{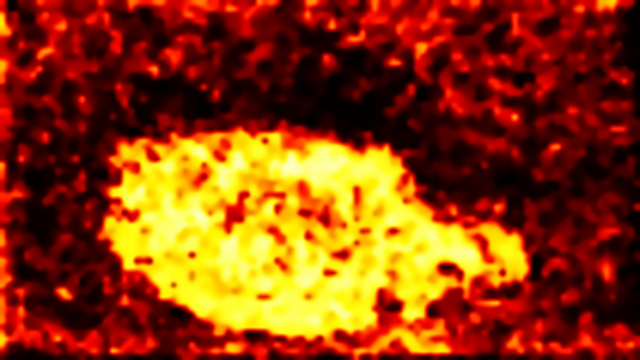}
				\\

				Image & Ground-Truth & Query1 $H_1$ &Query2 $H_2$ & Query3 $H_3$ & Query4 $H_4$  & Query5 $H_5$ &Query6 $H_6$ & Query7 $H_7$ & Query8 $H_8$ & Query PCA \\             
				\end{tabular}}
				\caption{Activations (first principle component) across decoder layers as
heatmaps. Top and bottom rows show two frames of a camouflaged fish.  Left-to-right: Input, Bounding box, decoder attention heads 1-8 ($H_1$ ...  $H_8$) and all attention heads.
                }
				\label{fig:interpret_viz_msdecoder}
				\end{center}
				\end{figure*}


\subsection{Multiscale encoder}

\textbf{Quantitative analysis.} 
In Figure~\ref{fig:static_vs_dynamic_bias_tkz} data points 5-12 correspond to transformer encoder layers. The layer-wise analysis shows that MED-VT consistently increases in dynamic information with increased depth compared to the baseline. 
These results reveal that MED-VT's multiscale encoder captures more dynamic information than its single scale baseline counterpart and it does so progressively with layer depth.

\textbf{Qualitative analysis.} Figure~\ref{fig:interpret_viz_msenc} visualizes different encoder layer activations. It is seen that going deeper in encoder layers provides an increase of distinguishable activation in the object region, with the effect most pronounced in the top row. For both cases, the target object is greatly enhanced compared to the backbone alone, \cf~Figure~\ref{fig:interpret_viz_multiscale}. Comparing encoding on the deepest layer backbone features, $W_{4\times1}$ to encoding on the penultimate layer backbone features, $W_{3\times1}$ reveals complementarity provided by multiscale processing: The penultimate scale provides more precise localization of the object boundaries, but with background noise; the coarser scale ameliorates noise, but compromises on the boundaries. Our multiscale approach can reap the benefits of both scales. Finally, comparison of between scale, $B_{3\times4}$, and within scale, $W_{3\times1}$, encoding shows the ability of between scale to enhance object delineation beyond what is provided by within scale alone, 
 as here is where the scales combine to exploit complementarity. For example, notice how the body of the object 
 better matches the final shape and attains more precise, less blurry boundaries, especially in the upper row.
\input{tkz_figures/lprop_roc}
\begin{figure}[ht]
	\begin{center}
		\setlength\tabcolsep{0.7pt}
		\def\arraystretch{0.75}
		\resizebox{0.48\textwidth}{!}{
			\begin{tabular}{ccccccc}
				\includegraphics[width=0.15\textwidth]{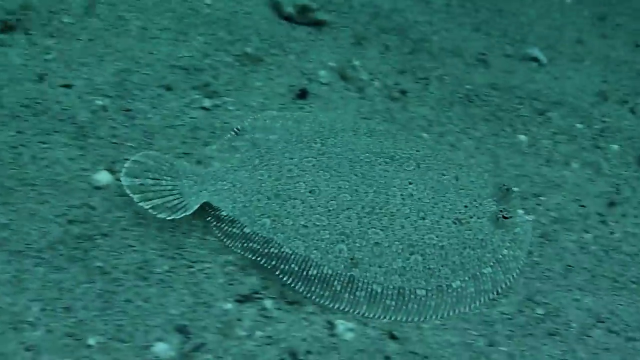}
				&
				\includegraphics[width=0.15\textwidth]{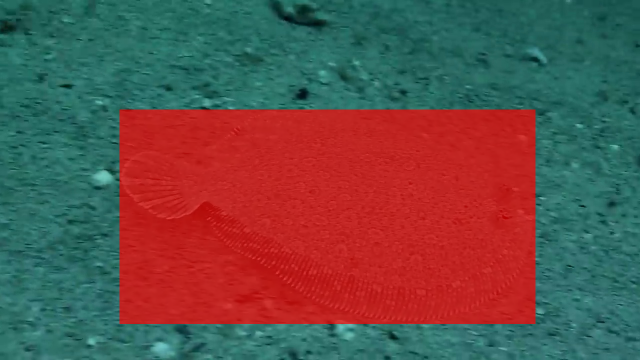}
				& 
				\includegraphics[width=0.15\textwidth]{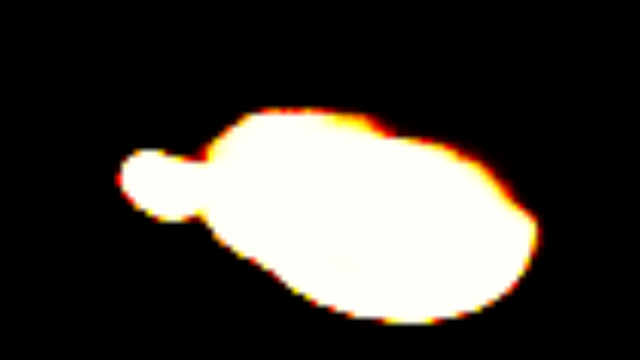}
				&
				\includegraphics[width=0.15\textwidth]{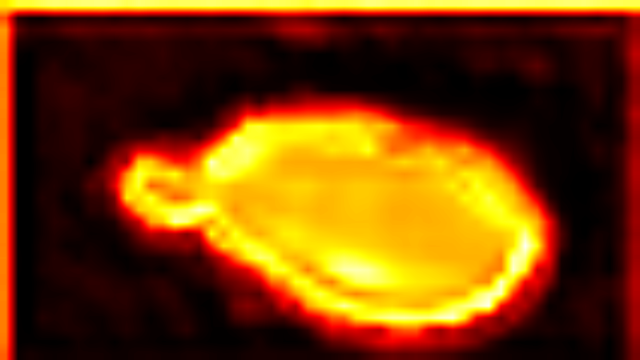}
				&
				\includegraphics[width=0.15\textwidth]{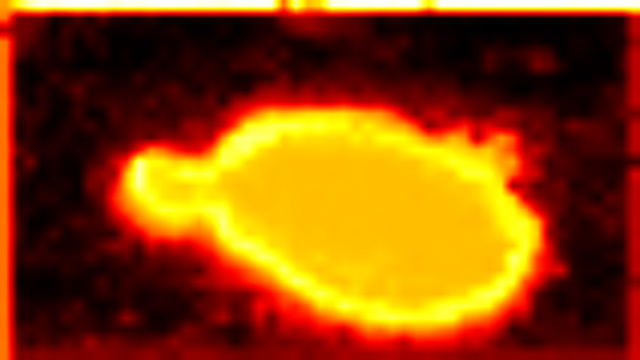}
				& 
				\includegraphics[width=0.15\textwidth]{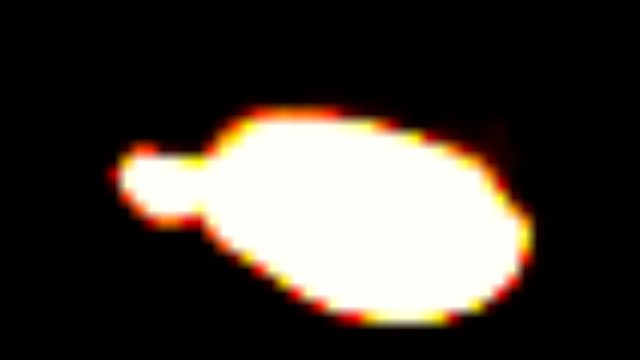}
				\\
				\includegraphics[width=0.15\textwidth]{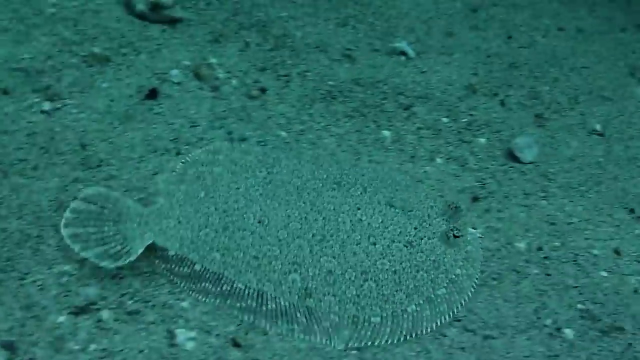}
				&
				\includegraphics[width=0.15\textwidth]{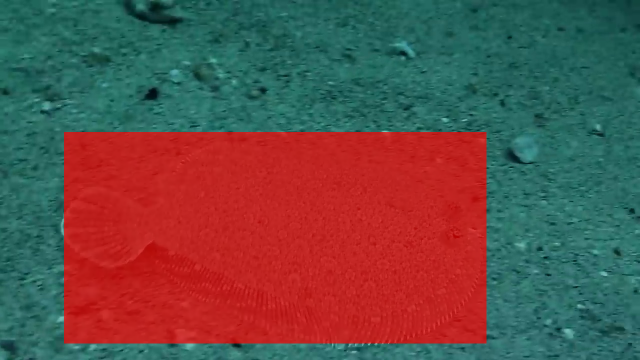}
				& 
				\includegraphics[width=0.15\textwidth]{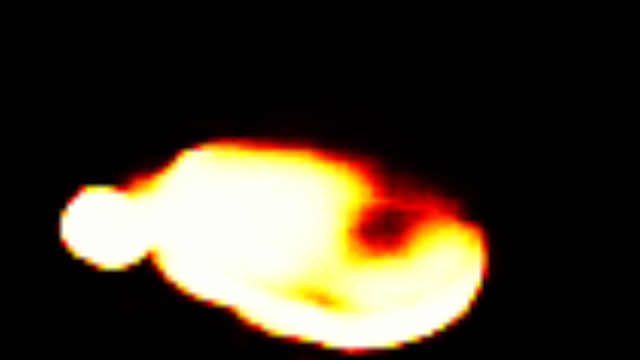}
				&
				\includegraphics[width=0.15\textwidth]{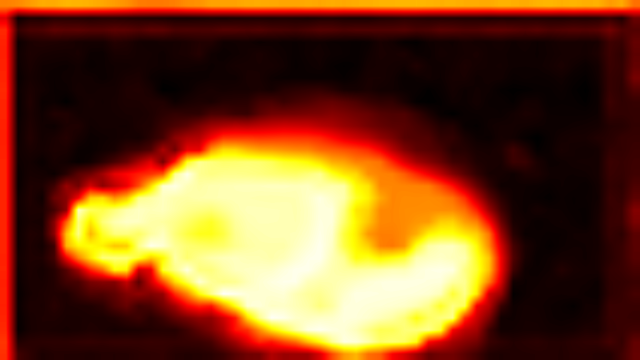}
				&
				\includegraphics[width=0.15\textwidth]{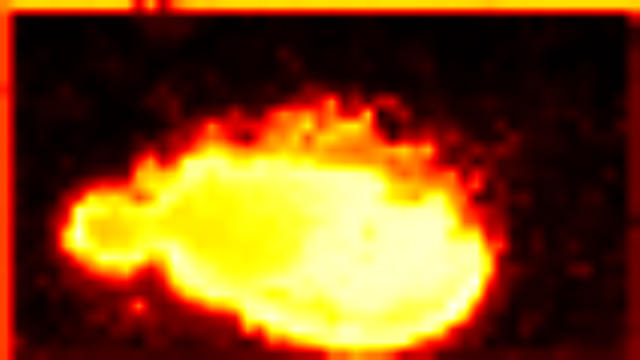}
				& 
				\includegraphics[width=0.15\textwidth]{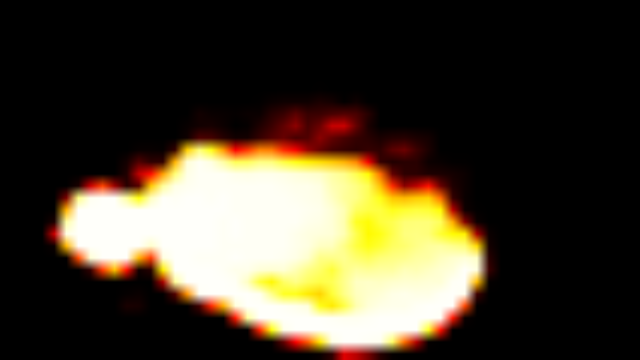}
				\\
				Image & Ground-Truth & Task-head & Mask encoding & Propagator encoding & Label propagator logits \\             
		\end{tabular}}
		\caption{Activations (first principle component) at different label propagation stages as heatmaps. 
  Top and bottom rows show two frames of a camouflaged fish.
Left-to-right: Image, ground-truth, task-head logits, mask-encoding, propagator mask encoding and label propagator. Both rows show label propagation correcting errors in a given frame (\eg missing parts), with second row providing stronger illustration.
        }
		\label{fig:interpret_lprop_stages}
	\end{center}
\end{figure}


\subsection{Multiscale decoder}
\textbf{Quantitative analysis.} In Figure~\ref{fig:static_vs_dynamic_bias_tkz} data points 12-21 correspond to decoder layers. The layer-wise results on MoCA show that the decoder maintains, but does not increase dynamic encoding. Ideally, the decoder's role is generation of adaptive queries without altering the latent representation learned by the encoder; so, the observed behaviour is consistent with that goal. Conversely, the dynamic factor decreases in the decoding stages on DAVIS,  which apparently seems opposite to the expected properties of MED-VT. This outcome could be attributed to the static bias that is inherent in the DAVIS dataset~\cite{kowal2022deeper}. The decoder generates dynamic queries based on encoded features to address the segmentation objective. On the DAVIS dataset, biased towards static samples, decoder queries prioritize static information over dynamics in encoder features. Conversely, the MoCA dataset, which is biased towards dynamic samples, enables the decoder to learn queries emphasizing dynamic factors from encoded features. Overall, these results support that the decoder can capitalize on the information provided by the encoder in a dataset depended fashion.

\textbf{Qualitative analysis.} 
Figure~\ref{fig:interpret_viz_msdecoder} shows the object attention generated by attention blocks, $\mathcal{A}^D$, per head. 
We observe that some attention heads learn to focus on various object aspects, including the object extremities. However, most attention heads generate well-localized saliency maps for the primary objects in the video. Indeed, the majority of attention heads in the examples yield activation for the primary object despite camouflage (\eg, $H_1, H_2, H_3, H_4, H_5, H_7$). In complement, however, $H_8$ and especially $H_6$ capture the background. These results offer insight into the nature of attention generated through decoder queries, a level of detail not previously presented in work on transformer decoders.


\subsection{Label propagation}


\textbf{Quantitative analysis.}
Figure~\ref{fig:lprop_roc_analysis} shows label propagation ROC analysis. The Area Under the ROC (AUROC) values obtained were 0.907, 0.859, and 0.828 for the original, static and dynamic perturbed videos, resp. Thus, both perturbations decrease performance, with the effect most pronounced for dynamic perturbation. These results support that MED-VT 
relies on dynamics for temporally consistent predictions through label propagation. 

\textbf{Qualitative analysis.}
Figure~\ref{fig:interpret_lprop_stages} visualizes label propagation. Both rows (frames) provide examples where label propagation recovers missing parts of the target object, with a more pronounced improvement in the second row. The example in the second row shows a large part of the fish with false negative labels in the task head logits; however, the label propagator associated information from surrounding frames to fill in the missing parts. The initial prediction from the task head considers the latent representation of each pixel independently, thereby limiting the ability to associate the object with the pixels of neighbouring frames. Label propagation associates the labels estimated in neighbouring frames to correct initially false negative labels. 

%% file: tkz_figures/medvt_stdyn.tex
\begin{figure*}[h]  
    \centering
    \begin{minipage}{0.71\textwidth}
    \resizebox{\textwidth}{!}{
    \begin{tikzpicture} \ref{legend_color}
    \begin{groupplot}[group style = {group size = 3 by 1, horizontal sep = 20pt}, width = 6.0cm, height = 6.4cm]
    \nextgroupplot[
      line width=1.0,
        title={\textbf{(a) Stylized MoCA}},
        title style={at={(axis description cs:0.5,0.92)},anchor=north},
        xlabel={Network Layer},
        ylabel={Number of Units (\%)},
        xmin=1, xmax=21,
        ymin=13, ymax=46,
        xtick={1,5,12,21},
        ytick={15, 20, 25, 30, 35, 40},
        x tick label style={font=\footnotesize},
        y tick label style={font=\footnotesize},
        x label style={at={(axis description cs:0.5,0.06)},anchor=north,font=\small}, 
        y label style={at={(axis description cs:0.17,.5)},anchor=south,font=\normalsize},
        width=6.5cm,
        height=5cm,        
        ymajorgrids=false,
        xmajorgrids=false,
        major grid style={dotted,green!20!black},
    ]
    \addplot[line width=1.2pt,solid,mark options={scale=0.8,solid},color=blue!100,mark=triangle*,]
        coordinates {(1,22.4)(2,22.1)(3,25.0)(4,24.7)(5,26)(6,26)(7,26.6)(8,26.8)(9,26.6)(10,26.8)(11,26.2)(12,26.3)(13,27.1)(14,26.3)(15,26.6)(16,26.3)(17,26.8)(18,26.8)(19,27.1)(20,26.6)(21,26.4)};
    \addplot[line width=1.2pt,mark size=1.1pt,color=red!100,mark=triangle*,]
        coordinates {(1,38.3)(2,37.8)(3,32.8)(4,33.3)(5,29.4)(5,29.7)(6,29.9)(7,29.9)(8,29.7)(9,29.7)(10,28.9)(11,29.3)(12,28.6)(13,26.8)(14,27.1)(15,27.1)(16,27.3)(17,25.9)(18,26.3)(19,26.2)(20,26.6)(21,27.1)};
    \addplot[line width=1.2pt,dashed,mark options={scale=0.8,solid},color=blue!100,mark=diamond*,]
        coordinates {(1,22.9)(2,22.7)(3,24.0)(4,23.4)(5,23.7)(6,23.7)(7,23.7)(8,24.0)(9,24.0)(10,24.2)(11,24.5)(12,24.5)(13,23.2)(14,23.4)(15,23.2)(16,23.2)(17,23.4)(18,23.2)(19,23.2)(20,23.2)(21,23.2)};
    \addplot[line width=1.2pt,dashed, mark options={scale=0.8,solid},color=red!100,mark=diamond*,]
        coordinates {(1,37.8)(2,37.8)(3,34.9)(4,34.6)(5,29.7)(6,29.4)(7,29.6)(8,29.4)(9,28.9)(10,28.7)(11,28.8)(12,28.1)(13,29.4)(14,29.2)(15,29.4)(16,29.4)(17,29.4)(18,29.7)(19,29.9)(20,29.9)(21,29.7)};
\hspace{5pt}
\nextgroupplot[
      line width=1.0,
        title={\textbf{(b) Stylized DAVIS}},
        title style={at={(axis description cs:0.5,0.92)},anchor=north,font=\normalsize},
        xlabel={Network Layer},
        xmin=1, xmax=21,
        ymin=10, ymax=48,
        xtick={1,5,12,21},
        ytick={15, 20, 25, 30, 35, 40},
        x tick label style={font=\footnotesize},
        y tick label style={font=\footnotesize},
        x label style={at={(axis description cs:0.5,0.06)},anchor=north,font=\small},   
        width=6.5cm,
        height=5cm,        
        ymajorgrids=false,
        xmajorgrids=false,
        major grid style={dotted,green!20!black},
        legend style={
        nodes={scale=0.87, transform shape},
        cells={anchor=west},
        legend style={at={(5.2,3.5)},anchor=south,row sep=0.15pt}, font =\normalsize},
        legend image post style={scale=0.9},
        legend columns=4,
        legend to name=legend_color,
    ]
    \addplot[line width=1pt,solid,mark options={scale=0.9,solid},color=blue!100,mark=triangle*,forget plot]
        coordinates {(1,18.8)(2,19.5)(3,25)(4,26)(5,31.2)(6,31.2)(7,32.3)(8,32.6)(9,32.3)(10,32.1)(11,26.8)(12,26.6)(13,26.9)(14,24.7)(15, 24.7)(16,23.7)(17,24.0)(18,23.7)(19,23.7)(20,23.7)(21,24.0)};
    \addplot[line width=1pt,mark options={scale=0.9,solid},color=red!100,mark=triangle*,forget plot]
        coordinates {(1,39.1)(2,37.8)(3,31.2)(4,30.2)(5,26.6)(6,26.8)(7,26.6)(8,26.6)(9,26.6)(10,26.6)(11,29.4)(12,28.1)(13,27.4)(14,26.6)(15,25)(16,26)(17,24.5)(18,25.3)(19,25)(20,26.8)(21,27.9)};
    \addplot[line width=1.2pt,dashed,mark options={scale=0.8,solid},color=blue!100,mark=diamond*,]
        coordinates {(1,18.8)(2,19.3)(3,23.7)(4,24.5)(5,22.1)(6,22.4)(7,22.1)(8,21.9)(9,21.6)(10,21.6)(11,21.4)(12,21.6)(13,21.1)(14,21.4)(15,21.4)(16,21.4)(17,21.6)(18,21.4)(19,21.4)(20,21.4)(21,21.4)};
    \addplot[line width=1.2pt,dashed,mark options={scale=0.8,solid},color=red!100,mark=diamond*,]
        coordinates {(1,39.1)(2,37.8)(3,30.5)(4,28.6)(5,25.3)(6,24.7)(7,24.2)(8,23.2)(9,22.7)(10,22.4)(11,22.1)(12,21.9)(13,25.3)(14,23.7)(15,24.2)(16,25)(17,24.5)(18,25)(19,25.3)(20,25.3)(21,25.3)};
    
    \addlegendimage{line width=1.2pt,color=red}\label{pgfplots:ar_stat}
    \addlegendentry[color=black]{Static}
    \addlegendimage{line width=1.2pt,solid,color=blue}\label{pgfplots:ar_dyn}
    \addlegendentry[color=black]{Dynamic}
    \addlegendimage{only marks,mark=triangle*,mark size=2.2pt,color=black!70}\label{pgfplots:ar_c1r1}
    \addlegendentry[color=black]{MED-VT  \hspace{1cm}}
    \addlegendimage{only marks,mark=diamond*,mark size=2.3pt,color=black!70}\label{pgfplots:ar_c1r4}
    \addlegendentry[color=black]{Baseline  \hspace{1cm}}
    \end{groupplot}
    \end{tikzpicture}
    }    
\end{minipage}

\caption[Static \textit{vs.} dynamic bias analysis]{Static vs. dynamic bias analysis of MED-VT compared to our single scale baseline on MoCA and DAVIS datasets. Left: MoCA layer-wise analysis, Right: DAVIS layer-wise analysis. Percentages per layer need not sum to $100$, because units can be biased to neither static nor dynamic information. 
  Overall, the analysis indicates that multiscale encoding-decoding enriches dynamic information in encoding that stays consistent during decoding. 
  }
  \label{fig:static_vs_dynamic_bias_tkz}
\end{figure*}

%% file: tkz_figures/lprop_roc.tex
\begin{figure}
\centering

\resizebox{0.31\textwidth}{!}{
\begin{tikzpicture}
\begin{axis} [
     title={\textbf{ROC Curve}},
     title style={at={(axis description cs:0.6,0.2)}, font=\tiny},
     xlabel={False-positive-rate},
     ylabel={True-positive-rate},
     xmin=0.0, xmax=1.0,
     ymin=0.0, ymax=1.0,
     xtick={0, 0.2, 0.4, 0.6, 0.8, 1.0},
     ytick={0, 0.2, 0.4, 0.6, 0.8, 1.0},
     x tick label style={font=\tiny},
     y tick label style={font=\tiny},
     x label style={at={(axis description cs:0.5,0.07)},anchor=north,font=\tiny},
     y label style={at={(axis description cs:0.17,.5)},anchor=south,font=\tiny},
     width=6.5cm,
     height=5cm,
     ymajorgrids=false,
     xmajorgrids=false,
     line legend,
     legend style={
        nodes={scale=0.86, transform shape},
        cells={anchor=west},
        legend style={at={(0.6,0.03)},anchor=south,row sep=0.01pt}, font =\tiny} ,
     legend image post style={scale=0.5},
     legend columns=2,
] 
\addplot[line width=0.5pt, color=red, style=solid] table [x=fpOrg, y=tpOrg, col sep=comma]{data.csv};
\addplot[line width=0.5pt, color=blue, style=solid] table [x=fpApp, y=tpApp, col sep=comma]{data.csv};
\addplot[line width=0.5pt, color=green, style=solid] table [x=fpMot, y=tpMot, col sep=comma]{data.csv};
\addplot[line width=0.5pt, color=black, style=solid] coordinates {(0,0)(0.1,0.1)(0.2,0.2)(0.3,0.3)(0.4,0.4)(0.5,0.5)(0.6,0.6)(0.7,0.7)(0.8,0.8)(0.9,0.9)(1.0,1.0)};

\legend{ \textbf{Original clip}, \textbf{Appearance perturbed},\textbf{Motion perturbed},\textbf{Chance level}}

\end{axis}
\end{tikzpicture}
}%
\caption{Receiver Operating Characteristic (ROC) curve for the label propagator encoding comparing three scenarios: i) Original input images ii) Appearance perturbed input clip iii) Motion perturbed input clip. 
The results indicate performance drop for both appearance and motion perturbation, with higher impact of motion perturbation. 
}
\label{fig:lprop_roc_analysis}
\end{figure}

%% file: medvt/7_conclusion.tex
\section{Conclusion}\label{con}

A novel multiscale video transformer (MED-VT) to unify multiscale processing throughout encoding and decoding has been introduced, along with many-to-many label propagation for temporal consistency. Furthermore, a seamless integration of an auxiliary modality (\eg audio) 
at encoding and context guided query generation at decoding 
has been included for multimodal video segmentation (MED-VT++). The key motivations of the design choices and efficacy of the components have been documented via an  interpretability analysis on the intermediate features. Our encoding provides temporally consistent features derived from only RGB input (\ie without optical flow), while decoding yields semantically informed, precise localization. Our analysis shows that our multiscale approach not only improves the delineation of object boundaries, but also enables detection and localization of camouflaged moving objects. MED-VT has been instantiated on three unimodal and one multimodal video prediction tasks to yield state-of-the-art performance on multiple datasets. The generality of MED-VT yields potential for additional video-based dense prediction tasks where there may be little known a priori about the objects of interest, yet precise delineation is desired (\eg anomalous behaviour detection in video \cite{kim2009,yu2017,nguyen2019}).